\documentclass[lettersize,journal]{IEEEtran}
\usepackage{url}
\usepackage{subfigure}
\usepackage[numbers,sort&compress]{natbib}
\usepackage[numbers]{natbib}
\usepackage{bm}
\usepackage{amssymb}
\usepackage{rotating}
\usepackage{amsmath}
\usepackage{algorithm}
\usepackage{algorithmicx}
\usepackage{algpseudocode}
\usepackage{makecell}
\usepackage{graphicx}
\usepackage{textcomp}
\usepackage{array} 
\usepackage{caption, subcaption}
\usepackage{longtable}
\usepackage{booktabs}
\usepackage{float}
 \usepackage{threeparttable}
 \usepackage{stfloats}
 \usepackage{float}
 \usepackage{epstopdf}
 \usepackage{tabularx}
\usepackage{adjustbox}
\usepackage{arydshln}
\usepackage{booktabs}
\usepackage{multirow}

\begin{document}

\title{Semi-supervised Symmetric Non-negative Matrix  Factorization with Low-Rank Tensor Representation}

\author{Yuheng~Jia,~\IEEEmembership{Member,~IEEE,}
        Jia-Nan~Li,
        Wenhui~Wu,~\IEEEmembership{Member,~IEEE,}
    Ran~Wang,~\IEEEmembership{Senior~Member,~IEEE}
\thanks{This work was supported in part by the National Natural Science Foundation of China (Grant 62106044, 62176160, 62376162), in part by Natural Science Foundation of Jiangsu Province under Grant BK20210221, in part by the Guangdong Basic and Applied Basic Research Foundation (Grants 2024B1515020109 and 2022A1515010791), and in part by the Guangdong Provincial Key Laboratory (Grant 2023B1212060076). (Corresponding author: Ran Wang)}
\thanks{Yuheng Jia and Jia-Nan Li are with the School of Computer Science and Engineering, Southeast University, Nanjing 211189, China, and also with the Key Laboratory of New Generation Artificial Intelligence Technology and Its Interdisciplinary Applications (Southeast University), Ministry of Education, China. (e-mail: yhjia@seu.edu.cn; jiananli@seu.edu.cn).}
\thanks{Wenhui Wu is with the College of Electronics and Information Engineering, Shenzhen University, Shenzhen 518060, China (e-mail: wuwenhui@szu.edu.cn).}
\thanks{Ran Wang is with the School of Mathematical Sciences, Shenzhen University, Shenzhen 518060, China, with the Guangdong Provincial Key Laboratory of Intelligent Information Processing, Shenzhen University, Shenzhen 518060, China, and also with the Shenzhen Key Laboratory of Advanced Machine Learning and Applications, Shenzhen University, Shenzhen 518060, China. (e-mail: wangran@szu.edu.cn).}
}
 
\markboth{Journal of \LaTeX\ Class Files,~Vol.~14, No.~8, August~2021}%
{Shell \MakeLowercase{\textit{et al.}}: Semi-supervised Symmetric Matrix Factorization with Low-Rank Tensor Representation}

\IEEEpubid{
  \begin{minipage}{\textwidth}
  \centering
    \copyright~2024 IEEE. Personal use of this material is permitted.
    \newline
    However, permission to use this material for any other purposes must be obtained from the IEEE by sending an email to pubs-permissions@ieee.org.
  \end{minipage}
}


\maketitle

\begin{abstract}
Semi-supervised symmetric non-negative matrix factorization (SNMF) utilizes the available supervisory information (usually in the form of pairwise constraints) to improve the clustering ability of SNMF. The previous methods introduce the pairwise constraints from the local perspective, i.e., they either directly refine the similarity matrix element-wisely or restrain the distance of the decomposed vectors in pairs according to the pairwise constraints, which overlook the global perspective, i.e., in the ideal case, the pairwise constraint matrix and the ideal similarity matrix possess the same low-rank structure. To this end, we first propose a novel semi-supervised SNMF model by seeking low-rank representation for the tensor synthesized by the pairwise constraint matrix and a similarity matrix obtained by the product of the embedding matrix and its transpose, which could strengthen those two matrices simultaneously from a global perspective. We then propose an enhanced SNMF model, making the embedding matrix tailored to the above tensor low-rank representation. We finally refine the similarity matrix by the strengthened pairwise constraints. We repeat the above steps to continuously boost the similarity matrix and pairwise constraint matrix, leading to a high-quality embedding matrix. Extensive experiments substantiate the superiority of our method. The code is available at https://github.com/JinaLeejnl/TSNMF.
\end{abstract}


\section{Introduction}\label{sec:introduction}
\IEEEPARstart{S}{ymmetric} non-negative matrix factorization (SNMF) \cite{kuang2015symnmf} takes a non-negative matrix $S \in \mathbb{R}^{n \times n}$ as input, and decomposes it as the product of two identical non-negative matrices, i.e.,
\begin{equation}
\label{eq4}
    \min _{V}\left\|S-V V^{\top}\right\|_{F}^{2}, \quad \text { s.t. } V \geq 0,
\end{equation}
where $V \in \mathbb{R}^{n \times k}$ is the embedding matrix, $V\geq 0$ means each element of $V$ is no less than zero, and $n$ and $k$ indicate the number of the samples and the feature dimension of the embedding matrix. When $S$ denotes a similarity matrix of a group of samples, SNMF becomes a well-known graph clustering method \cite{kuang2012symmetric}. For a typical graph clustering method like spectral clustering (SC) \cite{kuang2015symnmf}, \cite{ng2001spectral}, an embedding matrix is first generated, and then a post-processing like $K$-means \cite{hartigan1979algorithm} is required to be performed on the embedding matrix to get the final clustering result. Different from SC, SNMF can directly generate the clustering results utilizing the calculated embedding matrix owning to the non-negative constraints on $V$ \cite{kuang2012symmetric}. Due to this favorable point, SNMF has been applied to many applications like pattern clustering in gene expression \cite{6061964}, community detection \cite{shi2015community}, multi-document summarization \cite{wang2008multi}, etc. Many variants have also emerged based on SNMF. For instance, \citet{luo2021highly} proposed pointwise-mutual-information-incorporated and graph-regularized SNMF (PGSNMF). By implementing SNMF while preserving the geometrical information of the data, \citet{8637461} proposed graph regularized SNMF.

Many applications in the real world have a small amount of supervisory information available. For example, in the video face clustering, two faces in the same video frame cannot belong to the same person \cite{wu2013constrained}. That supervisory information can be generally transformed into two kinds of pairwise constraints, i.e., must-link (ML) and cannot-link (CL), which respectively indicate two data samples belong to the same class or just the opposite. Incorporating this supervisory information can improve the clustering ability of SNMF, which is known as semi-supervised SNMF \cite{chen2008non}. Recently, many semi-supervised SNMF methods have been proposed. For example, \citet{6985550} proposed to introduce MLs through a graph Laplacian regularization, hoping the embeddings of two samples with an ML to be close to each other. \citet{7167693} proposed SNMF-based constrained clustering (SNMFCC), which restricts the inner product of the embeddings of two samples with an ML to be large while that with a CL to be small. By simultaneously completing embedding learning and similarity matrix construction, \citet{8361078} proposed pairwise constraint propagation-induced SNMF (PCPSNMF). \citet{9543530} added an additional sparse regularizer and an additional smoothness regularizer on PCPSNMF. \citet{9013063} proposed the semi-supervised adaptive SNMF (SANMF) to emphasize CLs. See Section \ref{Related Work} for details on these algorithms.

\IEEEpubidadjcol{Existing semi-supervised SNMF methods introduce the supervisory information from a local perspective. For example, in \cite{6985550} and \cite{7167693}, if there exists an ML (resp. a CL) between the $i$-th and the $j$-th samples, their embeddings $V(i,:)$ and $V(j,:)$ are enforced to be similar (resp. dissimilar) to each other. In \cite{8361078,9543530,9013063}, they adjust the value of the similarity matrix $S$ according to the pairwise constraints, i.e., the similarity should be large for two samples with an ML and small for those with a CL. As these methods incorporate the pairwise constraints locally, i.e., element-wisely or sample-to-sample pair-wisely, we argue the global relationship between the pairwise constraints and the embedding matrix is ignored, which leads to inadequate utilization of supervisory information.}

\begin{figure*}[!t]
\centering
\includegraphics[width=5in]{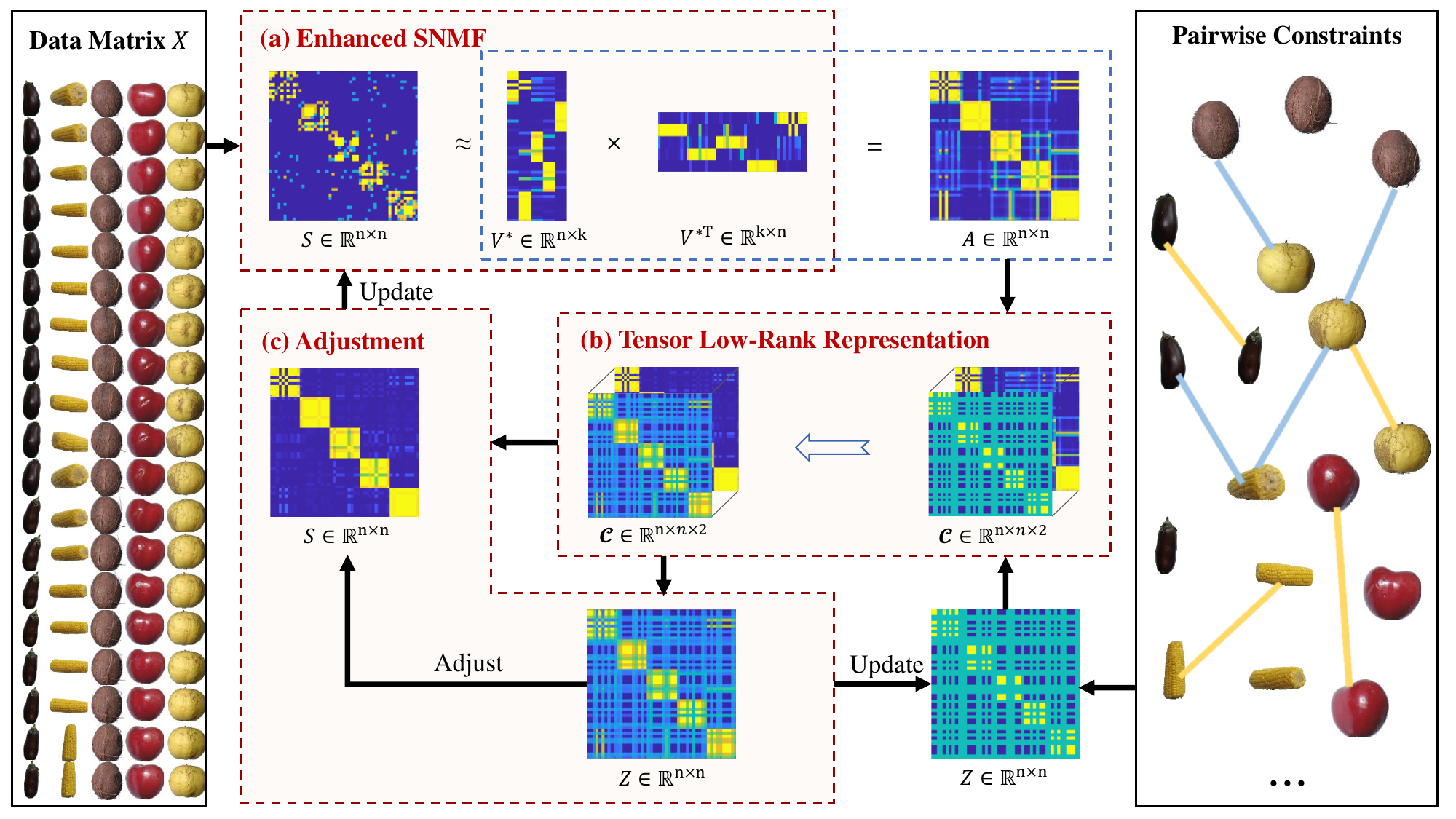}\\
\caption{Workflow of TSNMF. Given the data matrix $X$, the initial similarity matrix $S$ is obtained, and the supervisory information is converted into the initial pairwise constraint matrix $Z$, where the yellow line represents must-link and the blue line represents cannot-link. The following processes is mainly divided into three steps: (a) Enhanced SNMF, (b) Tensor low-rank representation and (c) using $Z$ to adjust $S$, where $V^*$ represents the embedding matrix obtained from step (a), $A$ is obtained from the product of $V^*$ with its transpose, and $\bm{\mathcal{C}}$ represents the tensor synthesized by similarity matrix $S$ and pairwise constraint matrix $Z$.}
\label{fig1}
\end{figure*}

We observe that when all the pairwise constraints are available, the pairwise constraint matrix is a binary block-diagonal matrix with size $n\times n$. At the same time, the ideal embedding matrix is an $n\times k$ binary matrix. Specifically, each row only has one element equaling to 1 whose column index suggests the clustering membership, while others equaling to 0. Moreover, the product of the embedding matrix with its transpose $VV^\top$ $\in \mathbb{R}^{n\times n}$ becomes a similarity matrix owing the same block-diagonal structure as the ideal pairwise constraint matrix. So we stack them into a 3-dimensional (3D) tensor and as the two slices of the tensor share the same block-diagonal structure, we could impose a global prior like tensor low-rank norm to capture the global relationship between the pairwise constraint matrix and the embedding matrix. By applying the tensor low-rank representation, more pairwise constraints can be interfered by exploring the information from the embedding matrix, and at the same time, the embedding matrix can also be promoted by the available supervisory information. Unfortunately, applying a low-rank prior on the formed tensor directly is ineffective as the similarity matrix composed by the embedding matrix is already low-rank, i.e., rank($VV^{\top}$) $\leq$ rank($V$) $\leq$ $\min(n,k)$ and $n$ is generally much greater than $k$. To this end, we propose an enhanced SNMF to make the learned embedding matrix have a larger rank to be tailored to the tensor low-rank prior. The proposed enhanced SNMF has the same input as the traditional SNMF but is more robust to initialization. After obtaining the embedding matrix by the enhanced SNMF, we apply the tensor low-rank norm on the formed 3D tensor, and then we refine the input similarity matrix by the promoted pairwise constraint matrix. These three steps are performed alternatively and iteratively. The overall framework is shown in Fig. \ref{fig1}. We compare the proposed method with 9 state-of-the-art (SOTA) methods on 9 datasets. Experiments show that our proposed model is superior to the SOTA methods.
Our work's contributions are summarized as follows:
\begin{itemize}
    \item We exploit the supervisory information from a global perspective by constructing a 3D tensor and imposing the tensor low-rank representation on it, which could concurrently promote the pairwise constraints and the input similarity matrix. 
    \item We propose an enhanced version of SNMF tailored to the tensor low-rank representation. It can produce a higher-quality and more stable embedding matrix with better clustering performance compared to SNMF.
    \item We conduct a series of experiments to prove the robustness and effectiveness of our method.
\end{itemize}

\if{}
The remainder of the paper is organized as follows. We briefly review some related works in Section \ref{Related Work}. In Section \ref{Proposed Model}, we detail the proposed method, and in Section \ref{Model Optimization}, we introduce how to optimize the method. In Section \ref{Experiments and Analysis} we carry out extensive experiments to compare the proposed method with the SOTA methods. Section \ref{Conclusion} concludes the paper. For convenience, the symbols used in the rest of this paper are summarized in Table \ref{tab:tb1}.
\fi

\begin{table}[!t]
  \centering
  \begin{threeparttable}
  \caption{List of Symbols}
  \label{tab:tb1}%
    \begin{tabularx}{\columnwidth}{lX}
    \toprule
    Symbol & Meaning \\
    \midrule
    $\text{R}^n$ & input space \\
    $\text{R}^m$ & original feature space \\
    $\text{R}^k$ & embedding feature space \\
    $\bm{\mathcal{C}}$ & tensor (bold calligraphy letter) \\
    $C$ & matrix (regular uppercase letter)\\
    $\bm{c}$  & vector (bold lowercase letter)\\
    $c$  & scalar (regular lowercase letter)\\
    $\boldsymbol{x}$ & sample \\
    $X$ & set of $n$ $\boldsymbol{x}$'s \\
    $S$ & symmetric weight matrix \\
    $V$, $V^*$ & embedding matrix \\
    $A$ & similarity among samples, i.e., $A=V^*V^{*\top}$ \\
    $Z$ & pairwise constraint matrix \\
    $\|\cdot\|_{F}$ &  Frobenius norm (the open root of the sum of the squares of all the elements of a matrix)\\
    $\|\cdot\|_{\circledast}$ & nuclear norm (the sum of the singular values of a matrix, and for a tensor, there are various different definitions. In this paper, we used the tensor nuclear norm defined in \cite{8606166}) \\
    \bottomrule
    \end{tabularx}%
\end{threeparttable}
\end{table}

\section{Related Work}\label{Related Work}
Table \ref{tab:tb1}. summarizes the symbols of this paper.
Given a sample matrix $X = \left[\bm{x}_{1}, \bm{x}_{2}, \ldots, \bm{x}_{n}\right]\in \mathbb{R}^{m \times n}$, which means there are $n$ samples in the $m$-dimensional space. $S \in \mathbb{R}^{n \times n}$ is a symmetric weight matrix, where $S_{ij}$ represents the similarity between sample $\bm x_i$ and $\bm x_j$. Typically, we can construct the weights according to the $p\mbox{-}\mathrm{NN}$ algorithm \cite{8361078}, i.e.,
\begin{equation}
\label{eq1}
 S_{ij}=\left\{\begin{array}{ll}
\exp \left(-\left\|\boldsymbol{x}_{i}-\boldsymbol{x}_{j}\right\|_{2}^{2} / \sigma^{2}\right), & \text { if }(i, j) \in \Omega \\
0, & \text {otherwise},
\end{array}\right.
\end{equation}
where $\Omega$ represents that two samples are the $p$ nearest neighbors of each other, and $\sigma$ is a hyper-parameter. In this paper, we set $\sigma$ as the average distance from a point to its $p$ nearest neighbors. We use matrix $Z$ to represent available constraints:
\begin{equation}
\label{PC matrix}
    \begin{split}
        Z_{i j} = \left\{\begin{aligned}
1, & \quad \left(\bm{x}_i, \bm{x}_j\right) \in \text{ML} \\
-1, & \quad \left(\bm{x}_i, \bm{x}_j\right) \in \text{CL} \\
0,& \quad \text{otherwise}. 
\end{aligned}\right.
    \end{split}
\end{equation}

\subsection{NMF and Symmetric NMF}
Given a non-negative matrix $X \in \mathbb{R}^{m \times n}$, NMF \cite{lee2001algorithms}, \cite{lee1999learning} approximately decomposes it into the product of two smaller non-negative matrices $H \in \mathbb{R}^{m \times r}$ and $V \in \mathbb{R}^{r \times n}$, that is, $X \approx H V$. NMF usually uses the Euclidean distance to measure the closeness between the input $X$ and its approximated decomposition, leading to the following cost function:
\begin{equation}
\label{eq3}
    \min _{H, V}\left\|X-H V\right\|_{F}^{2}, \quad \text { s.t. } H \geq 0, V \geq 0.
\end{equation}
$H$ represents the $r$-dimensional basis matrix, $V$ represents the embedding matrix (a.k.a. coefficient matrix), and $H \geq 0$ and $V \geq 0$ constrain each element in $H$ and $V$ to be non-negative, that is, only additive operations exist in NMF. In addition, NMF can be regarded as a clustering method \cite{4053063}, which is widely used in tasks such as consensus clustering \cite{4470293} and community detection \cite{9559733}.

NMF is not suitable for data with nonlinear cluster structures, while SNMF can better separate the data with nonlinear cluster structures. Unlike NMF, which takes the original data matrix as input, the input of SNMF is a similarity matrix $S \in \mathbb{R}^{n \times n}$, whose elements reflect the relationship between data points. Specifically, SNMF decomposes $S$ into the product of a non-negative matrix $V \in \mathbb{R}^{n \times r}$ and its transpose, $\text{i.e.}$,
\begin{equation}
\label{eq4}
    \min _{V}\left\|S-V V^{\top}\right\|_{F}^{2}, \quad \text { s.t. } V \geq 0.
\end{equation}
As $V$ is non-negative, the column index of the largest entry in each row of $V$ can directly indicate the cluster membership of a sample. Due to this favorable characteristic, SNMF is widely used in many clustering tasks such as signal and data analytics \cite{8653529}, biomedicine \cite{6508193}, semantic analysis of documents \cite{wang2008multi}, etc. 

\subsection{Semi-supervised SNMF}

Recently, many variants of SNMF were proposed to incorporate the supervisory information, which is known as the semi-supervised SNMF. For example, \citet{6985550} proposed GSNMF by integrating the must-links, i.e.,
\begin{equation}
\label{for_GSNMF}
    \min _{V}\left\|S-V V^{\top}\right\|_{F}^{2}+\frac{\lambda}{2} \sum_{i j}\left\|\boldsymbol{v}_{i}-\boldsymbol{v}_{j}\right\|^{2} M_{i j}, 
    \text { s.t. } V \geq 0,
\end{equation}
where $\boldsymbol{v}_{i} \in \mathbb{R}^{1 \times r}$ represents the $i$-th row of $V$ and $M_{ij}$ is defined as:
\begin{equation}
\label{eq8}
M_{i j}= \begin{cases}1, & \left(\bm x_i, \bm x_j\right) \in \text{ML} \text { or } i=j \\ 0, & \text {otherwise}.\end{cases}
\end{equation}

By minimizing Eq. (\ref{for_GSNMF}), two samples connected by an ML will have similar embeddings. To incorporate both MLs and CLs, \citet{7167693} proposed SNMFCC, i.e.,
\begin{equation}
\label{Eq_SNMFCC}
    \min _{V}\left\|S-V V^{\top}\right\|_{F}^{2}+\left\|B \odot \left(V V^{\top}-M\right)\right\|_{F}^{2}, \text { s.t. } V \geq 0,
\end{equation}
where $\odot$ means the element-wise Hadamard product. $B_{ij}$ is defined as:
\begin{equation}
    B_{i j}=\left\{\begin{array}{ll}
\alpha, & \left(\bm x_i, \bm x_j\right) \in \text{ML} \text { or } i=j \\
\beta, & \left(\bm x_i, \bm x_j\right) \in \text{CL}\\
0, & \text { otherwise, }
\end{array}\right.
\end{equation}
where $\alpha$ and $\beta$ are hyper-parameters, representing the importance of ML and CL respectively. SNMFCC effectively integrates both ML and CL by employing Eq. (\ref{Eq_SNMFCC}), ensuring that the product of the embedding matrix and its transpose reflects the pairwise constraints.

In order to ameliorate the predefined similarity matrix by the pairwise constraints, \citet{8361078} proposed PCPSNMF, whose formula is as follows:
\begin{equation}
\label{eq7}
\begin{split}
    &\min _{V, S}\left\|\tilde{S}-V V^{\top}\right\|_{F}^{2}+\mu \operatorname{Tr}(L S)+\alpha\|S-M\|_{F}^{2}, \\
&\text { s.t. } V \geq 0, \quad S \geq 0,
\end{split}
\end{equation}
where $L \in \mathbb{R}^{n \times n}$ is a graph Laplacian matrix. $\tilde{S}=\left(S+S^{\top}\right) / 2$ constructs a symmetric similarity matrix, $\mu > 0$ and $\alpha > 0$ are hyper-parameters. Eq. (\ref{eq7}) simultaneously updates the similarity matrix and the embedding matrix.

Realizing the ideal similarity matrix owns the block-diagonal structure, \citet{9543530} proposed S3NMF by adding an additional sparsity promoting term on the loss function of PCPSNMF, i.e.,
\begin{equation}
\label{eq9}
    \begin{split}
        &\min _{V, S}\left\|\tilde{S}-V V^{\top}\right\|_{F}^{2}+\mu \operatorname{Tr}(L S)+\lambda\|S\|_{1} +\frac{\beta}{2}\|S\|_{F}^{2}\\
        &\quad+\alpha \|S-M\|_{F}^{2}, \quad\text { s.t. } S \geq 0, V \geq 0,
    \end{split}
\end{equation}
where $\|\cdot\|_{1}$ is the $\ell_{1}$-norm (the sum of the absolute values of all elements of a matrix). $\beta > 0$ and $\lambda > 0$ are hyper-parameters, and $\lambda$ is used to introduce the sparsity promoting term. In order to make CLs play a greater role, \citet{9013063} proposed SANMF:
\begin{equation}
\label{eq10}
    \begin{split}
        \min _{V, D, S} &\left\|S-V V^{\top}\right\|_{F}^{2}+\eta\|D \odot S\|_{1}+\alpha \operatorname{Tr}\left(D L D^{\top}\right)\\
        &  +\beta \operatorname{Tr}\left(S L S^{\top}\right), \\
\text { s.t. }  &V \geq 0, D \geq 0, S \geq 0,\\
&D_{ij}=1, S_{ij}=0, \quad \forall\{\bm x_i, \bm x_j\} \in \text{CL}, \\
& D_{ij}=0, S_{ij}=1, \quad \forall\{\bm x_i, \bm x_j\} \in \text{ML}.
    \end{split}
\end{equation}
SANMF introduced CL propagation to distinguish CLs from unknown pairwise relations and to assist the propagation of MLs.

\subsection{Tensor-based Similarity Modeling}

Tensor denotes the multi-dimensional array, which is the extension of the matrix to a higher dimensional. Rank is one of the core concepts of matrix, but tensor rank is not well defined \cite{8606166}. Specifically, many definitions of tensor low-rank exist, such as the CP rank \cite{kolda2009tensor}, the Tucker rank \cite{kolda2009tensor}, the sum of nuclear norms \cite{6138863}, and the rank based on tensor SVD \cite{8606166}. 
Since data usually lie in a very high-dimensional space with an intrinsic low dimension, tensor low-rank representation becomes a practical approach to modeling it. The most well-known application is tensor principal component analysis \cite{8606166} that decomposes an input tensor into an intrinsic low-rank tensor and a sparse noise tensor.

Tensor rank can also indicate the similarity among its different slices because if a tensor's rank is low, its slices will be similar. This is a global similarity measure of different slices, distinct from the other local similarity measures. For this reason, tensor low-rank representation has been applied to many multiple-view clustering methods \cite{7410542,10102285}. The basic idea is to use tensor rank to measure the coherence between several similarity matrices. Jia et al. \cite{9336710} specifically designed a novel tensor low-rank norm tailored to multiple view clustering. Tensor low-rank representation has also been extended to solve the incomplete multiview clustering problem \cite{10232925,zhang2023enhanced} and the semi-supervised subspace clustering problem \cite{10007868}.

However, tensor low-rank representation has not been used to solve the semi-supervised SNMF problem as the size of the pairwise constraint matrix is $n\times n$, while that of the embedding matrix is $n\times v$, and it is difficult to construct a tensor directly. A naive approach is multiplying the embedding matrix and its transpose to get a similarity matrix and stack it with the pairwise constraint matrix to constitute the tensor. However, as the product of the embedding matrix and its transpose is already low-rank, seeking the low-rank representation for the formed tensor will not get the expected effect.

\section{Proposed Model}\label{Proposed Model}
\subsection{Motivation}
Although the existing semi-supervised SNMF methods have produced remarkable clustering performance, they all incorporate pairwise constraints from a local perspective. Specifically, they either constrain the distance between the embeddings of two samples with an ML or a CL like GSNMF and SNMFCC, or they refine the input similarity matrix element-wisely according to the pairwise constraints like PCPSNMF and SANMF. However, none of them consider the global structure of the pairwise constraint matrix. In the ideal case, the pairwise constraint matrix is a binary block-diagonal matrix if all the samples are aligned by the class membership. Moreover, the ideal similarity matrix shares the same global structure as the ideal pairwise constraint matrix. Motivated by this observation, we aim to include the pairwise constraints from the global perspective.

To this end, we propose a novel TSNMF method with the following three steps. First, we incorporate the pairwise constraints globally by tensor low-rank representation. Then, we propose an enhanced SNMF model to make SNMF tailored to the tensor low-rank representation. Finally, we refine the similarity matrix by the enhanced pairwise constraints. We alternatively repeat the above steps to enhance the pairwise constraint matrix and the similarity matrix. See Section \ref{Model Optimization} for the detailed optimization process. In the following, we will introduce the proposed method in detail.

\subsection{Incorporate the Pairwise Constraints by Tensor Low-Rank Representation}
Let $V^*$ represent the embedding matrix. The product between $V^*$ and its transpose $A=V^*V^{*\top} \in \mathbb{R}^{n \times n} $ can indicate the similarity among samples as $V^*$ is non-negative. Moreover, in the ideal case, the composed $A$ is also a binary block-diagonal matrix, i.e., $A_{ij} = 1$ if $x_i$ and $x_j$ belong to the same cluster and $A_{ij} = 0$ if $x_i$ and $x_j$ belong to different clusters \cite{6180173}. That means the ideal pairwise constraint matrix $Z$ and the ideal similarity matrix $A$ share the same block-diagonal structure. To capture this global prior, we propose to stack the matrix $A$ and the pairwise constraint matrix $Z$ into a 3D tensor and impose a tensor low-rank prior to the build tensor. By pursuing the tensor low-rank representation, the similarity matrix can be ameliorated by the available pairwise constraints while the initial pairwise is also boosted by the similarity matrix, i.e., more pairwise constraints can be inferred. Such a self-boosting strategy constantly improves both the pairwise constraint matrix and the similarity matrix, which is mathematically formulated as
\begin{equation}
\label{eq17}
    \begin{split}
        \min _{\bm{\mathcal{C}}, A, Z, E}&\|\bm{\mathcal{C}}\|_{\circledast}+\lambda\|E\|_{F}^{2}, \\
\text { s.t. } &\bm{\mathcal{C}}(:,:,1)=Z, \bm{\mathcal{C}}(:,:, 2)=A, A_{0}=A+E, \\
& Z_{i j}=1 ,  \forall\left(\bm{x}_i, \bm{x}_j\right) \in \text{ML}, \\
& Z_{i j}=-1, \forall\left(\bm{x}_i, \bm{x}_j\right) \in \text{CL},
    \end{split}
\end{equation}
where $Z \in \mathbb{R}^{n \times n}$ represents the pairwise constraint matrix, $A$ represents the refined similarity matrix, $\bm{\mathcal{C}}\in \mathbb{R}^{n \times n \times 2}$ denotes the 3D tensor stacked by $Z$ (the first slice) and $A$ (the second slice). $A_0$ represents the initial similarity matrix. For the refined similarity matrix $A$, we hope it is well aligned with pairwise constraints such that it can be improved by the supervisory information. Moreover, it should be close to the initial similarity matrix $A_0$, as the information in the $A_0$ is also valuable. Therefore, we assume that $A_0 = A + E$, where $E \in \mathbb{R}^{n \times n}$ denotes the error between $A_0$ and $A$, and we minimize the Frobenius norm of $E$. For the pairwise constraint matrix $Z$, we initialize it with available supervisory information and hope that more pairwise constraints are inferred to enhance $Z$. Finally, we apply nuclear norm $\|\cdot\|_{\circledast}$ defined in \cite{8606166} to obtain the low-rank representation. Other tensor low-rank norms are also applicable here. Note that this self-boosting strategy in Eq. (\ref{eq17}) is achieved by the global relationship between the two slices of the 3D tensor through the tensor low-rank representation, which is quite different from the previous local perspective.

However, directly using the $V^*\in \mathbb{R}^{n \times k}$ generated by SNMF cannot achieve the above target, since rank($V^* V^{*\top}$) $\leq$ rank($V^*$) $\leq$ $\min(n,k)$ and $n$ is generally much greater than $k$, the rank of $A_0$ is very small, and low-rank constraint on the tensor will lose the designed effect. Therefore, we propose an Enhanced SNMF to solve this problem.

\subsection{Enhanced SNMF}
To make the embedding of the SNMF fit the tensor low-rank representation in Eq. (\ref{eq17}), we propose an enhanced SNMF that can produce an embedding with a larger rank only using the same input as the traditional SNMF. Specifically, given a similarity matrix $S$, we first decompose it into a set of embedding matrices $\{V_i\in\mathbb{R}^{n\times k}\}_{i=1}^m$, where $m$ denotes the number of embeddings. Then, we construct a high-quality embedding matrix $V^* \in \mathbb{R}^{n \times k}$ as the final result by weighting the embedding matrices set $\left\{V_{i}\right\}_{i=1}^{m}$ with an adaptive weight vector $\bm{\alpha} \in \mathbb{R}^{m \times 1}$. The proposed enhanced SNMF incorporates the above steps into a joint optimization model, i.e.,
\begin{equation}
\label{eq18}
\begin{split}
        \min _{\bm{\alpha}, \left\{V_{i}\right\}_{i=1}^{m}, V^{*}} 
        &\sum_{i=1}^{m} \bm{\alpha}  _{i}\left\|S-V_{i} V_{i}^{\top}\right\|_{F}^{2}+\\
        &\sum_{i=1}^{m} \bm{\alpha} _{i}\left\|V_{i}-V^{*}\right\|_{F}^{2}+\beta\|\bm{\alpha}\|_{F}^{2}, \\
\text { s.t. } V_{i} \geqslant 0 ~\forall i, &V^{*} \geqslant 0, \sum_{i=1}^{m} \bm{\alpha } _{i}=1, 0 \leq \bm{\alpha } _{i} \leq 1 ~\forall i,
\end{split}
\end{equation}
where $\bm{\alpha}$ denotes the weight vector, and $\bm{\alpha}_i \geq 0$ ensures the validity of the weight value. We generate a set of embedding matrices $\left\{V_{i}\right\}_{i=1}^{m}$ from the same input similarity matrix $S$ by minimizing $\sum_{i=1}^{m} \bm{\alpha}  _{i}\left\|S-V_{i} V_{i}^{\top}\right\|_{F}^{2}$. To guarantee the diversity of $\{V_i\}_{i=1}^m$, we can simply initialize them with different initializations as SNMF is quite sensitive to the initialization \cite{9620082}. Then we obtain the final embedding $V^*$ through ${\rm min}\sum_{i=1}^{m} \bm{\alpha} _{i}\left\|V_{i}-V^{*}\right\|_{F}^{2}$ to make $V^*$ be consistent with $\{V_i\}_{i=1}^m$. Different from $V_i$ with a fixed rank, the constructed consistent embedding $V^*$ usually has a larger rank. Moreover, different initializations will lead to diverse $\{V_i\}_{i=1}^m$, but the embedding qualities of $\{V_i\}_{i=1}^m$ are also varied. To keep $V^*$ a high-quality embedding, we introduce a weight vector $\bm{\alpha}$ to measure the quality of each $V_i$. Specifically, if $V_i$ is superior in quality, the residual error is supposed to be small (i.e., $\|S-V_iV_i^T\|_F^2$ is small), and the corresponding weight $\bm{\alpha}_i$ should be large, otherwise $\bm{\alpha}_i$ should be small. We achieve this goal by $\min\sum_{i=1}^{m} \bm{\alpha}  _{i}\left\|S-V_{i} V_{i}^{\top}\right\|_{F}^{2}$. The constraints on $\bm{\alpha}$ (i.e., $\sum_{i=1}^m$ and $0\leq\bm{\alpha}_i\leq1 ~\forall i$) makes $\bm{\alpha}$ a well-defined weight vector. We also impose a regularization term on $\bm{\alpha}$ ($\|\bm{\alpha}\|_F^2$) to avoid the trivial solution that only one element of $\bm{\alpha}$ is $1$ and the remaining elements are $0$. The learned $\bm{\alpha}$ can be used to adjust the contribution of each $V_i$ to the construction of $V^*$ by minimizing $\sum_{i=1}^{m} \bm{\alpha} _{i}\left\|V_{i}-V^{*}\right\|_{F}^{2}$, i.e., a higher quality $V_i$ will contribute more in the learning of $V^*$. 

As a summary, by solving the problem in Eq. (\ref{eq18}), we get an embedding with a higher rank that is tailored to the low-rank representation problem in Eq. (\ref{eq17}). Moreover, the enhanced SNMF has the same input as the traditional SNMF but with improved robustness as it integrates different embeddings together with an adaptive weighting strategy.

\subsection{Similarity Matrix Refinement by the Enhanced Pairwise Constraints}
After solving Eq. (\ref{eq17}), we obtain an enhanced pairwise constraint matrix $Z$ and a promoted similarity matrix $A$, then we propose to further use the enhanced pairwise constraint matrix to adjust the similarity matrix promoted by Eq. (\ref{eq2}). Specifically, for a positive element in $Z$, it is likely to be an ML, we, therefore, increase the weight of the corresponding element in the similarity matrix. For the same reason, we decrease the weight of the similarity matrix for a negative element in $Z$. The adjustment strategy is formulated as
\begin{equation}
\label{eq2}
A_{ij}=\left\{\begin{array}{ll}
1-\left(1-Z_{ij}\right)\left(1-A_{ij}\right), & \text { if } Z_{ij} \geq 0 \\
\left(1+Z_{ij}\right) A_{ij} , & \text { if } Z_{ij}<0 .
\end{array}\right.
\end{equation}

\section{Optimization}\label{Model Optimization}

\subsection{Overall Optimization Method}
For better clustering performance, we first use the pairwise constraint matrix $Z$ to adjust the similarity matrix $S$ through Eq. (\ref{eq2}). Then, we input the preprocessed $S$ into Eq. (\ref{eq18}), and obtain the initial embedding matrix $V^*$ by the Enhanced SNMF. Finally, we utilize Eq. (\ref{eq17}) to simultaneously improve $S$ and the pairwise constraint matrix $Z$. The above steps are regarded as an iteration, and the optimized $S$ can be used as the input of the next iteration. Repeated iterations can make the similarity matrix and pairwise constraint matrix keep getting better. See Algorithm \ref{alg1} for details.

\begin{figure}[!t]
\renewcommand{\algorithmicrequire}{\textbf{Input:}}
\renewcommand{\algorithmicensure}{\textbf{Output:}}
  \begin{algorithm}[H]
    \caption{Proposed TSNMF}
    \label{alg1}
    \begin{algorithmic}[1]
      \Require data $X \in \mathbb{R}^{m \times n}$, pairwise constraints ML, CL, number of clusters $k$, $maxIter$, hyper-parameters: $\lambda$, $\beta$;   
      \Ensure embedding matrix $V^*$;
      \State Construct the initial similarity matrix $S$ and the pairwise constraint matrix $Z$ with Eq. (\ref{eq1}) and Eq. (\ref{PC matrix}) respectively.
      \State Initialize $\left\{V_i \in \mathbb{R}^{n \times k}\right\}_{i=1}^m$ and $V^{*} \in \mathbb{R}^{n \times k}$ with non-negative values randomly.
      \For{$t=1,\cdots,maxIter$}
        \State Adjust $S$ with $Z$ according to Eq. (\ref{eq2}).
        \State Obtain $V^*$ by Eq. (\ref{eq18}) (See Algorithm \ref{alg2} for details).
        \State $A_0=V^*V^{*\top}$.
        \State Obtain $A$ and $Z$ by Eq. (\ref{eq17}) (See Algorithm \ref{alg3} for details).
        \State Set $S=A$.
      \EndFor
      \State \Return $V^*$
    \end{algorithmic}
  \end{algorithm}
\end{figure}

\subsection{Algorithm for Solving the Enhanced SNMF in Eq. (\ref{eq18})}
Eq. (\ref{eq18}) is non-convex with multiple variables. It is difficult to find the global minimum, so we use an alternative method to find a local minimum, that is, first update $\left\{V_{i}\right\}_{i=1}^{m}$ with fixed $V^*$ and $\bm{\alpha}$, then update $V^*$ with fixed $\left\{V_{i}\right\}_{i=1}^{m}$ and $\bm{\alpha}$, and finally update $\bm{\alpha}$ with fixed $V^*$ and $\left\{V_{i}\right\}_{i=1}^{m}$. Eq. (\ref{eq18}) can be expressed as the following loss function:
\begin{equation}
\label{eq24}
\begin{split}
\mathcal{O}(\boldsymbol{\alpha}, V_{i}, V^{*}) = & \sum_{i=1}^{m}\boldsymbol{\alpha}_{i} \bigg\{ \operatorname{Tr}(SS^{\top}) - 2 \operatorname{Tr}(S V_i V_i^{\top}) \\
& + \operatorname{Tr}(V_{i} V_i^{\top} V_{i} V_i^{\top}) + \operatorname{Tr}(V_{i} V_{i}^{\top}) \\
& - 2 \operatorname{Tr}(V_{i} V^{* \top}) + \operatorname{Tr}(V^{*} V^{* \top}) \bigg\} \\
& + \beta \operatorname{Tr}(\boldsymbol{\alpha }\boldsymbol{\alpha}^{\top}), \\
\text{s.t. } V_{i} \geqslant 0 ~\forall i, & V^{*} \geqslant 0, \sum_{i=1}^{m} \bm{\alpha } _{i} = 1, 0 \leq \bm{\alpha } _{i} \leq 1 ~\forall i,
\end{split}
\end{equation}
where $\operatorname{Tr}(\cdot)$ means the trace of a matrix, i.e., the sum of the elements on the main diagonal of the matrix.

The $m$ embedding matrices are independent of each other, so we solve each $V_i$-problem separately. After removing the items irrelevant to $V_i$ in Eq. (\ref{eq24}), the Lagrange equation about $V_i$ can be expressed as follows:
\begin{equation}
\label{eq25}
\begin{split}
 \mathcal{L}_{V_i}= & -2 \boldsymbol{\alpha}_i \operatorname{Tr}\left(S V_i V_i^{\top}\right)
+\boldsymbol{\alpha}_i \operatorname{Tr}\left(V_i V_i^{\top}\right) \\
&+\boldsymbol{\alpha}_i \operatorname{Tr}\left(V_i V_i^{\top} V_i V_i^{\top}\right)-2 \boldsymbol{\alpha}_i \operatorname{Tr}\left(V_i V^{* \top}\right) \\
&+\operatorname{Tr}\left(\Phi^{\top}_i V_i\right),
\end{split}
\end{equation}
where $\Phi_i\in \mathbb{R}^{n \times k}$ is the Lagrangian multiplier matrix. Taking the first order derivative of Eq. (\ref{eq25}) with respect to $V_i$, we have
\begin{equation}
    \frac{\partial \mathcal{L}_{\boldsymbol{V}_{i}}}{\partial \boldsymbol{V}_{i}}=4 \bm{\alpha}_i V_i V_i^{\top} V_i-4 \bm{\alpha}_i S V_i-2 \bm{\alpha}_i V^*+2 \bm{\alpha}_i V_i+\Phi_i.
\end{equation}
Let $\frac{\partial \mathcal{L}_{\boldsymbol{V}_{i}}}{\partial \boldsymbol{V}_{i}}=0$, $V_i$ can be updated by
\begin{equation}
\label{eq27}
    V_{i_{n k}}^{t+1}=V_{i_{n k}}^t \sqrt[4]{\frac{\left(2 S V_i^t+V^{* t}\right)_{n k}}{\left(2 V_i^t\left(V_i^t\right)^{\top} V_i+V_i^t\right)_{n k}}},
\end{equation}
where $t$ represents the number of iterations. In the same way, the first derivative of Eq. (\ref{eq24}) with respect to $V^*$ is
\begin{equation}
    \frac{\partial\mathcal{O}}{\partial V^*}=-2 \sum_{i=1}^m \bm{\alpha}_i V_i+2 V^*.
\end{equation}
Then we update $V^*$ by $\frac{\partial \mathcal{L}_{\boldsymbol{V}^{*}}}{\partial \boldsymbol{V}^{*}}=0$, i.e.,
\begin{equation}
\label{eq29}
    V^{* t}=\sum_{i=1}^m \bm{\alpha}_i^t V_i^t.
\end{equation}

The sub-problem of Eq. (\ref{eq24}) about $\bm{\alpha}$ can be written as:
\begin{equation}
\label{eq30}
\begin{split}
&g(\bm{\alpha})=\beta\left\|\bm{\alpha}+\frac{1}{2 \beta} \bm{d}\right\|^2-\frac{1}{4 \beta} \bm{d}^2,\\
&\text { s.t. } \bm{\alpha } _{i}=1, 0 \leq \bm{\alpha } _{i} \leq 1 ~\forall i,
\end{split}
\end{equation}
where $\bm{d}\in \mathbb{R}^{1 \times m}$ and $\bm{d}_i=\left\|S-V_i V_i^{\top}\right\|_F^2+\left\|V_i-V^*\right\|_F^2$. This is a problem of computing the Euclidean projection of a point onto the capped simplex, which can be addressed by \cite{wang2015projection}. Algorithm \ref{alg2} shows the specific process of solving Eq. (\ref{eq18}).
\begin{figure}[!t]
\renewcommand{\algorithmicrequire}{\textbf{Input:}}
\renewcommand{\algorithmicensure}{\textbf{Output:}}
  \begin{algorithm}[H]
    \caption{Solution of Eq. (\ref{eq18})}
    \label{alg2}
    \begin{algorithmic}[1]
      \Require a set of random positive matrices $\left\{V_i \in \mathbb{R}^{n \times k}\right\}_{i=1}^m$ and $V^* \in \mathbb{R}^{n \times k}$, a similarity matrix $S$, the size of the embedding matrix set $m$, number of clusters $k$, $maxIter$, hyper-parameters: $\beta$;   
      \Ensure a consistent embedding matrix $V^*$;
      \State Initialize $\bm{\alpha}$ $=$ $\text{zeros}(m,1)$.
      \For{$t=1,\cdots,maxIter$}
        \For{$i=1,\cdots,m$}
        \State Updating $V_i$ with fixed $V^*$ and $\bm{\alpha}$ by Eq. (\ref{eq27}).
        \EndFor
        \State Updating $V^*$ with fixed $\left\{V_i\right\}_{i=1}^m$ and $\bm{\alpha}$ by Eq. (\ref{eq29}).
        \State Updating $\bm{\alpha}$ with fixed $V^*$ and $\left\{V_i\right\}_{i=1}^m$ by solving Eq. (\ref{eq30}).
      \EndFor
      \State Normalize each row of $V^*$ to [0,1].
      \State \Return $V^*$.
    \end{algorithmic}
  \end{algorithm}
\end{figure}

\subsection{Algorithm for Solving Eq. (\ref{eq17})}
We solve Eq. (\ref{eq17}) by the alternating direction method of multipliers (ADMM), which is very effective for solving problems with multiple variables and equality constraints \cite{chen2017note}, \cite{8962252}. We first introduce an auxiliary matrix $D \in \mathbb{R}^{n \times n}$, and let $D = Z$, then Eq. (\ref{eq17}) can be written as:
\begin{equation}
\label{eq31}
    \begin{split}
        \min _{\bm{\mathcal{C}}, A, Z, E, D}&\quad\|\bm{\mathcal{C}}\|_{\circledast}+\lambda\|E\|_{F}^{2}, \\
\text { s.t. } &\bm{\mathcal{C}}(:,:,1)=Z, \bm{\mathcal{C}}(:,:, 2)=A,A_{0}=A+E,\\
&D=Z , Z_{i j}=1 ,  \forall\left(\bm{x}_i, \bm{x}_j\right) \in \text{ML},\\
&Z_{i j}=-1, \forall\left(\bm{x}_i, \bm{x}_j\right) \in \text{CL}.
    \end{split}
\end{equation}

The augmented Lagrangian function of Eq. (\ref{eq31}) can be expressed as:
\begin{equation}
\label{eq32}
    \begin{split}
&\mathcal{L}(\boldsymbol{\mathcal{C}}, A, Z, E, D)\\
&=\|\boldsymbol{\mathcal{C}}\|_{\circledast}+\lambda\|E\|_{F}^{2}+\frac{\mu}{2}\left(\left\|A_{0}-A-E+Y_{1} / \mu\right\|_{F}^{2}\right. \\
&+\left\|Z-\boldsymbol{\mathcal{C}}(:, :, 1)+\boldsymbol{\mathcal{Y}} _{2}(:, :, 1) / \mu\right\|_{F}^{2}\\
&+\left\|A-\boldsymbol{\mathcal{C}}(:, :, 2)+\boldsymbol{\mathcal{Y}}_{2}(:, :, 2) / \mu\right\|_{F}^{2} \\
&\left.+\left\|Z-D+Y_{3} / \mu\right\|_{F}^{2}\right)+\operatorname{Tr}\left(\Psi ^{\top} A\right), \\
&\text { s.t. }  D_{i j}=1, \forall\left(\bm{x}_i, \bm{x}_j\right) \in \text{ML}, \\
&\quad D_{i j}=-1, \forall\left(\bm{x}_i, \bm{x}_j\right) \in \text{CL}.
\end{split}
\end{equation}
Among them, $Y_1 \in \mathbb{R}^{n \times n}$, $\bm{\mathcal{Y}_2} \in \mathbb{R}^{n \times n \times 2}$, and $Y_3 \in \mathbb{R}^{n \times n}$ are the Lagrangian multipliers corresponding to $A_{0}=A+E$, $\left\{\begin{array}{l}
\boldsymbol{\mathcal{C}} (: ,:, 1)=Z \\
\boldsymbol{\mathcal{C}} (:, :, 2)=A
\end{array}\right.$, and $D=Z$ respectively, $\Psi \in \mathbb{R}^{n \times n}$ is the Lagrangian multiplier matrix, and $\mu$ is the augmented Lagrangian coefficient. We also adopt an alternative iterative method to solve Eq. (\ref{eq32}). Specifically, the sub-problem with respect to $\bm{\mathcal{C}}$ can be written as follows:
\begin{equation}
\begin{split}
    \min _{\boldsymbol{\mathcal{C}}}&\quad\|\boldsymbol{\mathcal{C}}\|_{\circledast}+\frac{\mu}{2}(\left\|Z-\boldsymbol{\mathcal{C}}(:, :, 1)+\boldsymbol{\mathcal{Y}} _{2}(:, :, 1) / \mu\right\|_{F}^{2}\\
    &\quad+\left\|A-\boldsymbol{\mathcal{C}}(:, :, 2)+\boldsymbol{\mathcal{Y}}_{2}(:, :, 2) / \mu\right\|_{F}^{2} ).
\end{split}
\end{equation}
It has a closed-form solution by the tensor Singular Value Thresholding (t-SVT) operator \cite{8606166}, i.e., 
\begin{equation}
\label{eq34}
    \bm{\mathcal{C}}=\mathcal{S}_{\frac{1}{\mu}}\left(\bm{\mathcal{X}}+\bm{\mathcal{Y}_{2}}\right),
\end{equation}
where $\mathcal{S}$ is the t-SVT operator, ${\mathcal{X}}(:, :, 1)=Z$, ${\mathcal{X}}(:, :, 2)=A$.

Calculate the first-order derivative of $Z$ for Eq. (\ref{eq32}):
\begin{equation}
    \begin{split}
        \frac{\partial \mathcal{L}}{\partial Z}=4Z-2\bm{\mathcal{C}}(:,:,1)+2\bm{\mathcal{Y}}_{2}(:,:,1)/\mu-2D+2Y_3/\mu.
    \end{split}
\end{equation}
Let $\frac{\partial \mathcal{L}}{\partial Z}=0$, we get the alternate iteration formula of $Z$:
\begin{equation}
\label{eq36}
    Z=\left(\bm{\mathcal{C}}(:,: 1)-\bm{\mathcal{Y}}_{2}(:,:, 1) / \mu+D-Y_{3} / \mu\right) / 2.
\end{equation}
The first order derivative of Eq. (\ref{eq32}) with respect to $A$ is:
\begin{equation}
    \begin{split}
        \frac{\partial \mathcal{L}}{\partial A}&=4A-2A_0+2E-2\bm{\mathcal{C}}(:,:,2)\\
        &+2\bm{\mathcal{Y}}_{2}(:,:,2)/\mu-2Y_1/\mu+\Psi.
    \end{split}
\end{equation}
Since $\Psi_{ij} \times A_{ij}=0~(i, j \in\{1,2, \ldots, n\})$, we set $\frac{\partial \mathcal{L}}{\partial A}=0$ and get the update formula for $A$ as follows:
\begin{equation}
\label{eq38}
    A_{i j}=A_{i j} \frac{\left(A_{0}+Y_{1}^{+} / \mu+\bm{\mathcal{C}} ^{+}(:, :, 2)+E^{-}+\bm{\mathcal{Y}} _{2}^{-}(:, :, 2) / \mu\right) _{i j}}{\left(2 A+E^{+}+\bm{\mathcal{Y}} _{2}^{+}(:, :, 2) / \mu+Y_{1}^{-} / \mu+\bm{\mathcal{C}} ^{-}(:, :, 2)\right)_{i j}}.
\end{equation}
In Eq. (\ref{eq38}), we divide a matrix $K$ into positive and negative parts \cite{ding2010convex}, which are calculated as follows:
\begin{equation}
    K_{i j}^{+}=\left(\left|K_{i j}\right|+K_{i j}\right) / 2, \quad K_{i j}^{-}=\left(\left|K_{i j}\right|-K_{i j}\right) / 2.
\end{equation}

In the same way, after calculating the first-order derivative of $E$ in Eq. (\ref{eq32}), let the first-order derivative function be 0, and the update rule of $E$ can be obtained as follows:
\begin{equation}
\label{eq40}
    E=\left(\mu A_0+Y_1-\mu A\right) / \left(2 \lambda + \mu \right).
\end{equation}

The solution to the subproblem of $D$ is:
\begin{equation}
\label{eq41}
    D_{i j}=\left\{\begin{array}{ll}
1, & \forall\left(\bm{x}_i, \bm{x}_j\right) \in \text{ML} \\
-1, & \forall\left(\bm{x}_i, \bm{x}_j\right) \in \text{CL} \\
Z_{i j}+Y_{3 i j} / \mu, & \text { otherwise. }
\end{array}\right.
\end{equation}

In addition, the updated rules of the Lagrangian multipliers and augmented Lagrangian coefficient are as follows \cite{liu2013robust}:
\begin{equation}
\label{eq42}
    \begin{split}
        \left\{\begin{array}{l}
Y_{1}{ }^{(t+1)}=Y_{1}^{(t)}+\mu^{(t)}\left(A_{0}-A^{(t+1)}-E^{(t+1)}\right) \\
\bm{\mathcal{Y}} _{2}^{(t+1)}(:, :, 1)=\bm{\mathcal{Y}}_{2}^{(t)}(:, :, 1)+\mu^{(t)}\left(Z^{(t+1)}-\bm{\mathcal{C}}^{(t+1)}(:, :, 1)\right) \\
\bm{\mathcal{Y}}_{2}^{(t+1)}(:, :, 2)=\bm{\mathcal{Y}}_{2}^{(t)}(:, :, 2)+\mu^{(t)}\left(A^{(t+1)}-\bm{\mathcal{C}} ^{(t+1)}(:, :, 2)\right) \\
Y_{3}{ }^{(t+1)}=Y_{3}^{(t+1)}+Z^{(t+1)}-D^{(t+1)} \\
\mu^{(t+1)}=\min \left(\rho \mu^{(t)} ; \mu_{\max }\right),
\end{array}\right.
    \end{split}
\end{equation}
where $t$ represents the number of iterations and $\mu_{\rm max}$ is a predefined upper bound for $\mu$. For the specific solution process of Eq. (\ref{eq17}), see Algorithm \ref{alg3}.
\begin{figure}[!t]
\renewcommand{\algorithmicrequire}{\textbf{Input:}}
\renewcommand{\algorithmicensure}{\textbf{Output:}}
  \begin{algorithm}[H]
    \caption{Solve Eq. (\ref{eq17}) by ADMM}
    \label{alg3}
    \begin{algorithmic}[1]
      \Require the initial similarity matrix $A_0$, pairwise constraints ML, CL, $maxIter$, hyper-parameters: $\lambda$;   
      \Ensure the final similarity matrix $A$, the final pairwise constraint matrix $Z$;
      \State Initialize $\bm{\mathcal{C}}=\bm{\mathcal{Y}}_2=0\in \mathbb{R}^{n\times n \times 2}$, $A=Z=E=D=Y_1=Y_3=0\in \mathbb{R}^{n \times n}$, $\rho=1.1$, $\mu =1e-3$, $\mu _{max}=1e10$.
      \For{$t=1,\cdots,maxIter$}
        \State Updating $\bm{\mathcal{C}}$ using Eq. (\ref{eq34})
        \State Updating $Z$ using Eq. (\ref{eq36}).
        \State Updating $A$ using Eq. (\ref{eq38}).
        \State Updating $E$ using Eq. (\ref{eq40}).
        \State Updating $D$ using Eq. (\ref{eq41}).
        \State Updating the Lagrangian multipliers and augmented Lagrangian coefficient $Y_1$, $\bm{\mathcal{Y}}_2$, $Y_3$ and $\mu$ using Eq. (\ref{eq42}).
      \EndFor
    \end{algorithmic}
  \end{algorithm}
\end{figure}

\subsection{Computational Complexity}
The computational complexity of Algorithm \ref{alg1} is mainly determined by steps 5 and 7. Step 5 corresponds to Algorithm \ref{alg2}, whose main computational complexity lies in the updates of $\left\{V_i\right\}_{i=1}^m$, $V^*$ and $\bm{\alpha}$, i.e., $\mathrm{O}\left(n^{2} k m\right)$, $\mathrm{O}\left(n k m\right)$ and $\mathrm{O}\left(m^{2}\right)$, respectively. As $m$ is much smaller than $n$, so the computational complexity of Algorithm \ref{alg2} in one iteration is $\mathrm{O}\left(n^{2} k m\right)$. Besides, the computational complexity of step 7, namely Algorithm \ref{alg3}, mainly comes from steps 3-5 in Algorithm \ref{alg3}. Specifically, the solution of $\bm{\mathcal{C}}$ uses the t-SVD of an $n \times n \times 2$ tensor and its computational complexity is $\mathrm{O}\left(n^{3}\right)$ \cite{8606166}. The updates of $A$ and $Z$ include matrix addition, subtraction, and dot division operations with the complexity of $\mathrm{O}\left(n^{2}\right)$. So the computational complexity of each iteration of Algorithm \ref{alg3} is $\mathrm{O}\left(n^{3}\right)$. For general data sets, $km < n$, so Algorithm \ref{alg3} has higher computational complexity than Algorithm \ref{alg2}. Assuming that $T$ is the number of iterations of Algorithm \ref{alg3}, the complexity of each iteration of Algorithm \ref{alg1} is $\mathrm{O}\left(n^{3} T\right)$. Empirically, Algorithm \ref{alg1} can be stopped in 3 iterations. See the detailed experimental result and analysis in Section \ref{IterAlg1}.

\section{Experiments and Analysis}\label{Experiments and Analysis}
\subsection{Experimental Settings}
To demonstrate the effectiveness of our method, we compared it with the following 9 SOTA methods.
\begin{itemize}
    \item NMF \cite{lee2001algorithms} obtains the lower-rank approximation by decomposing the data matrix, and the clustering result is obtained by performing $K$-means \cite{hartigan1979algorithm} on the embedding matrix.
    \item SNMF \cite{kuang2012symmetric} decomposes a similarity matrix as the product of an embedding matrix and its transpose, where the embedding matrix acts as the clustering indicator.
    \item GNMF \cite{5674058} incorporates the geometrical information of the data matrix on the basis of NMF.
    \item GSNMF \cite{6985550} extends SNMF by adding ML supervisory information by a Laplacian graph regularization.
    \item SNMFCC \cite{7167693} uses the pairwise constraints to regularize the product of the embedding matrix and its transpose.
    \item PCPSNMF \cite{8361078} simultaneously updates the similarity matrix and the embedding matrix according to the pairwise constraints.
    \item S3NMF \cite{9543530} uses the block-diagonal structure prior to achieving semi-supervised SNMF.
    \item SANMF \cite{9013063} adopts the adversarial pairwise constraint propagation to construct a similarity matrix for SNMF.
    \item MVCHSS \cite{10076472} is a semi-supervised SNMF model for multiview data. In the experiment, we fixed the number of its views to 1.
\end{itemize}

\begin{table}[!t]
  \centering
  \begin{threeparttable}
  \caption{Dataset Characteristics}
  \label{tab:tb2}%
    \begin{tabular}{ccccc}
    \hline\hline
    Dataset & $n$ & $m$ & $k$ & Data Split \\
    \hline
    Libras & 360   & 90    & 15    & 7,8,9,10,11,12,13,14,15 \\
    Yale & 165  & 1024  & 15    & 7,8,9,10,11,12,13,14,15 \\
    ORL   & 400   & 1024  & 40    & 8,12,16,20,24,28,32,36,40 \\
    Leaf  & 340   & 14  & 30    & 14,16,18,20,22,24,26,28,30 \\
    UMIST & 575   & 644   & 20    & 12,13,14,15,16,17,18,19,20 \\
    BinAlpha & 1404   & 320   & 36    & 12,15,18,21,24,27,30,33,36 \\
    \hline\hline
    \end{tabular}%
    \begin{tablenotes}
    \item $n$ is the number of samples, $m$ is the dimensionality of each sample, $k$ is the number of categories.
    \end{tablenotes}
\end{threeparttable}
\end{table}%
We evaluated all the methods on 6 datasets covering face images, digit images, hand movements, and so on. 
Table \ref{tab:tb2} summarizes the specific information of the 6 datasets.

For a fair comparison, the similarity matrices of all methods were generated using the $p\mbox{-}\mathrm{NN}$ graph \cite{8361078}, where $p$ was empirically set to $\left\lfloor\log _{2}(n)\right\rfloor$ ($\left\lfloor \cdot \right\rfloor$ returns the largest integer less than the original value), and in order to ensure the symmetry of the similarity matrix, $W=\left(W+W^{\top}\right) / 2$ was used as post-processing. Since the clustering performance of MVCHSS is greatly affected by $p$, in order to achieve the best performance, we set $p=5$ in MVCHSS, which is suggested by its original paper. We tuned the hyper-parameters of different methods according to the scope of the original literature and performed each method 10 times and reported the average performance. Further, we set the maximum number of iterations for each method to be 500, and 10$\%$ of the labels are randomly selected as supervisory information. The proposed method has three hyper-parameters $\beta$, $\lambda$, and $m$. We fixed $m=10$, and $\beta$ and $\lambda$ were adjusted in the ranges of $\{0.1,1,5,10\}$ and $\{0.01, 0.03, 0.1, 0.3, 1\}$.

We used two metrics to measure the quality of the clustering results, namely ACC and NMI \cite{8361078}, which represent clustering accuracy and normalized mutual information respectively. Their value ranges are both in [0,1] and the larger value indicates the better the clustering performance.

\begin{table*}[!t]
\scriptsize

  \centering
  \begin{threeparttable}
  \caption{Clustering Performance on all the Datasets}
    \begin{tabularx}{\textwidth}{@{\hspace{5.8pt}}c@{\hspace{5.8pt}}c@{\hspace{5.8pt}}c@{\hspace{5.8pt}}c@{\hspace{5.8pt}}c@{\hspace{5.8pt}}c@{\hspace{5.8pt}}c@{\hspace{5.8pt}}c@{\hspace{5.8pt}}c@{\hspace{5.8pt}}c@{\hspace{5.8pt}}c@{\hspace{5.8pt}}}
    \toprule
    \toprule
    \textbf{ACC} & \textbf{NMF\cite{lee2001algorithms}} & \textbf{SNMF\cite{kuang2012symmetric}} & \textbf{GNMF\cite{5674058}} & \textbf{GSNMF\cite{6985550}} & \textbf{SNMFCC\cite{7167693}} & \textbf{PCPSNMF\cite{8361078}} & \textbf{S3NMF\cite{9543530}} & \textbf{SANMF\cite{9013063}} & \textbf{MVCHSS\cite{10076472}} & \textbf{TSNMF} \\
    \hdashline
    BinAlpha & 0.183±0.010 $\bullet$ & 0.425±0.016 $\bullet$ & 0.409±0.013 $\bullet$ & 0.913±0.028 $\bullet$ & 0.535±0.029 $\bullet$ & 0.941±0.033 $\bullet$ & 0.932±0.027 $\bullet$ & 0.938±0.022 $\bullet$ & \underline{0.945±0.025} $\bullet$ & \textbf{1.000±0.000} \\
    Leaf & 0.311±0.015 $\bullet$ & 0.506±0.019 $\bullet$ & 0.399±0.016 $\bullet$ & 0.637±0.014 $\bullet$ & 0.515±0.025 $\bullet$ & \underline{0.762±0.038} $\bullet$ & 0.750±0.028 $\bullet$ & 0.754±0.160 $\bullet$ & 0.583±0.022 $\bullet$ & \textbf{0.874±0.042} \\
    Libras & 0.352±0.038 $\bullet$ & 0.487±0.016 $\bullet$ & 0.462±0.025 $\bullet$ & 0.774±0.066 $\bullet$ & 0.523±0.045 $\bullet$ & \underline{0.911±0.022} $\bullet$ & 0.908±0.050 $\bullet$ & 0.846±0.074 $\bullet$ & 0.745±0.050 $\bullet$ & \textbf{1.000±0.000} \\
    ORL & 0.341±0.020 $\bullet$ & 0.632±0.013 $\bullet$ & 0.417±0.013 $\bullet$ & 0.791±0.031 $\bullet$ & 0.623±0.026 $\bullet$ & \underline{0.805±0.021} $\bullet$ & 0.784±0.015 $\bullet$ & 0.773±0.023 $\bullet$ & 0.775±0.017 $\bullet$ & \textbf{0.898±0.020} \\
    UMIST & 0.349±0.028 $\bullet$ & 0.514±0.022 $\bullet$ & 0.456±0.031 $\bullet$ & \underline{0.920±0.054} $\bullet$ & 0.612±0.043 $\bullet$ & 0.795±0.071 $\bullet$ & 0.788±0.041 $\bullet$ & 0.793±0.071 $\bullet$ & 0.864±0.017 $\bullet$ & \textbf{0.963±0.036} \\
    Yale & 0.375±0.028 $\bullet$ & 0.499±0.030 $\bullet$ & 0.402±0.022 $\bullet$ & 0.642±0.039 $\bullet$ & 0.509±0.013 $\bullet$ & 0.769±0.092 $\bullet$ & \underline{0.773±0.063} $\bullet$ & 0.728±0.083 $\bullet$ & 0.607±0.032 $\bullet$ & \textbf{0.850±0.048} \\
    \midrule
    \textbf{NMI} & \textbf{NMF\cite{lee2001algorithms}} & \textbf{SNMF\cite{kuang2012symmetric}} & \textbf{GNMF\cite{5674058}} & \textbf{GSNMF\cite{6985550}} & \textbf{SNMFCC\cite{7167693}} & \textbf{PCPSNMF\cite{8361078}} & \textbf{S3NMF\cite{9543530}} & \textbf{SANMF\cite{9013063}} & \textbf{MVCHSS\cite{10076472}} & \textbf{TSNMF} \\
    \hdashline
    BinAlpha & 0.313±0.013 $\bullet$ & 0.582±0.008 $\bullet$ & 0.557±0.009 $\bullet$ & 0.959±0.010 $\bullet$ & 0.645±0.017 $\bullet$ & \underline{0.981±0.010} $\bullet$ & 0.978±0.010 $\bullet$ & 0.980±0.006 $\bullet$ & 0.967±0.012 $\bullet$ & \textbf{1.000±0.000} \\
    Leaf & 0.515±0.012 $\bullet$ & 0.691±0.008 $\bullet$ & 0.601±0.017 $\bullet$ & 0.775±0.007 $\bullet$ & 0.695±0.012 $\bullet$ & \underline{0.866±0.020} $\bullet$ & 0.861±0.014 $\bullet$ & 0.859±0.131 $\bullet$ & 0.738±0.011 $\bullet$ & \textbf{0.919±0.018} \\
    Libras & 0.415±0.030 $\bullet$ & 0.624±0.012 $\bullet$ & 0.592±0.020 $\bullet$ & 0.853±0.030 $\bullet$ & 0.641±0.030 $\bullet$ & 0.961±0.019 $\bullet$ & \underline{0.962±0.018} $\bullet$ & 0.925±0.038 $\bullet$ & 0.794±0.032 $\bullet$ & \textbf{1.000±0.000} \\
    ORL & 0.576±0.016 $\bullet$ & 0.779±0.007 $\bullet$ & 0.642±0.013 $\bullet$ & 0.880±0.019 $\bullet$ & 0.776±0.013 $\bullet$ & \underline{0.880±0.011} $\bullet$ & 0.874±0.011 $\bullet$ & 0.874±0.013 $\bullet$ & 0.867±0.009 $\bullet$ & \textbf{0.939±0.010} \\
     UMIST & 0.498±0.026 $\bullet$ & 0.701±0.020 $\bullet$ & 0.627±0.029 $\bullet$ & \underline{0.957±0.022} $\bullet$& 0.766±0.018 $\bullet$ & 0.918±0.029 $\bullet$ & 0.909±0.018 $\bullet$ & 0.877±0.029 $\bullet$ & 0.915±0.007 $\bullet$ & \textbf{0.977±0.018} \\
    Yale & 0.442±0.034 $\bullet$ & 0.541±0.013 $\bullet$ & 0.472±0.018 $\bullet$ & 0.678±0.029 $\bullet$ & 0.538±0.008 $\bullet$ & 0.836±0.062 $\bullet$ & \underline{0.838±0.046} $\bullet$ & 0.796±0.058 $\bullet$ & 0.642±0.029 $\bullet$ & \textbf{0.874±0.030} \\
    \bottomrule
    \bottomrule
    \end{tabularx}%
  \begin{tablenotes}
    \item The best ACC/NMI in each dataset is presented in bold, while the second-best one is underlined. $\bullet$/$\circ$ indicates whether TSNMF is significantly better than the compared algorithm according to pairwise $t$-test at the significance level of 0.05.
    \end{tablenotes}
  \label{tab:complete}%
  \end{threeparttable}
\end{table*}%

\subsection{Comparisons of Clustering Results}
Table \ref{tab:complete} lists the average clustering performance and standard deviation of all methods over 10 repetitions on 6 datasets. According to it, we can draw the following conclusions:

\begin{figure*}[!t] 
\centering
\subfigure[]{
\label{ACC_0.1_BinAlpha}
\includegraphics[width=2.2in, height=1.7in]{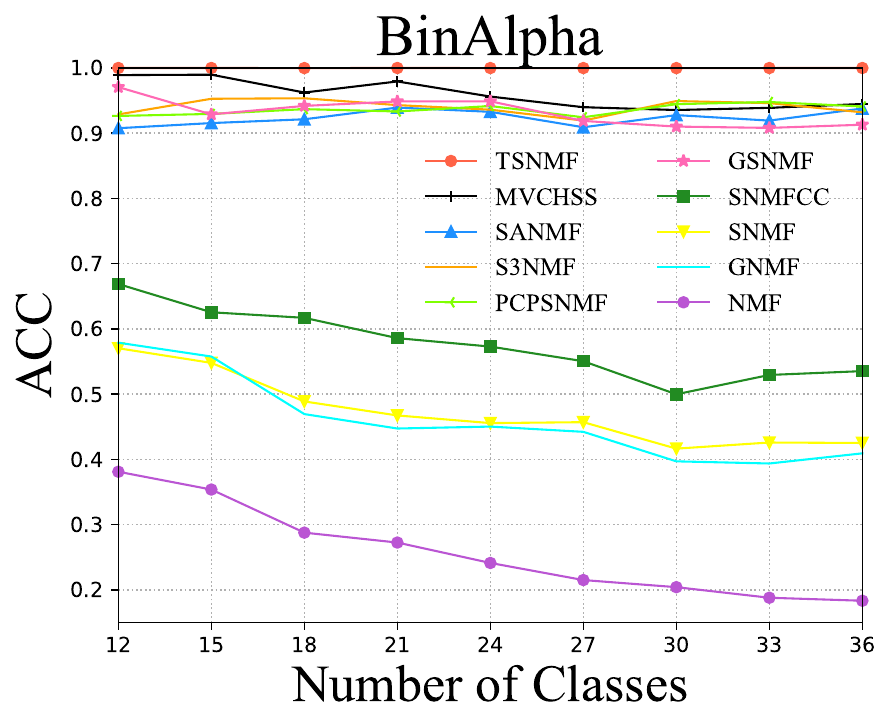}
}
\subfigure[]{
\label{ACC_0.1_Leaf}
\includegraphics[width=2.2in, height=1.7in]{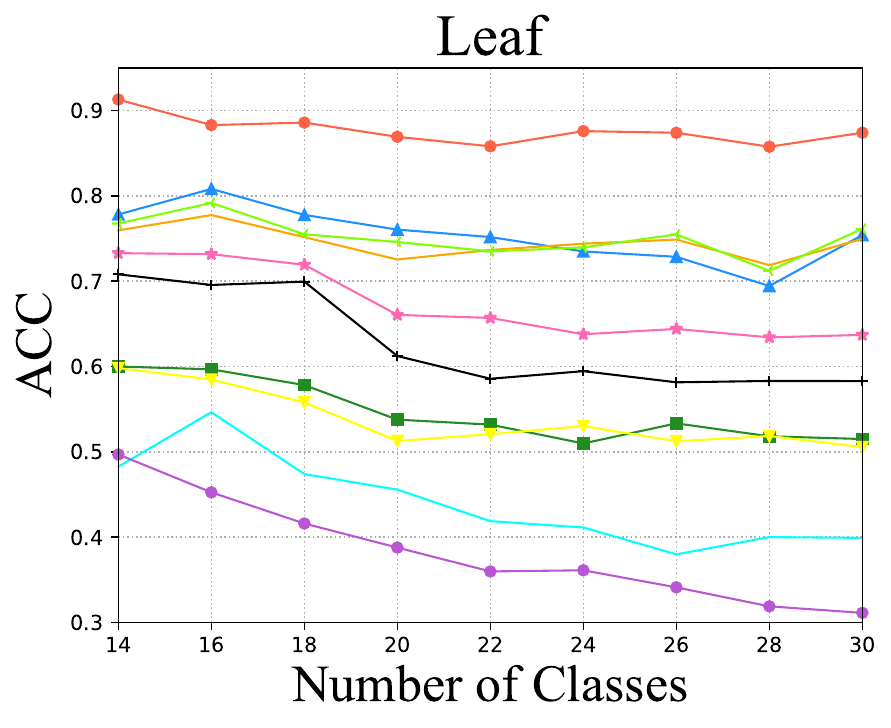}
}
\subfigure[]{
\label{ACC_0.1_Libras}
\includegraphics[width=2.2in, height=1.7in]{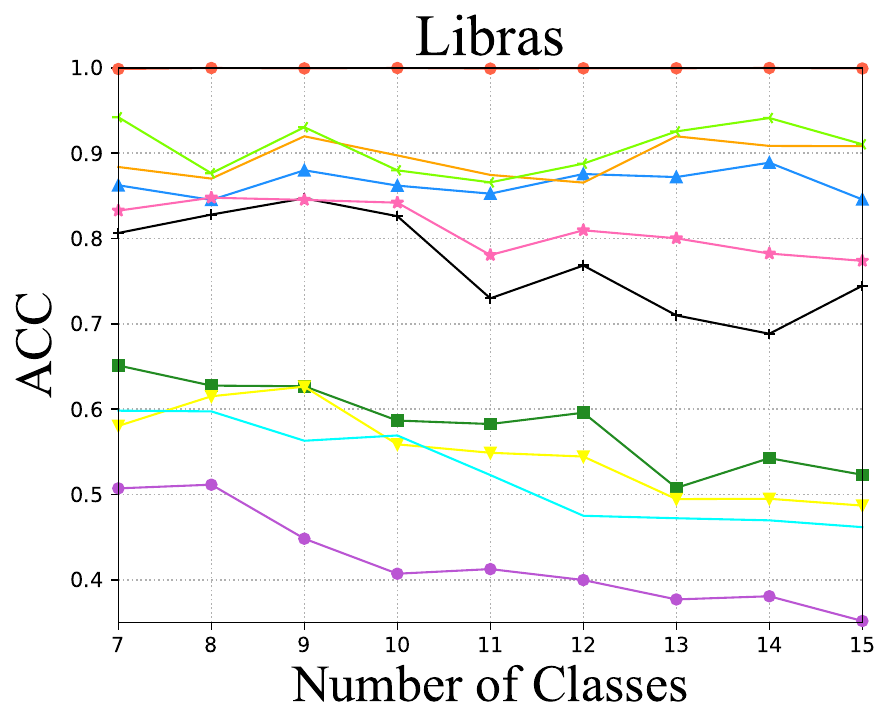}
}
\\
\subfigure[]{
\label{ACC_0.1_ORL} 
\includegraphics[width=2.2in, height=1.7in]{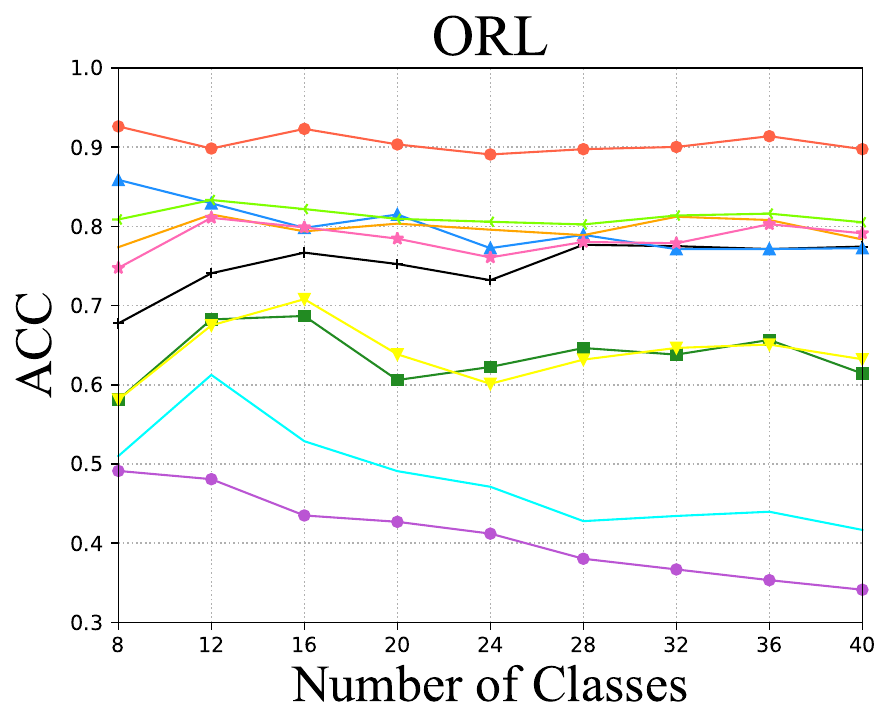} 
}
\subfigure[]{
\label{ACC_0.1_UMIST}
\includegraphics[width=2.2in, height=1.7in]{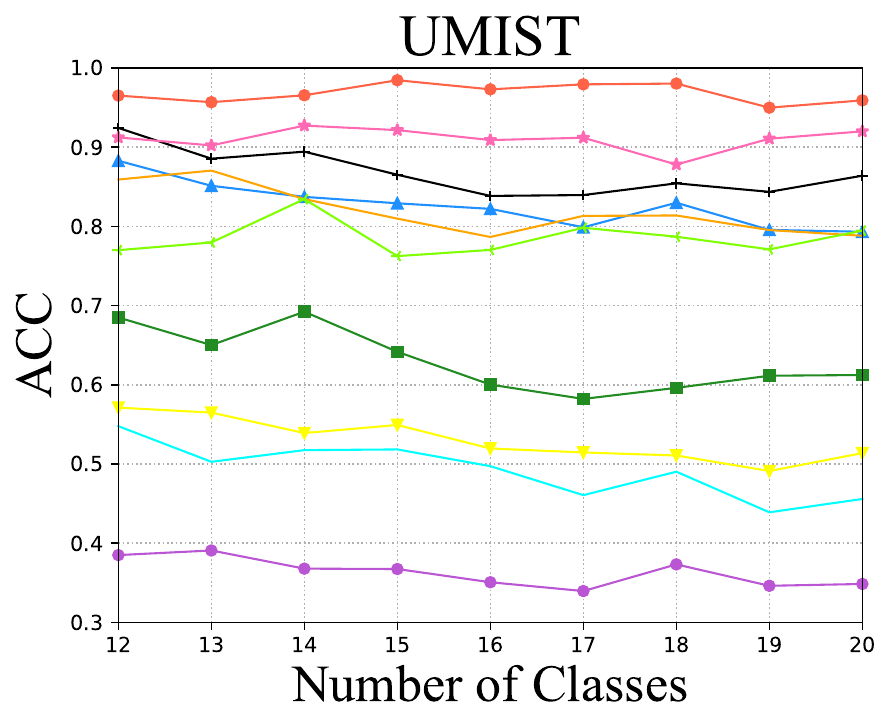}
}
\subfigure[]{
\label{ACC_0.1_Yale} 
\includegraphics[width=2.2in, height=1.7in]{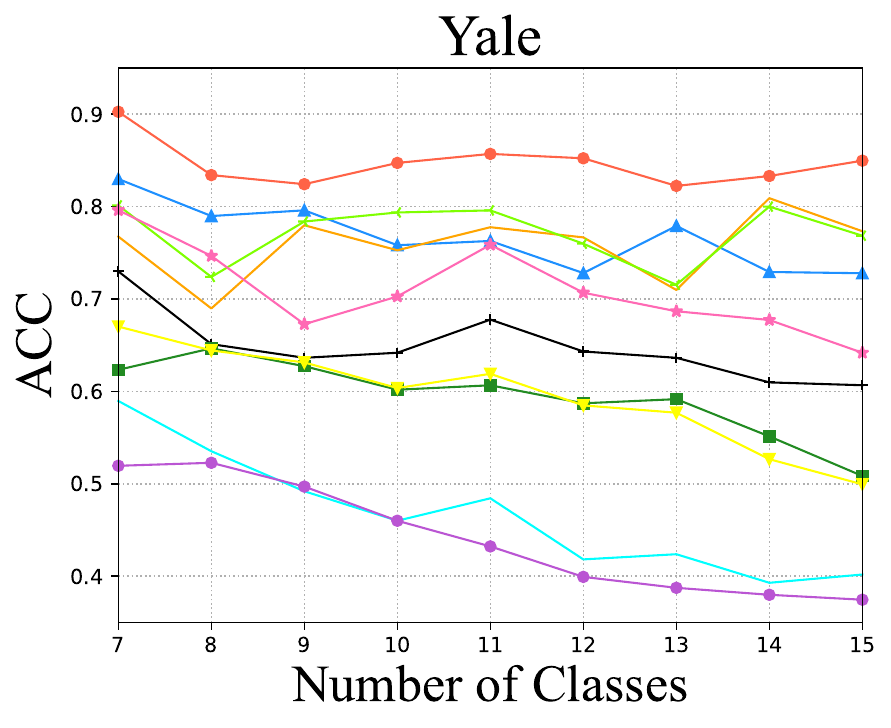} 
}
\caption{Comparisons of the accuracy of different methods on 6 datasets with different numbers of classes. The legends of the 6 subfigures are consistent and shown in \subref{ACC_0.1_BinAlpha}.}
\label{fig2}
\end{figure*}

\begin{figure*}[!t] 
\centering
\subfigure[]{
\includegraphics[width=2.2in, height=1.7in]{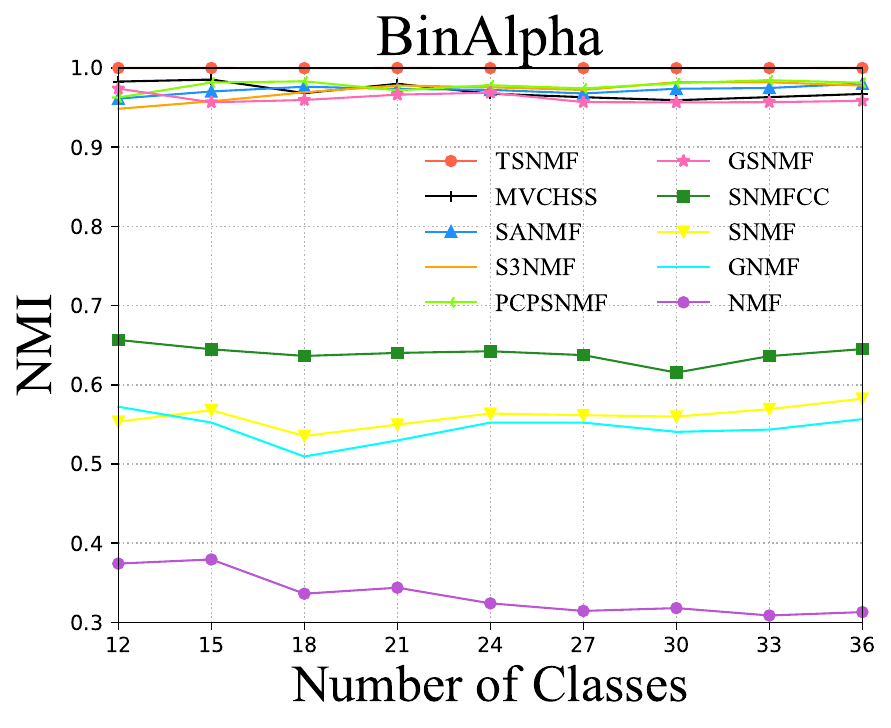}
\label{NMI_0.1_BinAlpha}
}
\subfigure[]{
\includegraphics[width=2.2in, height=1.7in]{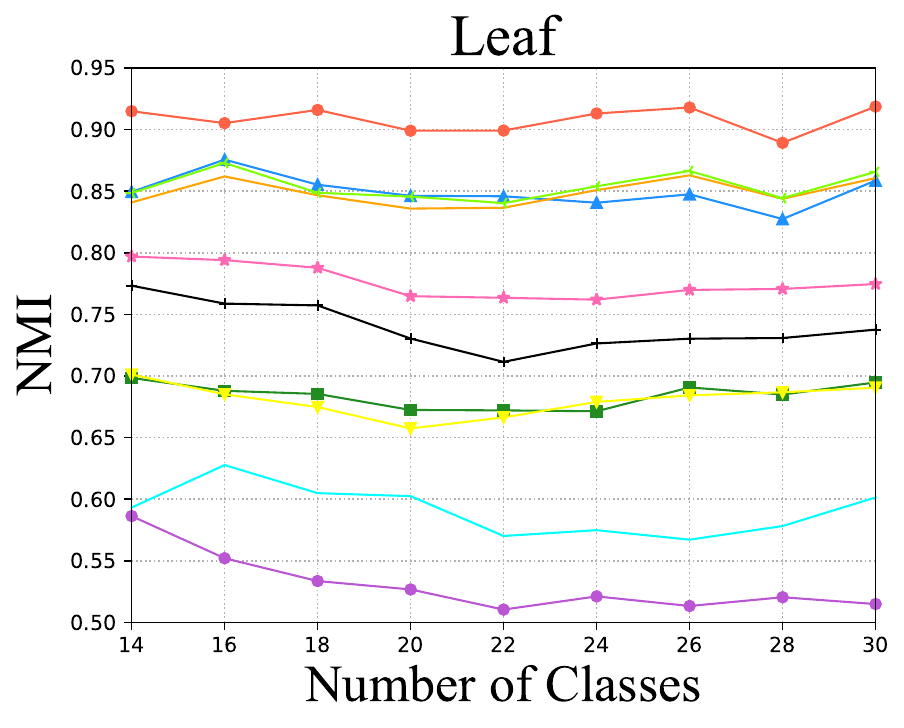}
\label{NMI_0.1_Leaf}
}
\subfigure[]{
\includegraphics[width=2.2in, height=1.7in]{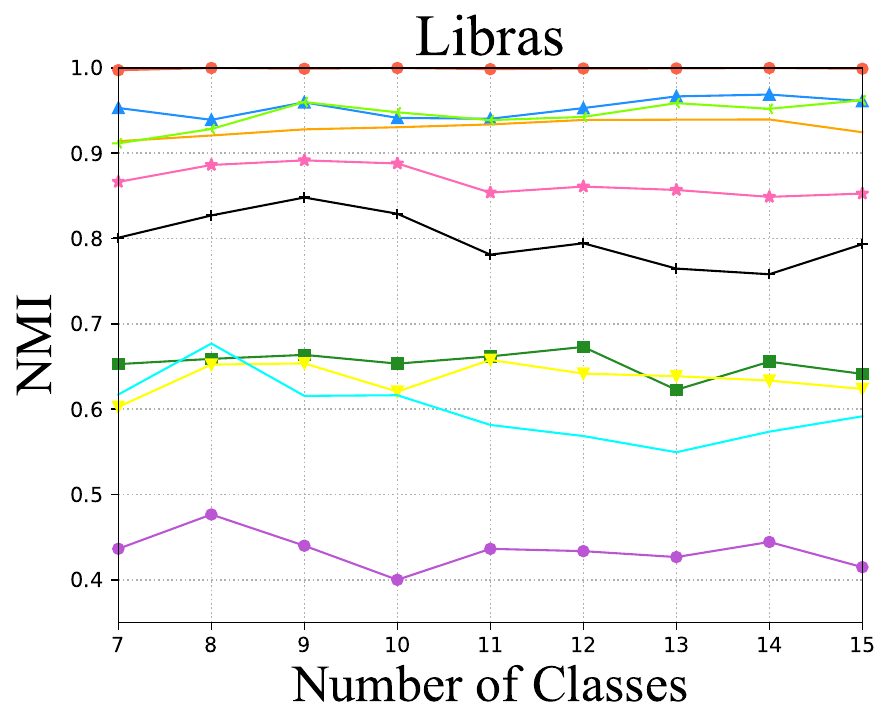}
\label{NMI_0.1_Libras}
}
\quad

\subfigure[]{
\includegraphics[width=2.2in, height=1.7in]{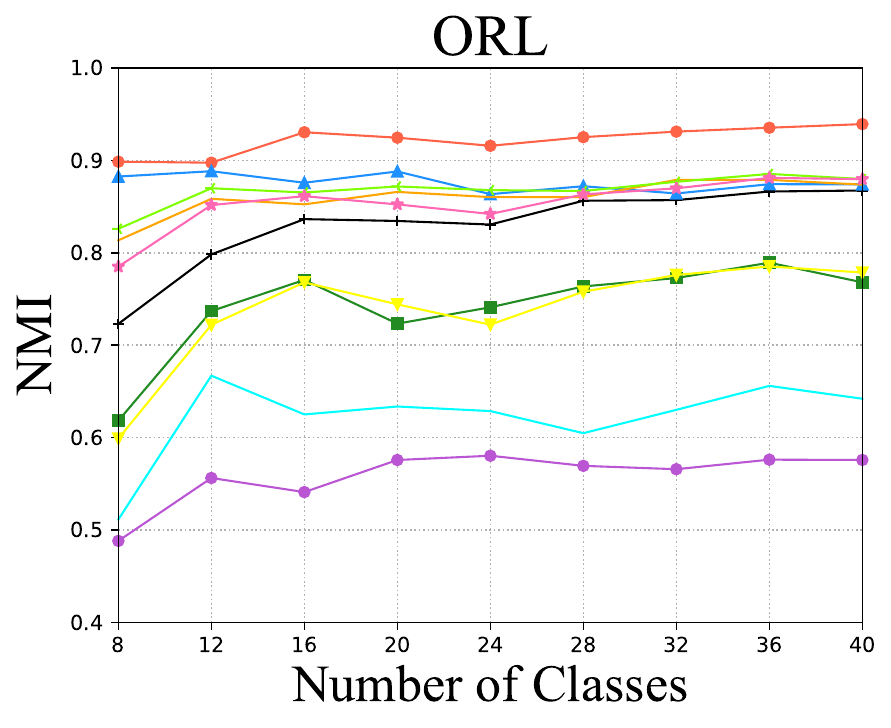} 
\label{NMI_0.1_ORL} 
}
\subfigure[]{
\includegraphics[width=2.2in, height=1.7in]{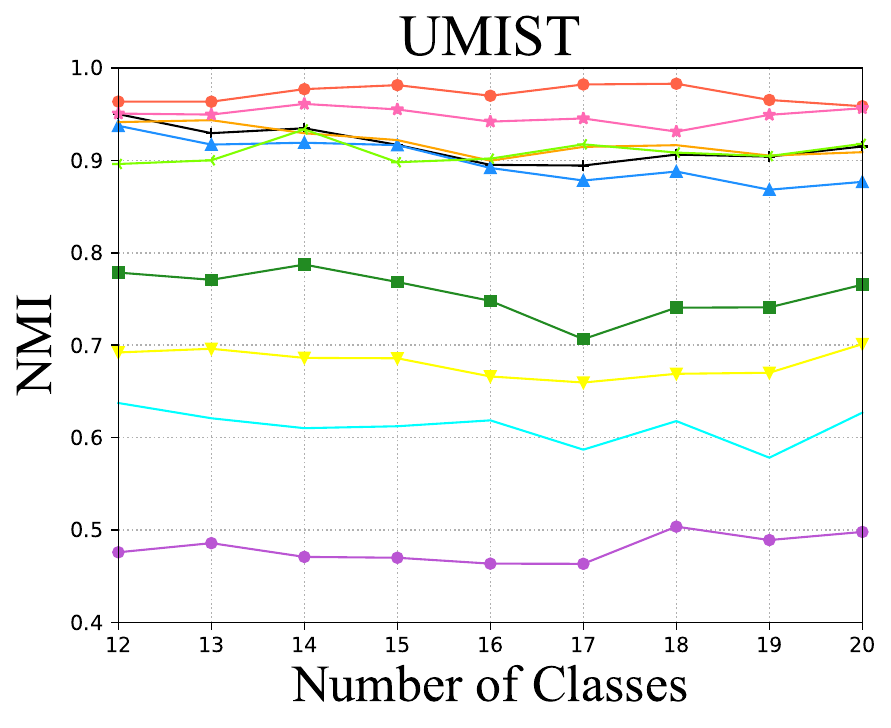}
\label{NMI_0.1_UMIST}
}
\subfigure[]{
\includegraphics[width=2.2in, height=1.7in]{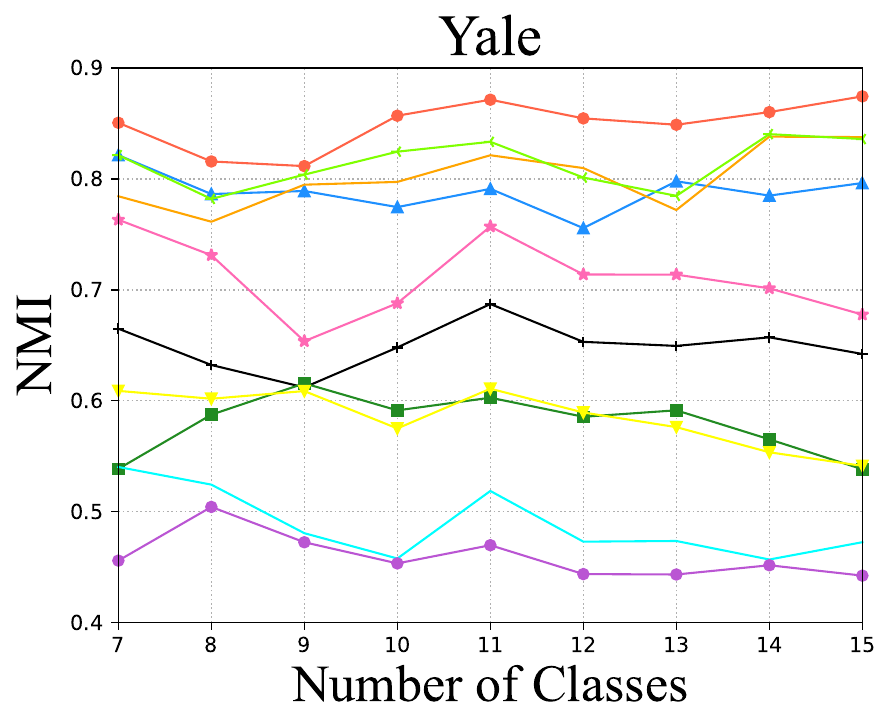} 
\label{NMI_0.1_Yale} 
}
\caption{Comparisons of the NMI of different methods on 6 datasets with different number of classes. The legends of the 6 subfigures are consistent and shown in \subref{NMI_0.1_BinAlpha}.}
\label{fig3}
\end{figure*}

\begin{itemize}
    \item The proposed algorithm demonstrates significant improvement in clustering performance on 6 datasets compared to the previous methods. Specifically, the average ACC and NMI of TSNMF is higher than all the SOTA methods on all six datasets. According to the pairwise t-test, in all 108 cases, the improvements are significant. For example, on ORL, TSNMF improves the ACC from 0.805 to 0.898 compared with the second-best one. Those observations prove the effectiveness and robustness of TSNMF.
    \item The clustering performances of both GSNMF and SNMFCC are better than SNMF, proving the importance of introducing supervisory information.
    \item PCPSNMF, S3NMF, SANMF and MVCHSS perform better than SNMFCC in most cases. It may be because SNMFCC only directly utilizes the supervisory information without propagating known pairwise constraints to unknown ones, thus limiting the role of supervisory information. Besides, PCPSNMF, S3NMF and SANMF iteratively strengthen the similarity matrix in the optimization process, while SNMFCC uses a fixed predefined similarity matrix.
    \item The reasons why TSNMF is superior to the currently SNMF-based semi-supervised algorithms may be as follows: First of all, TSNMF utilizes the low-rank constraints on tensor to realize the global propagation of supervisory information and maximize the role of supervisory information. Second, the proposed enhanced SNMF improves the quality of the embedding matrix and similarity matrix to a certain extent.
\end{itemize}

\subsection{Influence of the Number of Categories}
In order to evaluate the influence of the number of categories of the datasets on the proposed method, we selected different subsets containing different numbers of categories for each dataset to conduct experiments. In particular, we selected the top $k$ categories of each dataset, and the setting of $k$ is shown in Table \ref{tab:tb2}. We randomly selected 10$\%$ of the ground-truth labels as the supervisory information. We repeated each method 10 times with different supervisory information and took the mean value as the final result. The results are shown in Figs. \ref{fig2} and \ref{fig3}, where we can see that as the number of categories increases, the clustering task becomes more difficult, and the accuracy of all algorithms shows a slight downward trend on some datasets. In addition, TSNMF consistently achieves the best clustering performance with the different number of categories, which further proves the superiority of our algorithm.

\subsection{Influence of the Amount of Supervisory Information}
\begin{figure*}[!t] 
\centering
\begin{center}
\subfigure[]{
\includegraphics[width=1.09in, height=0.85in]{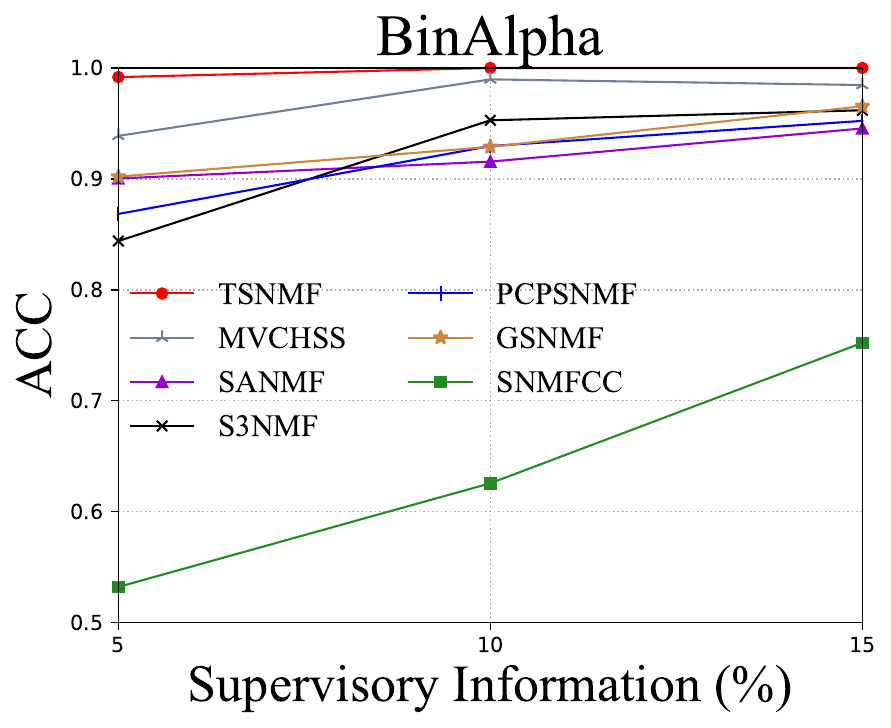}
\label{ACC_label_BinAlpha}
}\hspace{-1mm}
\subfigure[]{
\includegraphics[width=1.09in, height=0.85in]{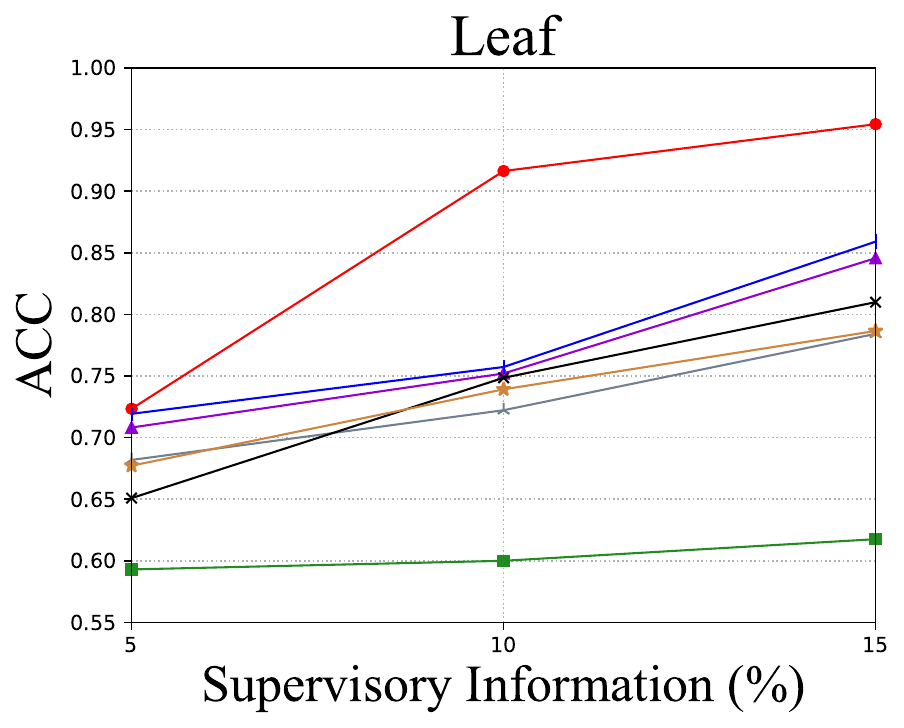}
\label{ACC_label_Leaf}
}\hspace{-1mm}
\subfigure[]{
\includegraphics[width=1.09in, height=0.85in]{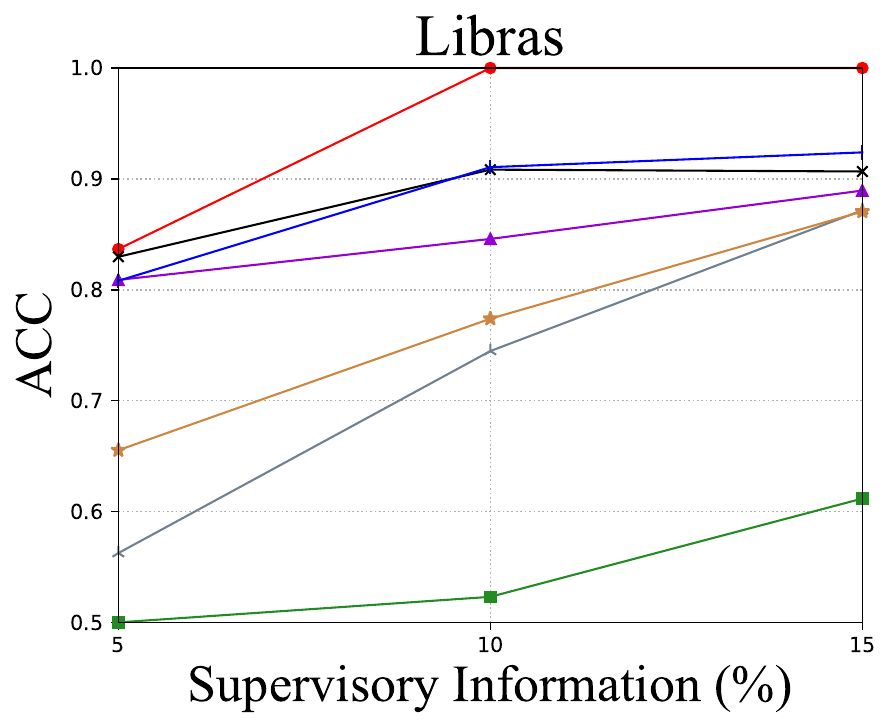}
\label{ACC_label_Libras}
}\hspace{-1mm}
\subfigure[]{
\includegraphics[width=1.09in, height=0.85in]{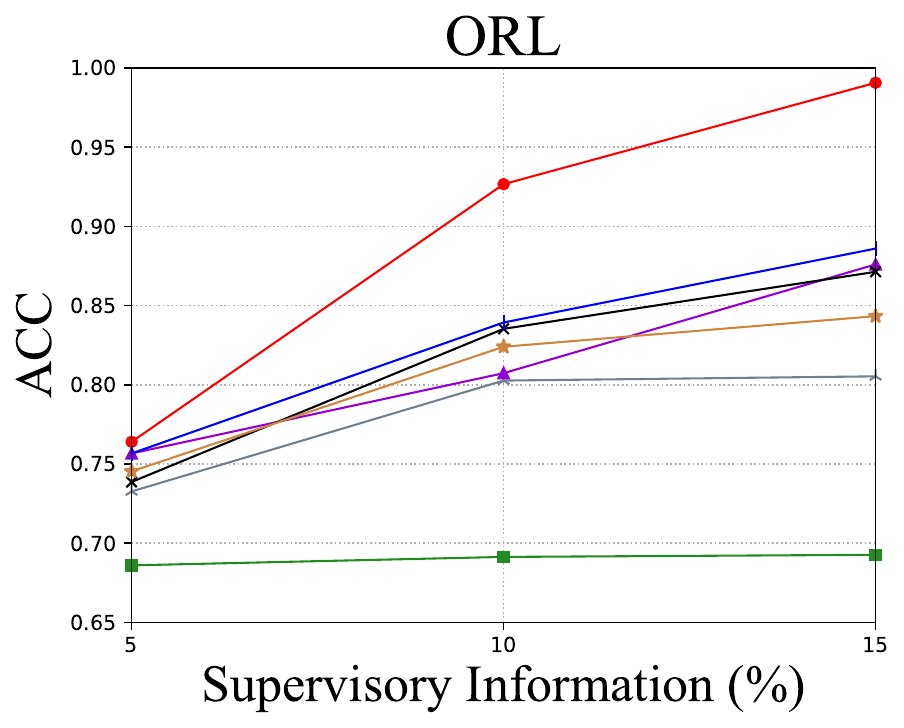} 
\label{ACC_label_ORL} 
}\hspace{-1mm}
\subfigure[]{
\includegraphics[width=1.09in, height=0.85in]{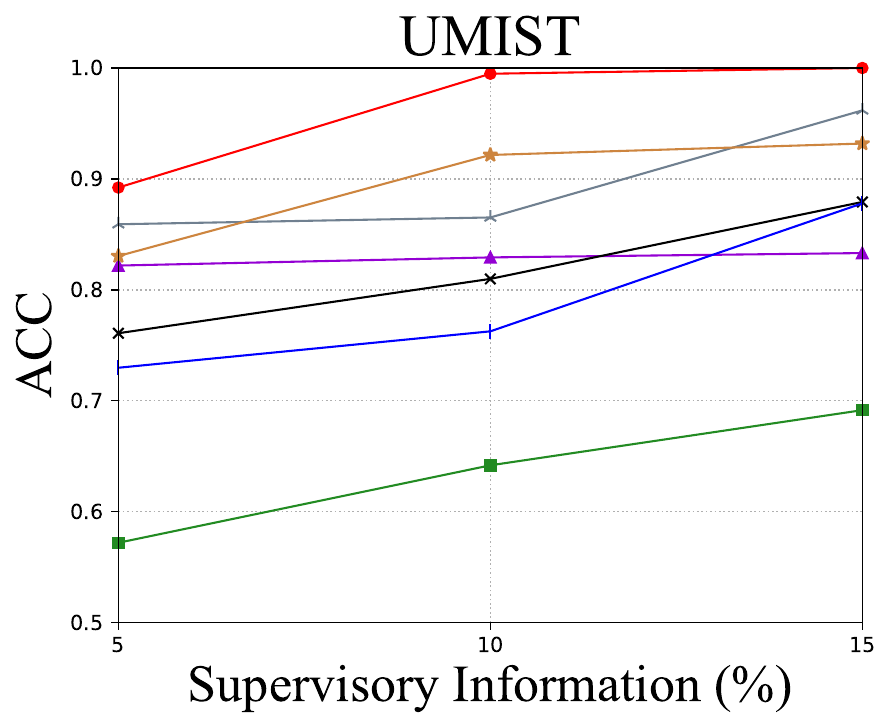}
\label{ACC_label_UMIST}
}\hspace{-1mm}
\subfigure[]{
\includegraphics[width=1.09in, height=0.85in]{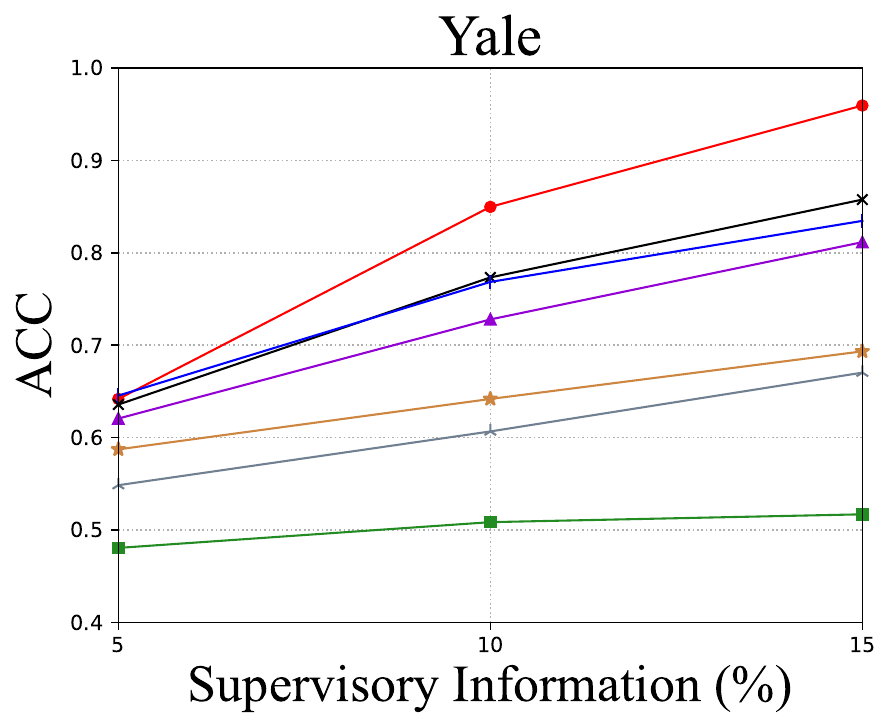} 
\label{ACC_label_Yale} 
}\hspace{-1mm}
\end{center}
\caption{Accuracy of SNMF-based semi-supervised methods on different amounts of supervisory information. The legends of the 6 subfigures are consistent and shown in \subref{ACC_label_BinAlpha}.}
\label{fig4}
\end{figure*}

\begin{figure*}[!t] 
\centering
\begin{center}
\subfigure[]{
\includegraphics[width=1.09in, height=0.85in]{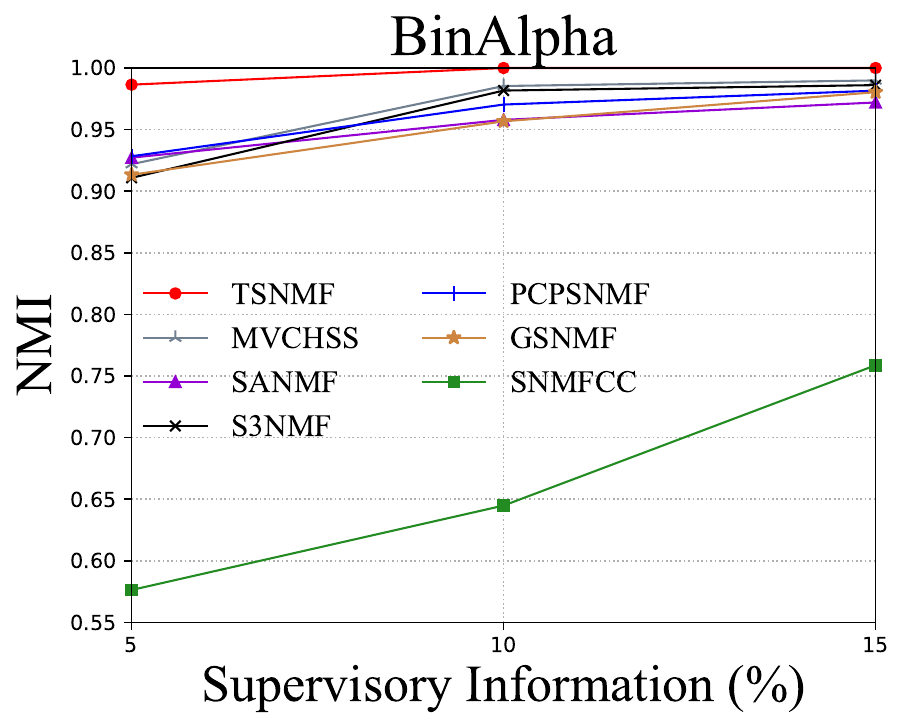}
\label{NMI_label_BinAlpha}
}\hspace{-1mm}
\subfigure[]{
\includegraphics[width=1.09in, height=0.85in]{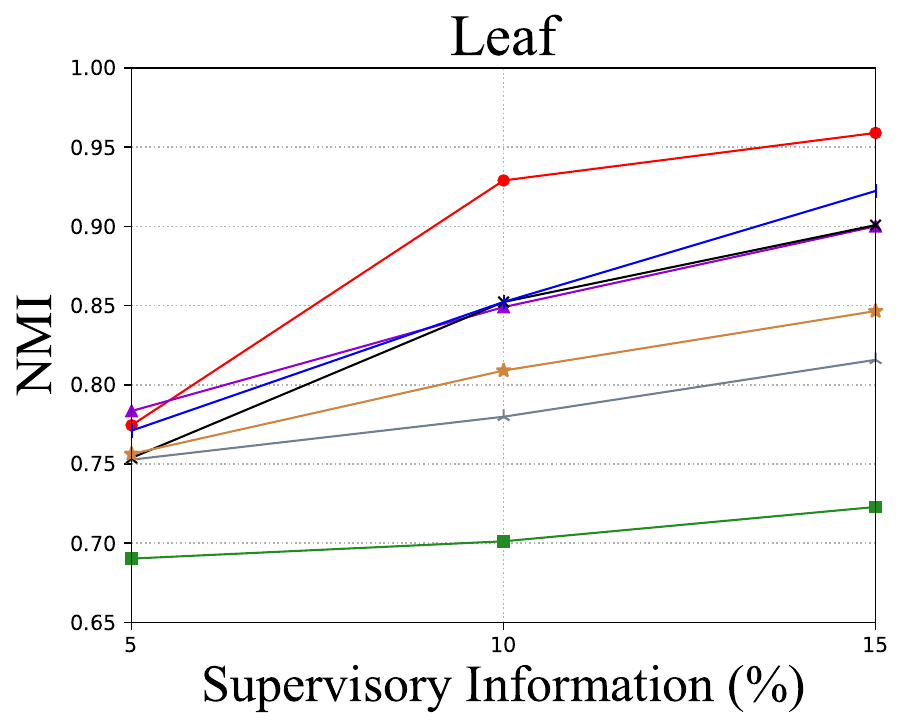}
\label{NMI_label_Leaf}
}\hspace{-1mm}
\subfigure[]{
\includegraphics[width=1.09in, height=0.85in]{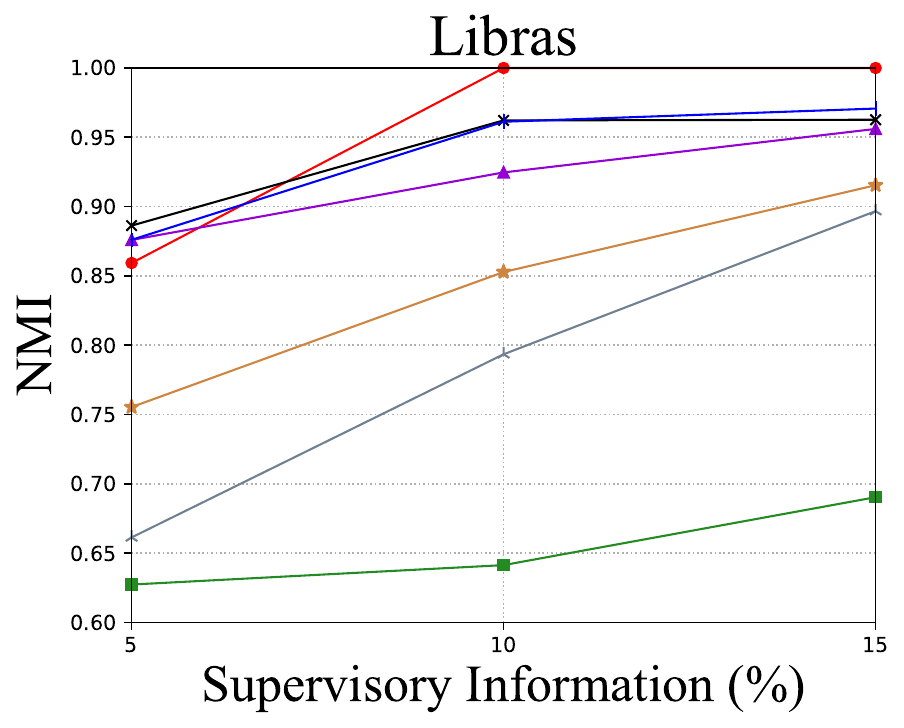}
\label{NMI_label_Libras}
}\hspace{-1mm}
\subfigure[]{
\includegraphics[width=1.09in, height=0.85in]{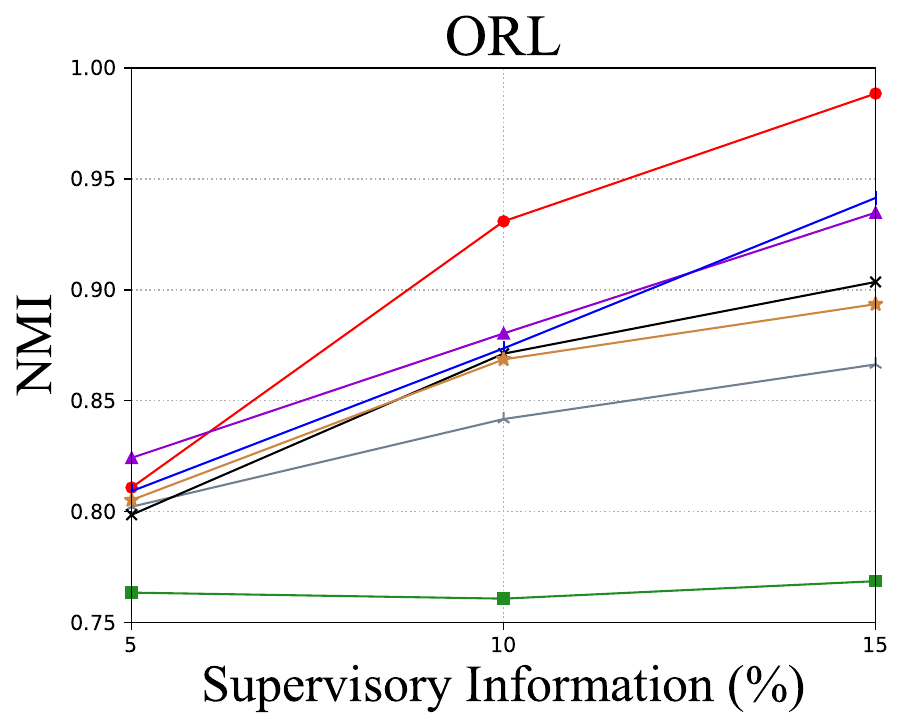} 
\label{NMI_label_ORL} 
}\hspace{-1mm}
\subfigure[]{
\includegraphics[width=1.09in, height=0.85in]{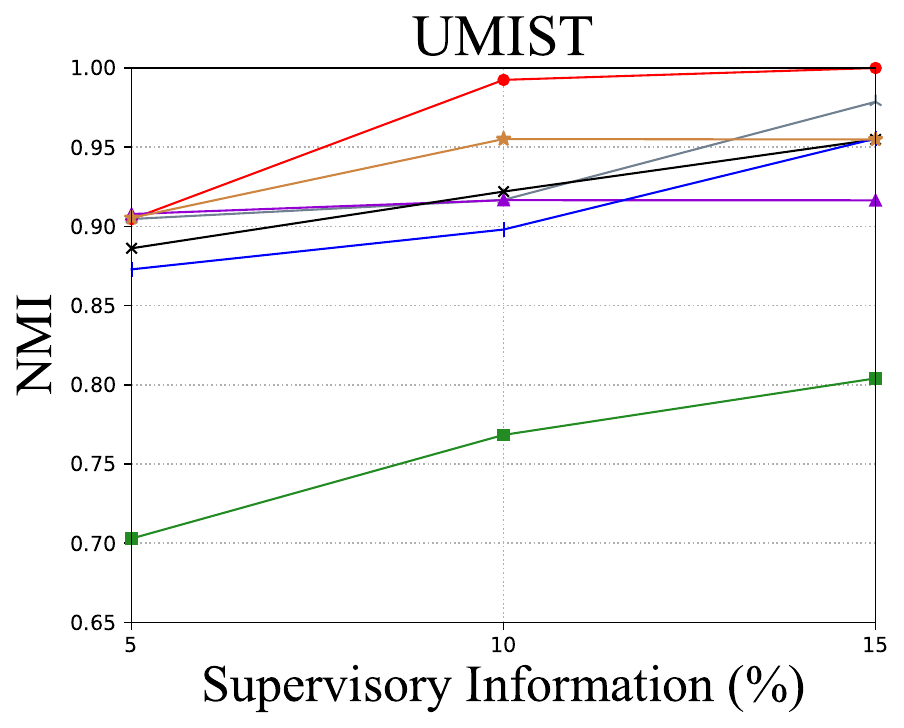}
\label{NMI_label_UMIST}
}\hspace{-1mm}
\subfigure[]{
\includegraphics[width=1.09in, height=0.85in]{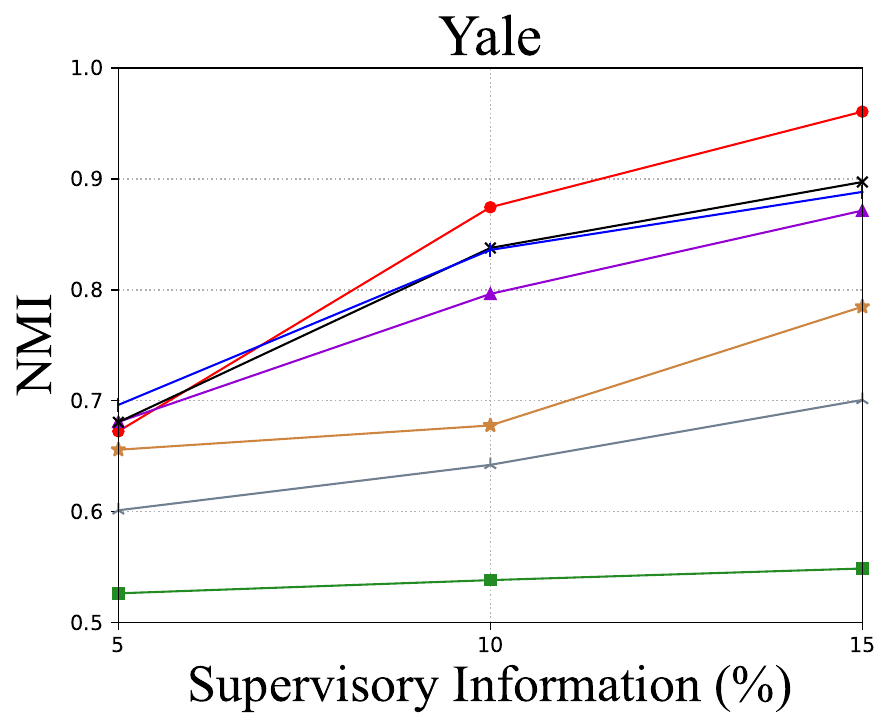} 
\label{NMI_label_Yale} 
}\hspace{-1mm}
\end{center}
\caption{NMI of SNMF-based semi-supervised methods on different amounts of supervisory information. The legends of the 6 subfigures are consistent and shown in \subref{NMI_label_BinAlpha}.}
\label{fig5}
\end{figure*}

In order to investigate the influence of the amount of supervisory information, we selected 5$\%$, 10$\%$, and 15$\%$ of the labels to generate pairwise constraint matrices for the semi-supervised SNMF methods, which are TSNMF, MVCHSS, SANMF, S3NMF, PCPSNMF, GSNMF, and SNMFCC. We selected the first 15 classes from each dataset, and in order to reduce randomness, we repeated each method 10 times on different supervisory information and reported the mean value as the final result. The results are shown in Figs. \ref{fig4} and \ref{fig5}, from which we can draw the following conclusions:

\begin{itemize}
    \item With the increase of supervisory information, the clustering performance of the semi-supervised methods continuously becomes better, indicating that the introduction of supervisory information is important for improving the clustering performance.
    \item The proposed method performs best among the compared algorithms in most cases, indicating that TSNMF is robust to different amounts of supervisory information. It may be because TSNMF exploits the supervisory information more comprehensively than other semi-supervised algorithms, and applies a global prior to incorporating the supervisory information.
    \item The NMI of TSNMF is slightly worse than that of SANMF in the case of Libras and ORL datasets with 5$\%$ supervisory information. However, after slightly increasing the supervisory information, TSNMF outperforms SANMF significantly. For example, when the ORL dataset takes 10$\%$ supervisory information, the NMI of SANMF only increases from 0.82 to 0.88, while that of our method is improved from 0.81 to 0.93.
\end{itemize}

\begin{figure*}[!t] 
\centering
\begin{center}
\subfigure[]{
\includegraphics[width=0.85in, height=0.85in]{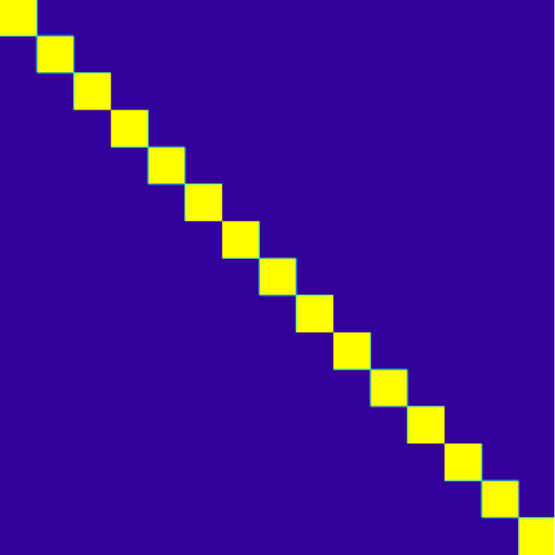} 
\label{standardS} 
}\hspace{-3mm}
\subfigure[]{
\includegraphics[width=0.85in, height=0.85in]{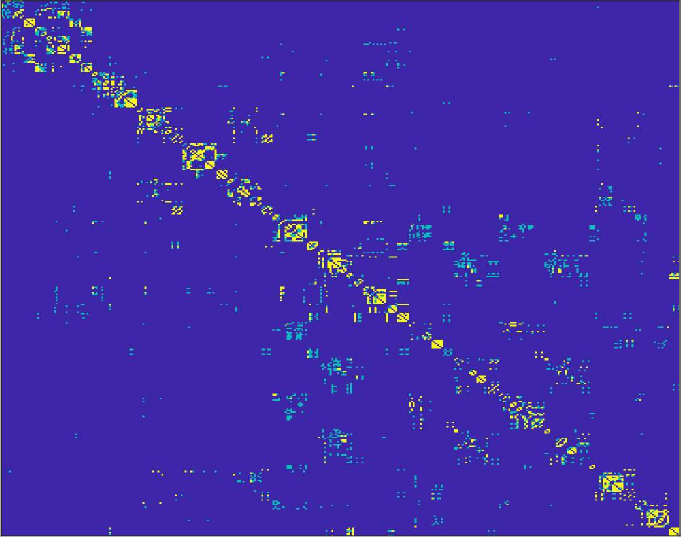} 
\label{initialS} 
}\hspace{-3mm}
\subfigure[]{
\includegraphics[width=0.85in, height=0.85in]{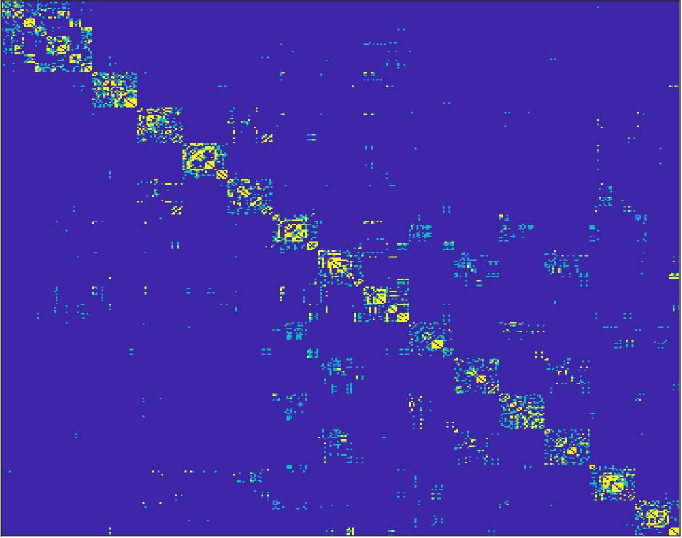}
\label{A1}
}\hspace{-3mm}
\subfigure[]{
\includegraphics[width=0.85in, height=0.85in]{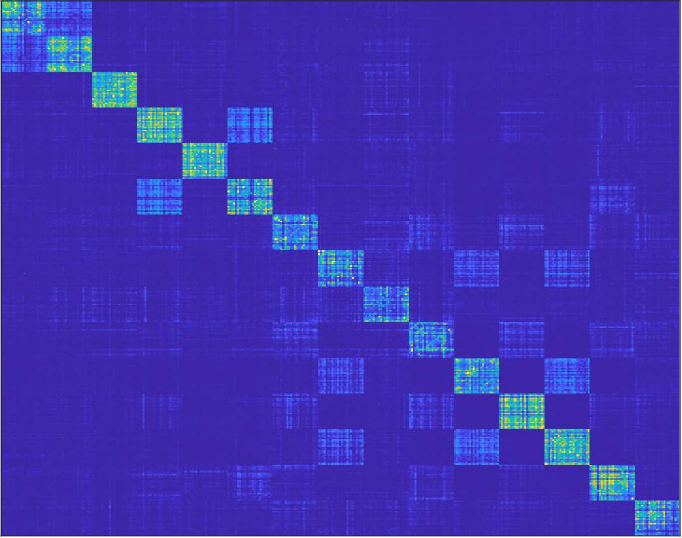}
\label{A2}
}\hspace{-3mm}
\subfigure[]{
\includegraphics[width=0.85in, height=0.85in]{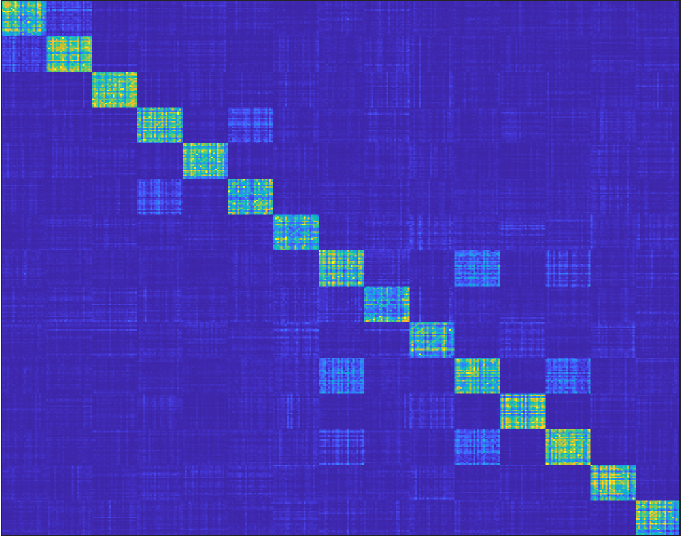} 
\label{A3} 
}\hspace{-3mm}
\subfigure[]{
\includegraphics[width=0.85in, height=0.85in]{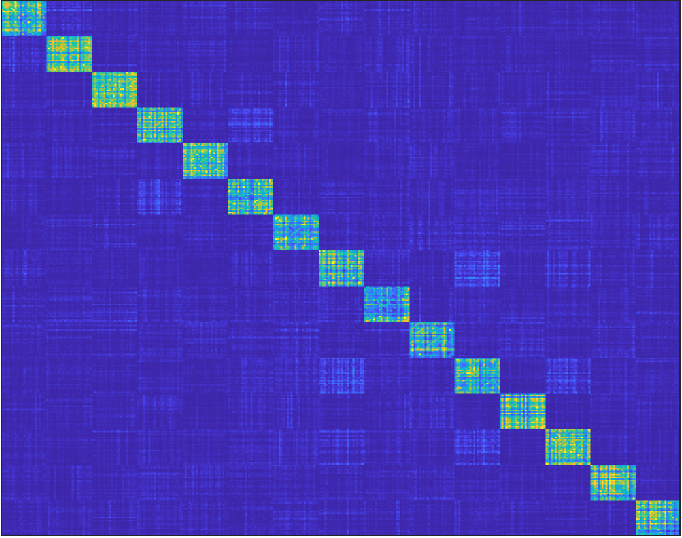}
\label{A4}
}\hspace{-3mm}
\subfigure[]{
\includegraphics[width=0.85in, height=0.85in]{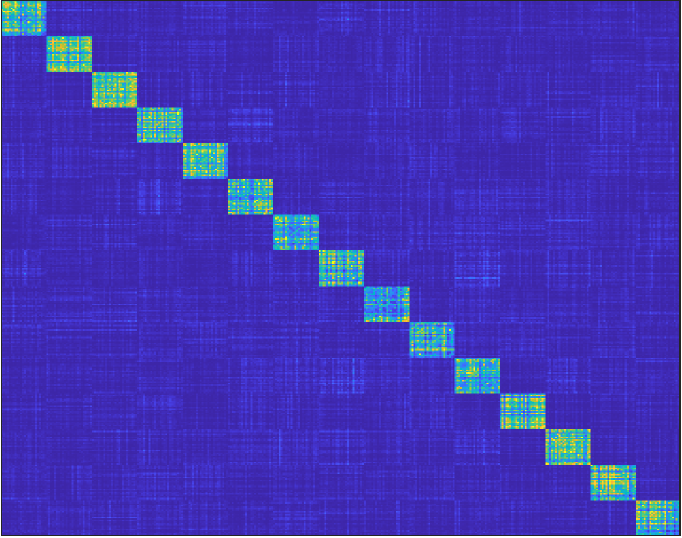}
\label{A5}
}\hspace{-3mm}
\subfigure[]{
\includegraphics[width=0.85in, height=0.85in]{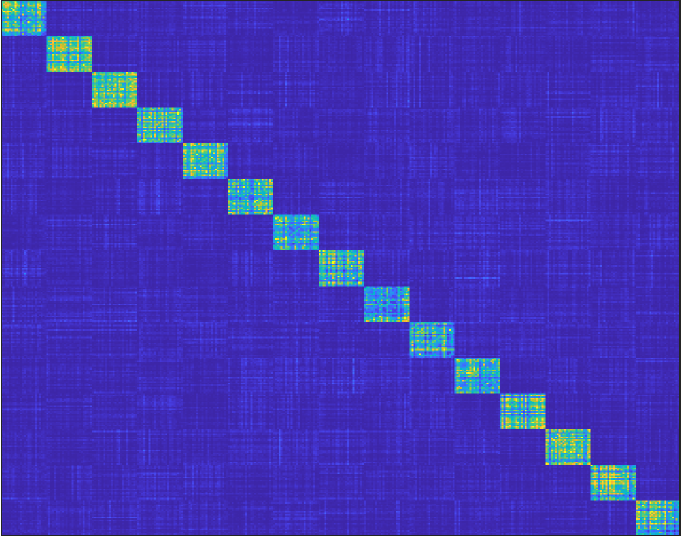}
\label{A6}
}\hspace{-3mm}
\end{center}
\caption{Similarity matrices obtained by our method on the Libras dataset in iterations 1-6. \subref{standardS} The ideal similarity matrix. \subref{initialS} Similarity matrix obtained by the $p\mbox{-}\mathrm{NN}$ method. \subref{A1} Similarity matrix for the first iteration obtained after adjusting \subref{initialS} with the initial pairwise constraint matrix by Eq. (\ref{eq2}). \subref{A2}-\subref{A6} Similarity matrices at iterations 2-6.}
\label{fig6}
\end{figure*}


\begin{figure*}[!t] 
\centering
\begin{center}
\subfigure[]{
\includegraphics[width=1.09in, height=0.85in]{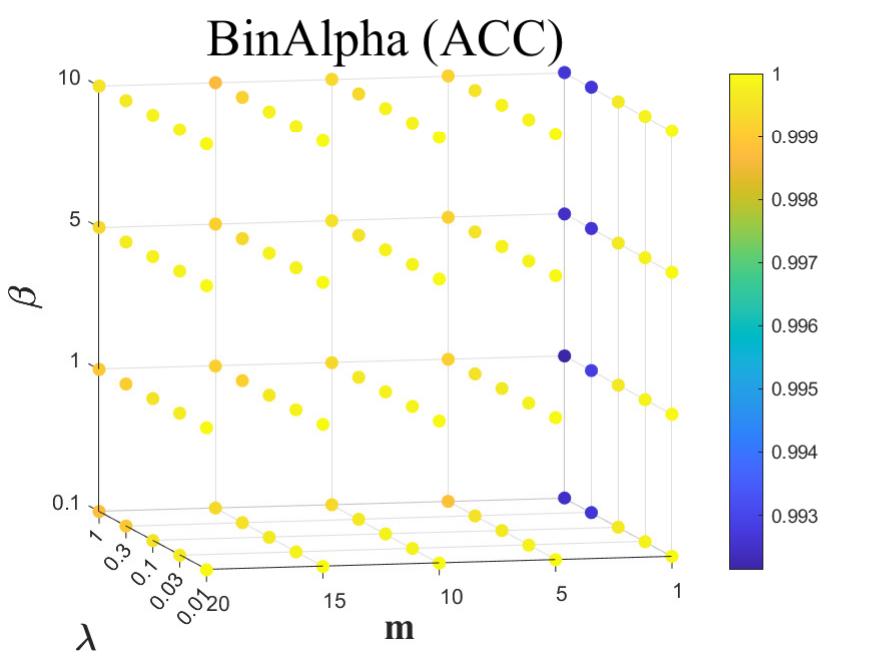}
\label{ACC_para_BinAlpha}
}\hspace{-1mm}
\subfigure[]{
\includegraphics[width=1.09in, height=0.85in]{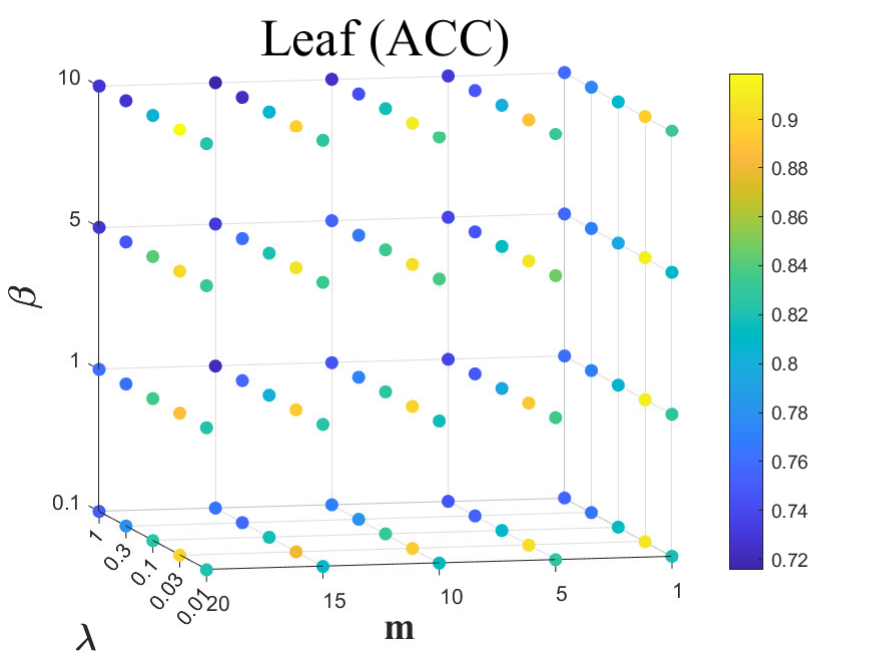} 
\label{ACC_para_Leaf} 
}\hspace{-1mm}
\subfigure[]{
\includegraphics[width=1.09in, height=0.85in]{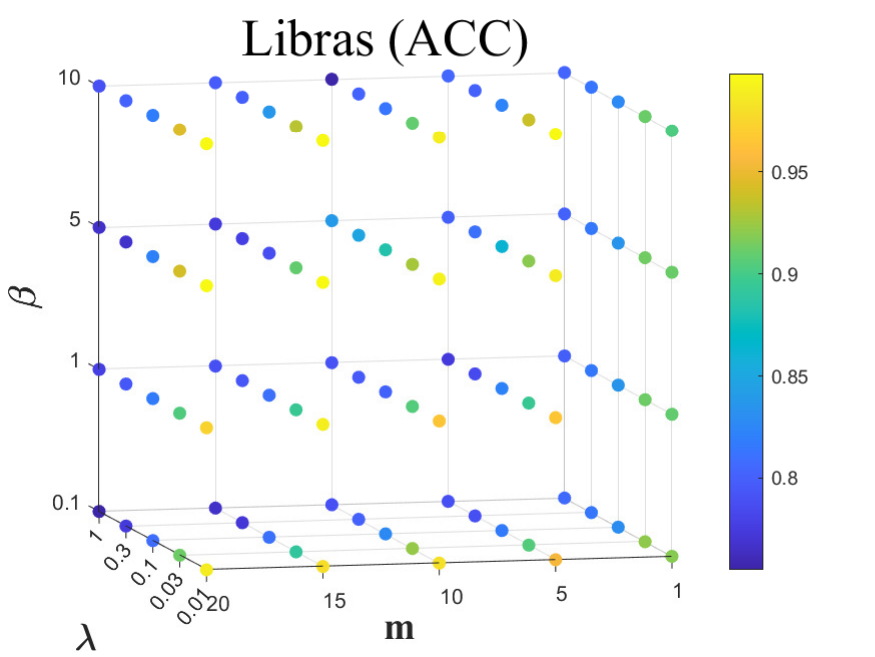}
\label{ACC_para_Libras}
}\hspace{-1mm}
\subfigure[]{
\includegraphics[width=1.09in, height=0.85in]{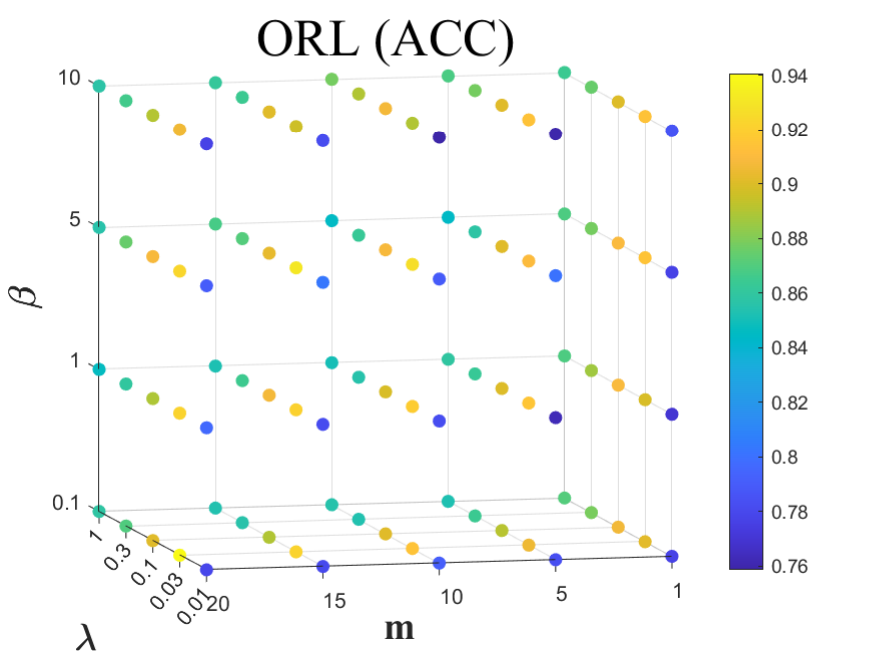} 
\label{ACC_para_ORL} 
}\hspace{-1mm}
\subfigure[]{
\includegraphics[width=1.09in, height=0.85in]{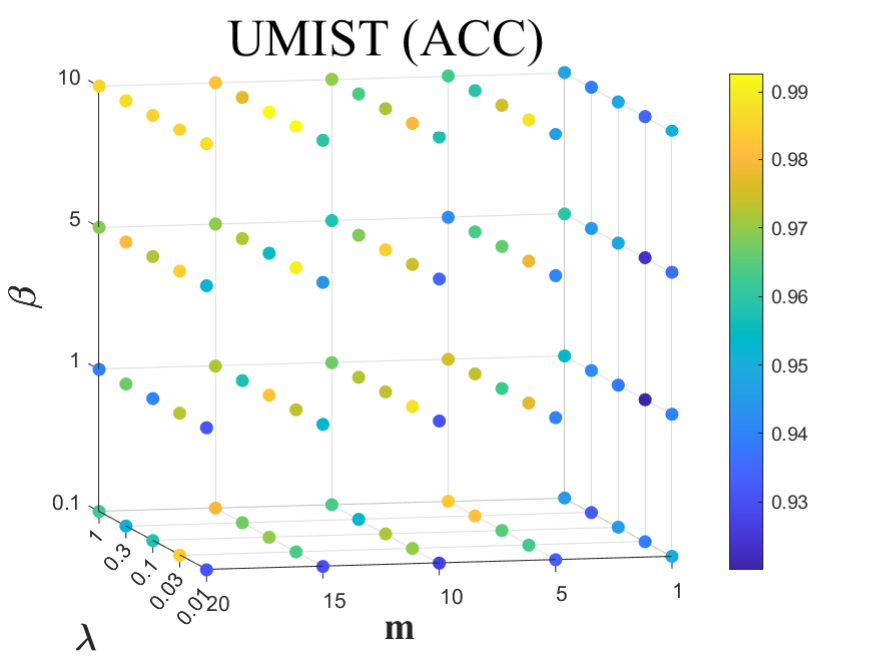}
\label{ACC_para_UMIST}
}\hspace{-1mm}
\subfigure[]{
\includegraphics[width=1.09in, height=0.85in]{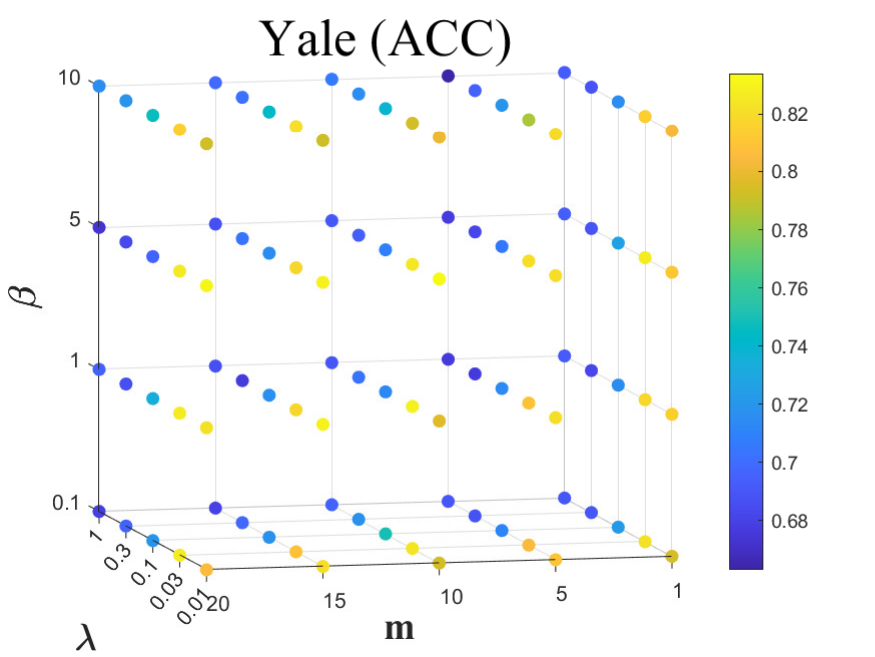}
\label{ACC_para_Yale}
}\hspace{-1mm}
\end{center}
\caption{Impact of parameters $\lambda$, $\beta$ and $m$ on the accuracy of TSNMF. The above figures are all 4-D figures, where the accuracy value is represented by the scatter color indicated in the color bar.}
\label{fig7}
\end{figure*}

\begin{figure*}[!t] 
\centering
\begin{center}
\subfigure[]{
\includegraphics[width=1.09in, height=0.85in]{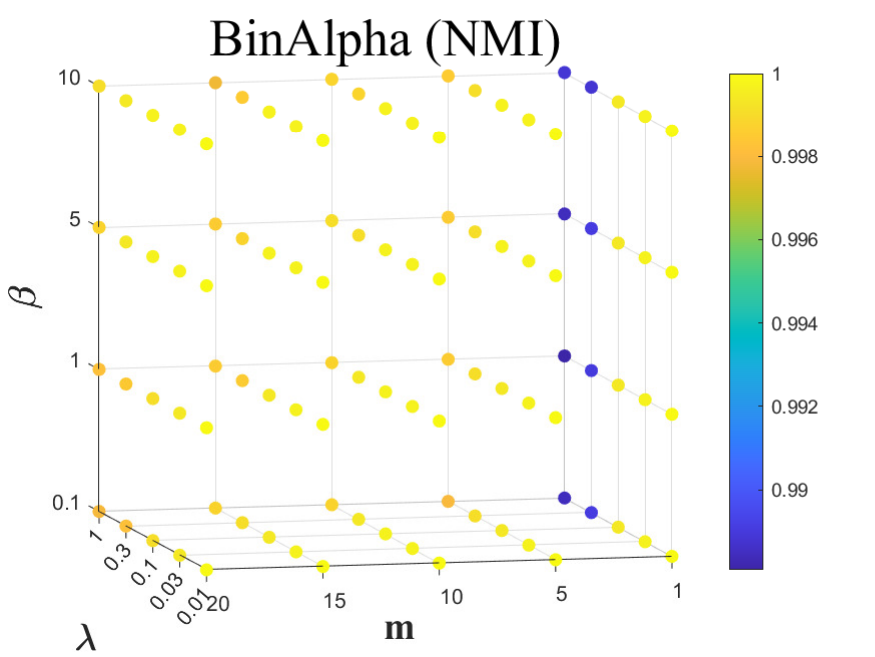}
\label{NMI_para_BinAlpha}
}\hspace{-1mm}
\subfigure[]{
\includegraphics[width=1.09in, height=0.85in]{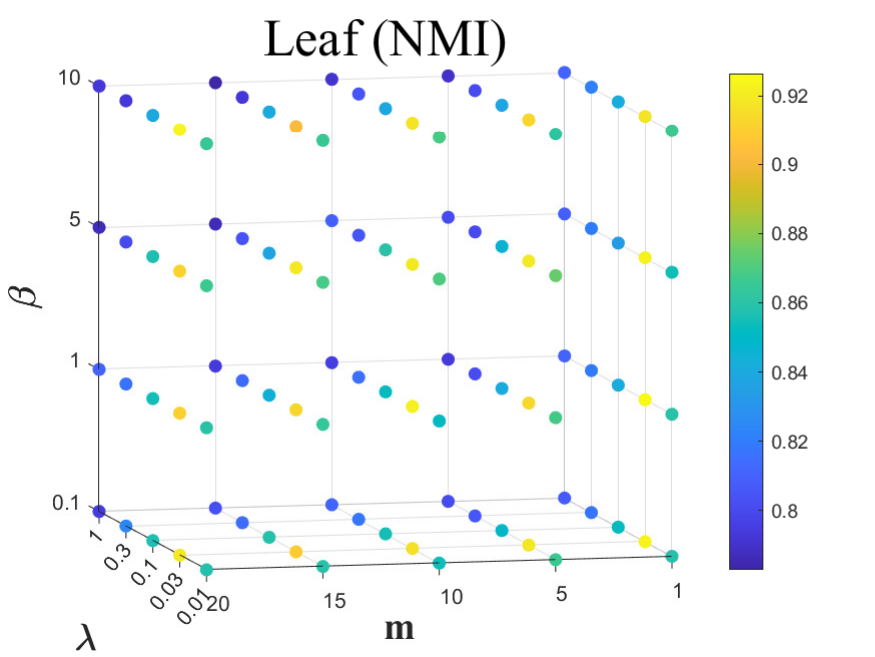} 
\label{NMI_para_Leaf} 
}\hspace{-1mm}
\subfigure[]{
\includegraphics[width=1.09in, height=0.85in]{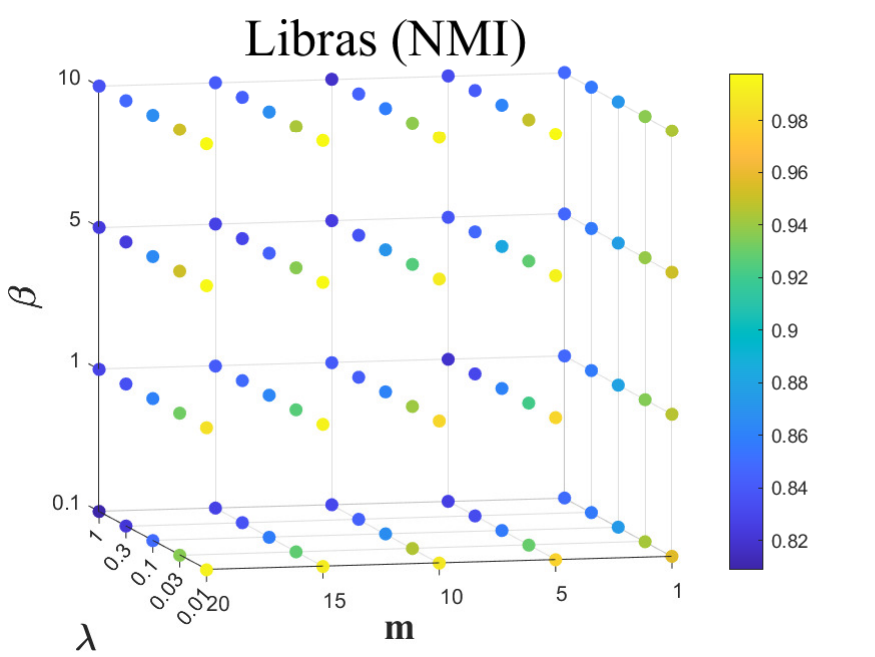}
\label{NMI_para_Libras}
}\hspace{-1mm}
\subfigure[]{
\includegraphics[width=1.09in, height=0.85in]{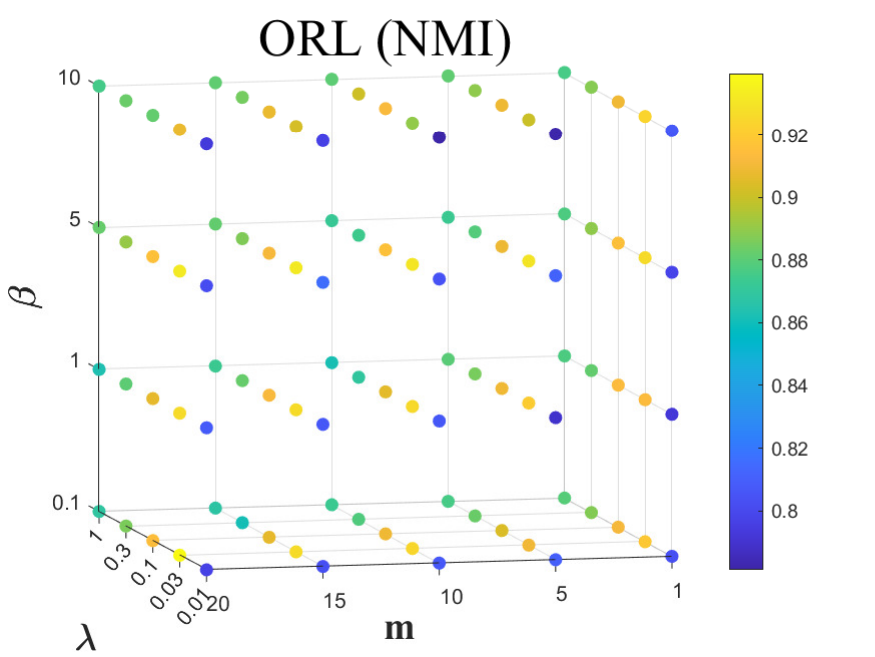} 
\label{NMI_para_ORL} 
}\hspace{-1mm}
\subfigure[]{
\includegraphics[width=1.09in, height=0.85in]{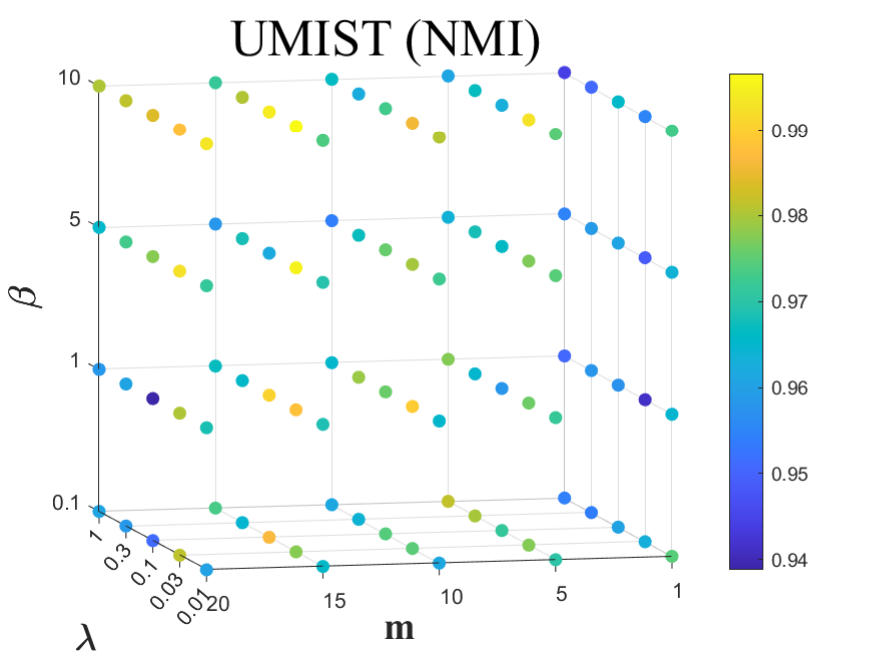}
\label{NMI_para_UMIST}
}\hspace{-1mm}
\subfigure[]{
\includegraphics[width=1.09in, height=0.85in]{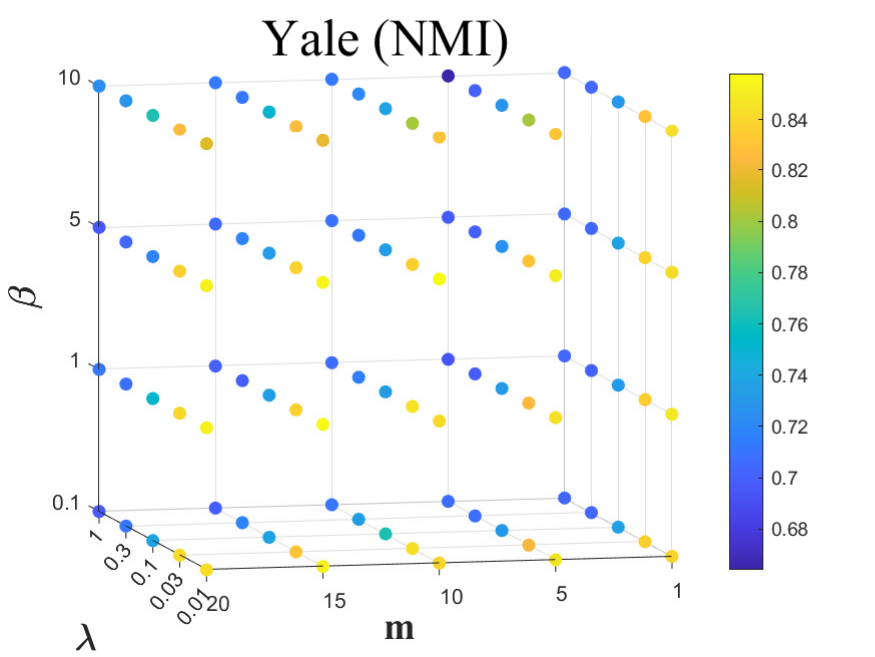}
\label{NMI_para_Yale}
}\hspace{-1mm}
\end{center}
\caption{Impact of parameters $\lambda$, $\beta$ and $m$ on the NMI of TSNMF. The above figures are all 4-D figures, where the NMI value is represented by the scatter color indicated in the color bar.}
\label{fig8}
\end{figure*}

\begin{table}[!t]
\begin{threeparttable}
  \centering
  \caption{ACC for Each Iteration}
    \begin{tabular}{c|c|c|c|c|c|c}
    \toprule
    \textbf{Iters} & \textbf{1} & \textbf{2} & \textbf{3} & \textbf{4} & \textbf{5} & \textbf{6} \\
    \midrule
    Libras & 0.604  & 0.956  & ~\textbf{1.000 } & 1.000  & 1.000  & 1.000  \\
    Leaf & 0.661  & 0.877  & 0.895 & 0.930 & 
    ~\textbf{0.936 }  & 0.924  \\
    UMIST & 0.819  & 0.929  & \textbf{0.946} & 0.909  & 0.909  & 0.909  \\
    Yale   & 0.576  & \textbf{0.879}  & 0.849   & 0.824 & 0.746  & 0.691  \\
    ORL   & 0.774  & 0.915  & ~\textbf{0.923 } & 0.915  & 0.886  & 0.858  \\
    BinAlpha & 0.881  & 0.999  & ~\textbf{1.000 } & 1.000  & 1.000  & 1.000  \\
    \bottomrule
    \end{tabular}%
  \label{tab:ACCIter}%
  \begin{tablenotes}
    \item Iters represents the number of iterations of Algorithm \ref{alg1}.
    \end{tablenotes}
  \end{threeparttable}
\end{table}%
\begin{table}[!t]
\begin{threeparttable}
  \centering
  \caption{NMI for Each Iteration}
    \begin{tabular}{c|c|c|c|c|c|c}
    \toprule
    \textbf{Iters} & \textbf{1} & \textbf{2} & \textbf{3} & \textbf{4} & \textbf{5} & \textbf{6} \\
    \midrule
    Libras & 0.716  & 0.977  & ~\textbf{1.000 } & 1.000  & 1.000  & 0.999  \\
    Leaf & 0.755  & 0.893  & 0.909 & 0.917 & ~\textbf{0.928 }  & 0.914  \\
    UMIST & 0.876  & 0.966  & ~\textbf{0.974 } & 0.974  & 0.974  & 0.974  \\
    Yale   & 0.614  & \textbf{0.860}  & 0.828   & 0.805 & 0.743  & 0.687  \\
     ORL   & 0.813  & 0.920  & ~\textbf{0.928 } & 0.914  & 0.877  & 0.850  \\
    BinAlpha & 0.896  & 0.999  & ~\textbf{1.000 } & 1.000  & 1.000  & 1.000  \\
    \bottomrule
    \end{tabular}%
  \label{tab:NMIIter}%
   \begin{tablenotes}
    \item Iters represents the number of iterations of Algorithm \ref{alg1}.
    \end{tablenotes}
  \end{threeparttable}
\end{table}%

\subsection{Analysis of the Iteration Times of Algorithm \ref{alg1}}\label{IterAlg1}
We study the optimal number of iterations for Algorithm \ref{alg1} on different datasets. According to Tables \ref{tab:ACCIter} and \ref{tab:NMIIter}, the proposed method shows a significant improvement in the second iteration compared to the first one on most datasets, which proves that applying the global low-rank constraint to tensor composed of the similarity matrix and the pairwise constraint matrix exploits the supervisory information substantially. Since the second iteration, the clustering performance increases slowly and then remains relatively stable. Therefore, we suggest setting $maxIter=3$ for Algorithm \ref{alg1}. 

\if{}
\begin{figure}[!t]
  \centering
  \includegraphics[width=\linewidth]{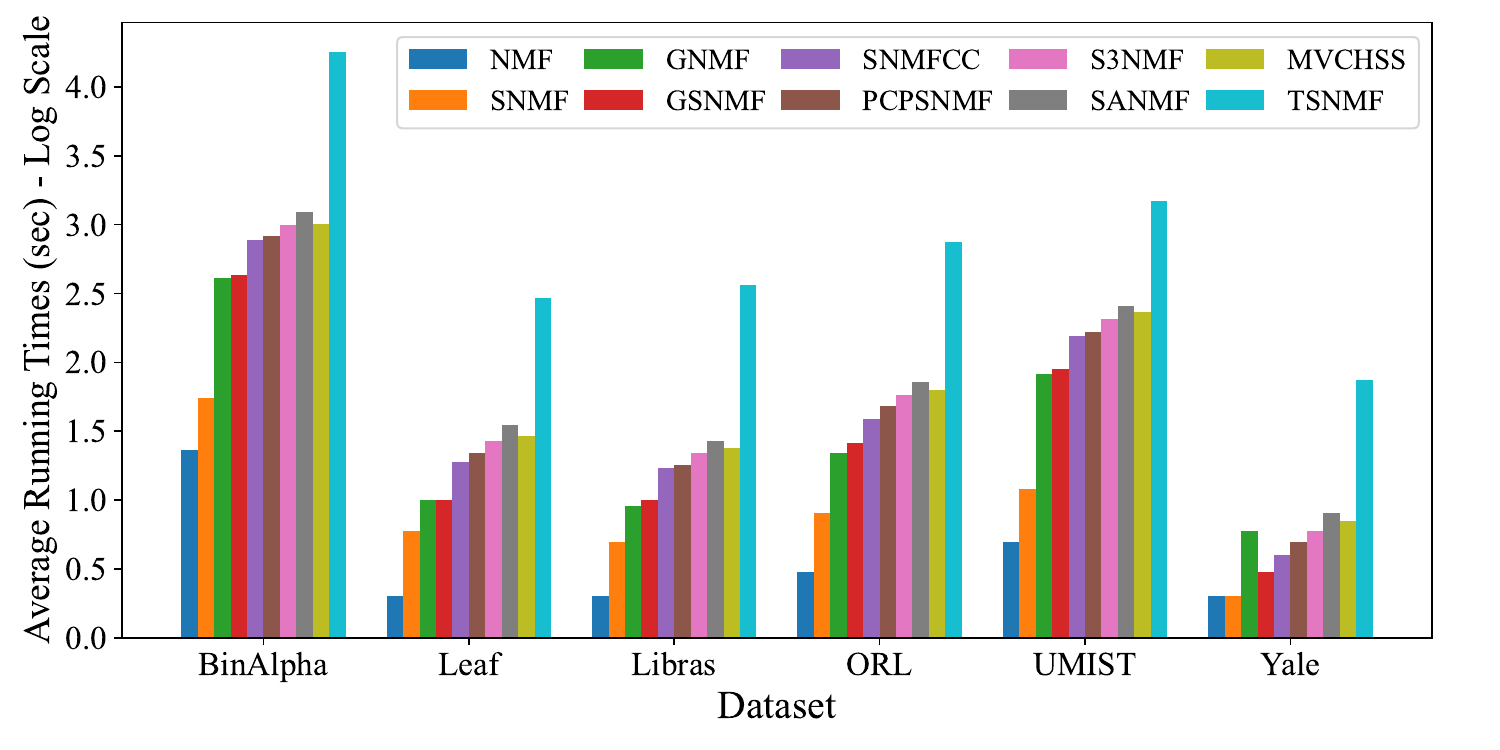}
  \caption{The results of the running time (in seconds) comparison.}
  \label{fig:run_time}
\end{figure}
\fi

Fig. \ref{fig6} shows that as the number of iterations increases, the similarity matrix continues to strengthen. Among them, Fig. \ref{initialS} is the similarity matrix generated directly by $p\mbox{-}\mathrm{NN}$ \cite{8361078}, and Figs. \ref{A1}-\subref{A6} are the similarity matrices input into Algorithm \ref{alg2} in the 1st-5th iteration of Algorithm \ref{alg1}. It can be seen that compared with Fig. \ref{A1}, Fig. \ref{A2} removes many incorrect connections and increases the number of correct connections, which is consistent with the experimental results that the clustering performance from the first iteration to the second iteration is greatly improved. In Figs. \ref{A3}-\subref{A6}, there are fewer and fewer incorrect connections, and the similarity matrix is getting denser, which explains why the clustering performance will be improved as the number of iterations of Algorithm \ref{alg1} increases.

\subsection{Hyper-parameter Sensitives}
There are three hyper-parameters in our method, which are the number of the embedding matrices $m$, the coefficient of the error term $\lambda$, and the smoothness term $\beta$. In this subsection, we study their impact on the clustering performance. The value ranges of $m$, $\lambda$ and $\beta$ are $\{1, 5, 10, 15, 20\}$, $\{0.01, 0.03, 0.1, 0.3, 1\}$ and $\{0.1,1,5,10\}$, respectively. We still take the first 15 classes from each dataset and record the average results of 10 repetitions in Figs. \ref{fig7} and \ref{fig8}.

It can be seen from Figs. \ref{fig7} and \ref{fig8} that when the value of $m$ is small, the clustering performance is poor, as a small $m$ means large randomness of the initialization. After increasing the value of $m$, ACC and NMI become better and tend to be stable. Satisfactory performance can already be achieved when $m=10$. In addition, it is observed that TSNMF performs better when the value of $\lambda$ is relatively small, that is, between $\{0.01, 0.03, 0.1\}$. Different from $\lambda$, when the value of $\beta$ is relatively large, namely $\{1, 5, 10\}$, the ACC and NMI are larger, indicating the importance of item $\|\bm{\alpha}\|_{F}^{2}$. The reason is that the consistent embedding matrix needs to involve more embedding matrices to prevent bad results from being dominated by a poor embedding matrix. Moreover, almost all datasets achieve the best clustering performance when $\lambda$ is about 0.03 and $\beta$ is about 5. Taking the above analyses into account, we suggest $m=10$, $\lambda=0.03$, $\beta=5$ for our method.

\begin{table}[!]
\scriptsize
\begin{threeparttable}
  \centering
  \caption{Clustering Performance Comparison between the Proposed Enhanced SNMF and SNMF}
    \begin{tabular}{@{\hspace{6pt}}c@{\hspace{6pt}}c@{\hspace{6pt}}c@{\hspace{6pt}}|c@{\hspace{6pt}}c@{\hspace{6pt}}c@{\hspace{6pt}}c@{\hspace{6pt}}c@{\hspace{6pt}}c@{\hspace{6pt}}c@{\hspace{6pt}}c@{\hspace{6pt}}}
    \hline
    \textbf{ACC} & \textbf{ESNMF} & \textbf{SNMF} & \textbf{NMI} & \textbf{ESNMF} & \textbf{SNMF} \\
    \hline
    BinAlpha & 0.435±0.014 & 0.425±0.016 & BinAlpha & 0.587±0.006 & 0.582±0.008 \\
    Leaf  & 0.518±0.011 & 0.506±0.019 & Leaf  & 0.695±0.009 & 0.691±0.008 \\
    Libras & 0.499±0.020 & 0.487±0.016 & Libras & 0.628±0.016 & 0.624±0.012 \\
    ORL   & 0.637±0.014 & 0.632±0.013 & ORL   & 0.783±0.005 & 0.779±0.007 \\
    UMIST & 0.518±0.023 & 0.514±0.022 & UMIST & 0.707±0.012 & 0.701±0.020 \\
    Yale & 0.504±0.028 & 0.499±0.030 & Yale & 0.542±0.018 & 0.541±0.013 \\
    \hline
    \end{tabular}%
  \label{tab:ESNMF}%
  \begin{tablenotes}
    \item ESNMF stands for the proposed enhanced SNMF.
    \end{tablenotes}
  \end{threeparttable}
\end{table}%

\subsection{Effectiveness of the Enhanced SNMF}
In Table \ref{tab:ESNMF}, we compare the clustering performance of the proposed Enhanced SNMF with SNMF. Specifically, we set the maximum number of iterations to 500 and fixed $m=10$. It is clear that both the ACC and NMI of the Enhanced SNMF are superior to those of SNMF on 6 datasets because the Enhanced SNMF can automatically select several embedding matrices and weight them to obtain a highly consistent embedding matrix as the final clustering result, which improves the clustering performance.

Moreover, we evaluate the performance of our method with SNMF as input in Table \ref{tab:Enhanced}, where we find that incorporating the enhanced SNMF in our method will significantly outperform that with SNMF on all datasets. The reason is that the embedding matrix generated by SNMF is already low-rank rank; accordingly, using it in tensor low-rank representation will lose the designed effect.

\begin{table}
  \caption{The Performance of Our Method under the Enhanced SNMF and SNMF, Respectively.}
  \label{tab:Enhanced}
  \centering
  \resizebox{\linewidth}{!}{\begin{tabular}{llcccccc}
    \toprule
    Metric & Method & Libras & Leaf & UMIST & Yale & ORL & BinAlpha \\
    \midrule
    \multirow{2}{*}{ACC} & TSNMF w/ Enhanced SNMF & 1.000 & 0.874 & 0.963 & 0.850 & 0.898 & 1.000 \\
     &  TSNMF w/ SNMF & 0.928 & 0.763 & 0.803 & 0.726 & 0.697 & 0.981 \\
    \midrule
    \multirow{2}{*}{NMI} & TSNMF w/ Enhanced SNMF & 1.000 & 0.919 & 0.977 & 0.874 & 0.939 & 1.000 \\
     & TSNMF w/ SNMF & 0.967 & 0.791 & 0.877 & 0.764 & 0.739 & 0.989 \\
    \bottomrule
  \end{tabular}}
\end{table}

\subsection{Effectivness of Eq. \eqref{eq2}}
We evaluate the effectiveness of Eq. \eqref{eq2} in Table \ref{tab:eq15}, where we can find that this operation significantly improves the clustering performance on all datasets. For example, on Leaf, the ACC has been increased from 0.828 to 0.875, and on ORL, the NMI has been increased from 0.882 to 0.939. Therefore, we can conclude that Eq. \eqref{eq2} is effective in improving the clustering performance.

\begin{table}
  \caption{Effectivness of Eq. \eqref{eq2}.}
  \label{tab:eq15}
  \centering
  \resizebox{\linewidth}{!}{\begin{tabular}{llcccccc}
    \toprule
    Metric & Method & BinAlpha & Leaf & Libras & ORL & UMIST & Yale\\
    \midrule
    \multirow{2}{*}{ACC} & TSNMF & 1.000 & 0.874 & 1.000 & 0.898 & 0.963 & 0.850\\
     & TSNMF w/o Eq. \eqref{eq2} & 0.965 & 0.828 & 0.965 & 0.843 & 0.929 & 0.848 \\
    \midrule
    \multirow{2}{*}{NMI} & TSNMF & 1.000 & 0.919 & 1.000 & 0.939 & 0.977 & 0.874\\
    & TSNMF w/o Eq. \eqref{eq2} & 0.983 &  0.878 & 0.981 & 0.882 & 0.960 & 0.852\\
    \bottomrule
  \end{tabular}}
\end{table}

\subsection{Results on Large-scale Datasets}

We evaluated our method on three datasets with more samples and a larger number of clusters. Information about the datasets can be found in Table \ref{tab:Dataset}. The experimental results are shown in Table \ref{tab:largeMainresults}, where we find that our method still significantly outperforms the previous methods on all those large-scale datasets. For example, on Bookmark, the clustering ACC is increased from 0.713 to 0.892. 

\begin{table}[!t]
  \centering
  \caption{Information About Three Large-scale Datasets. 
  }
  \label{tab:Dataset}%
    \begin{tabular}{ccccc}
    \toprule
    Dataset & $n$ & $m$ & $k$\\
    \midrule
    TMC & 8670 & 981 & 18 \\
    Delicious & 1409  & 1389  & 20\\
    Bookmark & 2500 & 1413 & 57 \\
    \bottomrule
    \end{tabular}%
\end{table}

\begin{table*}[!h]
\caption{Results on Large-scale Datasets.}
\label{tab:largeMainresults}
\centering
\resizebox{\linewidth}{!}{
\begin{threeparttable}
\begin{tabular}{lcccccccccc}
    \toprule
    \toprule
    \textbf{ACC} & \textbf{NMF\cite{lee2001algorithms}} & \textbf{SNMF\cite{kuang2012symmetric}} & \textbf{GNMF\cite{5674058}} & \textbf{GSNMF\cite{6985550}} & \textbf{SNMFCC\cite{7167693}} & \textbf{PCPSNMF\cite{8361078}} & \textbf{S3NMF\cite{9543530}} & \textbf{SANMF\cite{9013063}} & \textbf{MVCHSS\cite{10076472}} & \textbf{TSNMF} \\
    \midrule
    TMC & 0.067±0.012 $\bullet$ & 0.211±0.023 $\bullet$ & 0.155±0.018 $\bullet$ & 0.475±0.024 $\bullet$ & 0.230±0.035 $\bullet$ & \underline{0.963±0.021} $\bullet$ & 0.938±0.014 $\bullet$ & 0.928±0.021 $\bullet$ & 0.941±0.023 $\bullet$ & \textbf{1.000±0.000} \\
    Delicious & 0.131±0.020 $\bullet$ & 0.154±0.018 $\bullet$ & 0.132±0.032 $\bullet$ & 0.693±0.042 $\bullet$ & 0.181±0.019 $\bullet$ & \underline{0.709±0.087} $\bullet$ & 0.704±0.077 $\bullet$ & 0.688±0.089 $\bullet$ & 0.643±0.029 $\bullet$ & \textbf{0.956±0.035} \\
    Bookmark & 0.150±0.034 $\bullet$ & 0.151±0.023 $\bullet$ & 0.137±0.025 $\bullet$ & 0.182±0.041 $\bullet$ & 0.201±0.044 $\bullet$ & 0.682±0.057 $\bullet$ & 0.620±0.046 $\bullet$ & 0.699±0.068 $\bullet$ & \underline{0.713±0.063} $\bullet$ & \textbf{0.892±0.055}\\
    \midrule
    \midrule
    \textbf{NMI} & \textbf{NMF\cite{lee2001algorithms}} & \textbf{SNMF\cite{kuang2012symmetric}} & \textbf{GNMF\cite{5674058}} & \textbf{GSNMF\cite{6985550}} & \textbf{SNMFCC\cite{7167693}} & \textbf{PCPSNMF\cite{8361078}} & \textbf{S3NMF\cite{9543530}} & \textbf{SANMF\cite{9013063}} & \textbf{MVCHSS\cite{10076472}} & \textbf{TSNMF} \\
    \midrule
    TMC & 0.030±0.014 $\bullet$ & 0.047±0.018 $\bullet$ & 0.113±0.015 $\bullet$ & 0.603±0.008 $\bullet$ & 0.182±0.018 $\bullet$ & \underline{0.981±0.009} $\bullet$ & 0.967±0.008 $\bullet$ & 0.959±0.014 $\bullet$ & 0.963±0.018 $\bullet$ & \textbf{1.000±0.001} \\
    Delicious & 0.047±0.027 $\bullet$ & 0.541±0.013 $\bullet$ & 0.044±0.025 $\bullet$ & 0.759±0.034 $\bullet$ & 0.170±0.015 $\bullet$ & 0.855±0.073 $\bullet$ & \underline{0.870±0.056} $\bullet$ & 0.786±0.067 $\bullet$ & 0.792±0.023 $\bullet$ & \textbf{0.946±0.033} \\
    Bookmark & 0.325±0.028 $\bullet$ & 0.370±0.036 $\bullet$ & 0.367±0.022 $\bullet$ & 0.473±0.048 $\bullet$ & 0.488±0.039 $\bullet$ & 0.777±0.069 $\bullet$ & 0.714±0.073 $\bullet$ & 0.825±0.077 $\bullet$ & \underline{0.827±0.063} $\bullet$ & \textbf{0.899±0.062}\\
    \bottomrule
    \bottomrule
\end{tabular}

\begin{tablenotes}
    \item The best ACC/NMI in each dataset is presented in bold, while the second-best is underlined. $\bullet$/$\circ$ indicates whether TSNMF is significantly better than the compared algorithm according to the pairwise $t$-test at the significance level of 0.05.
\end{tablenotes}
\end{threeparttable}
}
\end{table*}

\section{Conclusion}\label{Conclusion}
In this paper, we have presented a novel SNMF-based model TSNMF that incorporates pairwise constraints by seeking the tensor low-rank representation. Compared with the previous semi-supervised algorithm, we improve the propagation of the supervisory information from a local perspective to a global perspective, making greater use of supervisory information. We also propose an enhanced SNMF tailored to the tensor low-rank representation. We provide an iterative and alternative optimization algorithm to solve the proposed model. The similarity matrix and the pairwise constraint matrix are continuously strengthened during the iterative process, and we give the empirical optimal number of iterations. The proposed model outperforms the SOTA methods significantly on extensive datasets and settings.

Although the proposed model produces significantly better clustering performance than the previous SOTA methods, it needs to iteratively and alternatively update tensor low-rank representation and symmetric non-negative matrix factorization, leading to a complexity of $\mathcal{O}(n^3T)$, where $n$ is the number of samples and $T$ is the number of iterations for Algorithm 1. Although we have empirically shown that $T$ can be set to a quite small value, i.e., $T=3$, the proposed model still has relatively high computational complexity. Reducing computational costs is an important future direction.

\vspace{-15pt}
\begin{IEEEbiography}[{\includegraphics[width=1in,height=1.25in,clip,keepaspectratio]{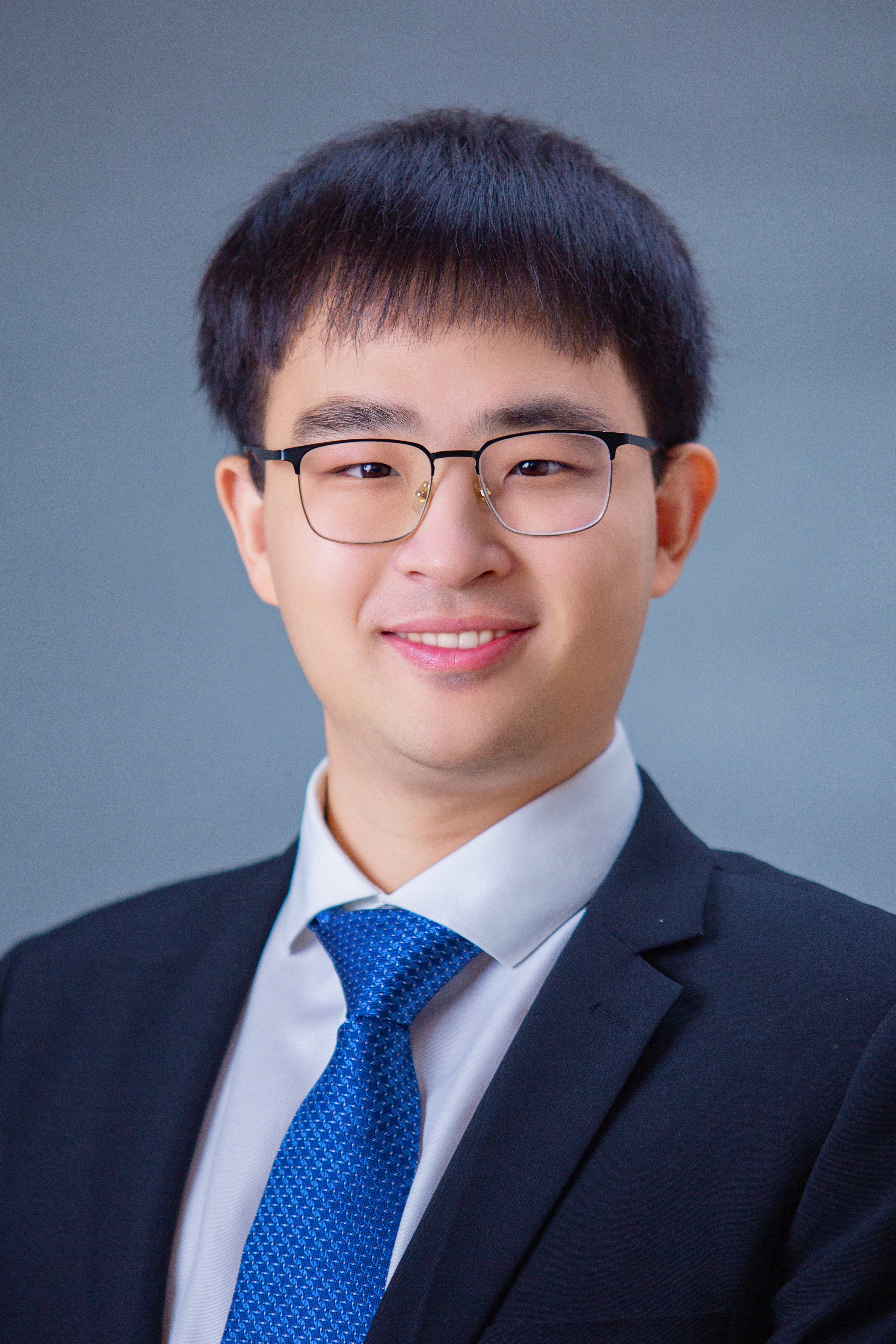}}]{Yuheng Jia}
(Member, IEEE) received the B.S. degree in automation and the M.S. degree in control theory and engineering from Zhengzhou University, Zhengzhou, China, in 2012 and 2015, respectively, and the Ph.D. degree in computer science from the City University of Hong Kong, Hong Kong, China, in 2019. He is currently an Associate Professor at the School of Computer Science and Engineering, Southeast University, Nanjing, China. His research interests broadly include topics in machine learning and data representation, such as un/semi/weakly-supervised learning, high-dimensional data modeling and analysis, and low-rank tensor/matrix approximation and factorization.
\end{IEEEbiography}

\vspace{-15pt}

\begin{IEEEbiography}[{\includegraphics[width=1in,height=1.25in,clip,keepaspectratio]{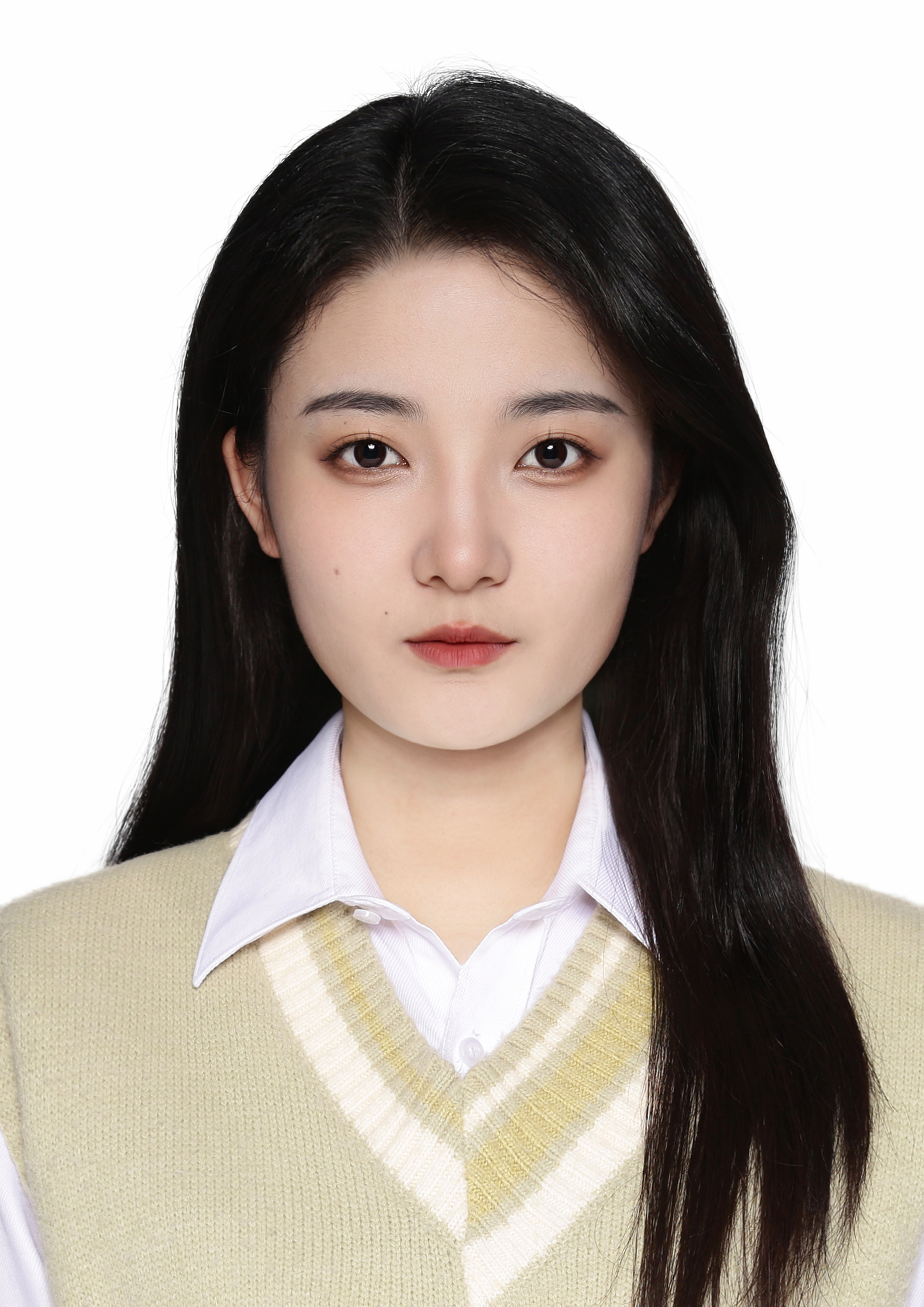}}]{Jia-Nan Li}
is currently pursuing a bachelor’s degree with the School of Computer Science and Engineering, Southeast University, Nanjing, China. Her research interests include semi-supervised learning and machine learning.
\end{IEEEbiography}

\vspace{-15pt}

\begin{IEEEbiography}[{\includegraphics[width=1in,height=1.25in,clip,keepaspectratio]{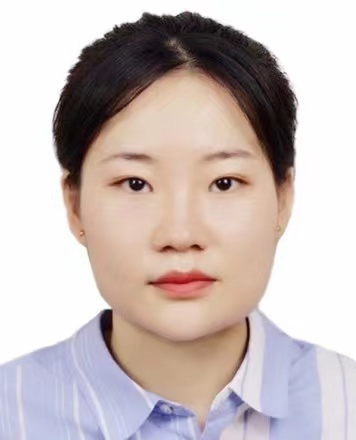}}]{Wenhui Wu}
received the B.S. and M.S. degrees from Xidian University, Xian, China, in 2012 and 2015, respectively, and the Ph.D. degree in computer science from the City University of Hong Kong, Hong Kong, China, in 2019. She is currently an Associate Professor with the College of Electronics and Information Engineering, Shenzhen University, Shenzhen, China. Her current research interests include machine learning, image enhancement, and network data clustering.
\end{IEEEbiography}

\vspace{-15pt}

\begin{IEEEbiography}[{\includegraphics[width=1in,height=1.25in,clip,keepaspectratio]{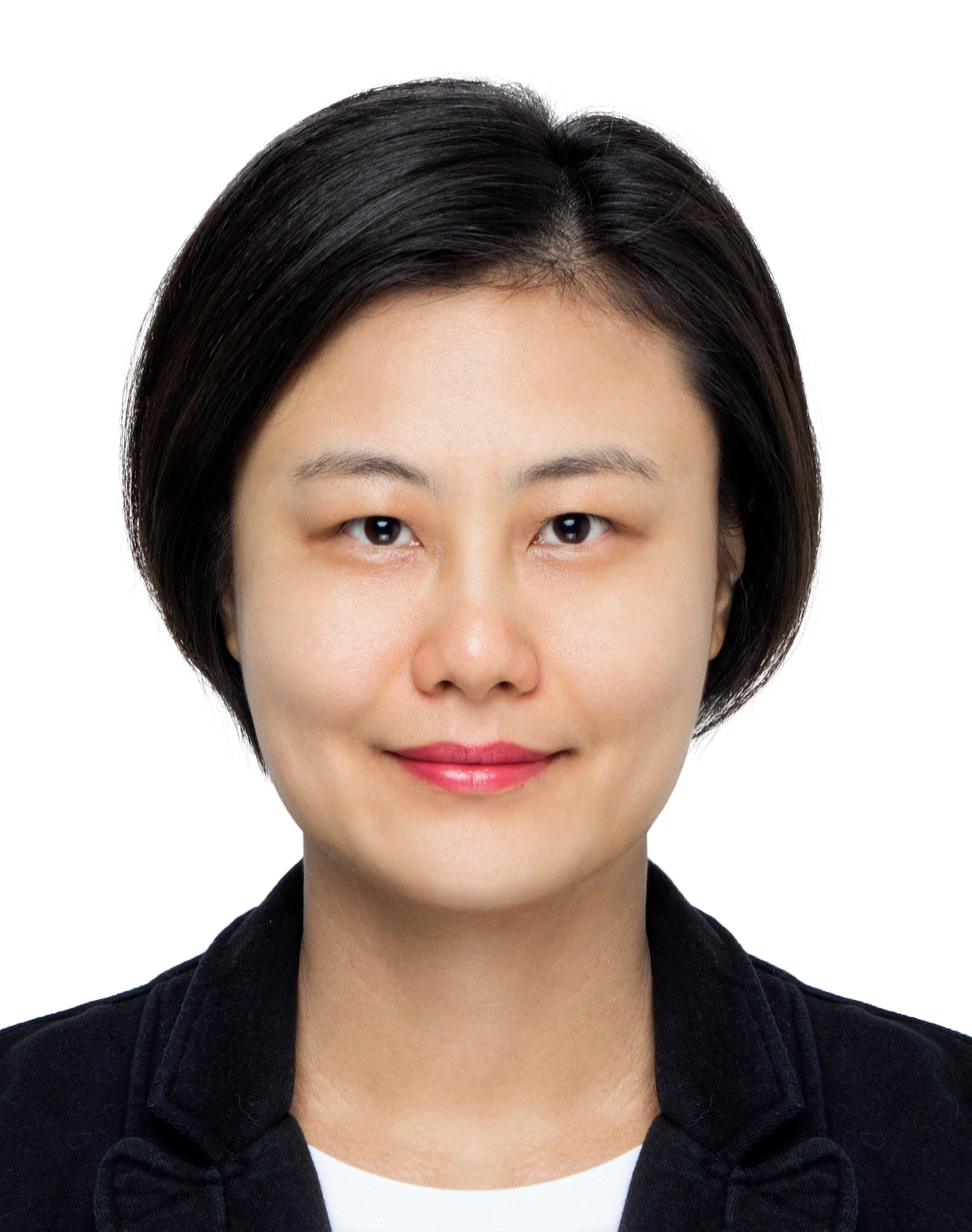}}]{Ran Wang}
(Senior Member IEEE) received her B.Eng. degree in computer science from the College of Information Science and Technology, Beijing Forestry University, Beijing, China, in 2009, and her Ph.D. degree from the Department of Computer Science, City University of Hong Kong, Hong Kong SAR, China, in 2014.

From 2014 to 2016, she was a postdoctoral researcher with the Department of Computer Science, City University of Hong Kong. She is currently an associate professor with the School of Mathematical Sciences, Shenzhen University, China. Her research interests broadly include topics in machine learning and pattern recognition, such as active learning, multi-label learning, deep learning, adversarial robustness, evolutionary optimization, fuzzy sets and fuzzy logic, and their related applications.
\end{IEEEbiography}


\vfill


\begin{thebibliography}{1}
\bibliographystyle{IEEEtran}
\if{}
\bibitem[Lee and Seung(2001)]{lee2001algorithms}
D.~D. Lee and H.~S. Seung, ``Algorithms for non-negative matrix
  factorization,'' in \emph{NeruIPS},
  2001, pp. 556--562.

\bibitem[Lee and Seung(1999)]{lee1999learning}
D.~D. Lee and H.~S. Seung, ``Learning the parts of objects by non-negative
  matrix factorization,'' \emph{Nature}, vol. 401, no. 6755, pp. 788--791,
  1999.

\bibitem[Kuang et~al.(2012)Kuang, Ding, and Park]{kuang2012symmetric}
D.~Kuang, C.~Ding, and H.~Park, ``Symmetric nonnegative matrix factorization
  for graph clustering,'' in \emph{ICDM}.\hskip 1em plus 0.5em minus 0.4em\relax SIAM,
  2012, pp. 106--117.

\bibitem[Kuang et~al.(2015)Kuang, Yun, and Park]{kuang2015symnmf}
D.~Kuang, S.~Yun, and H.~Park, ``Symnmf: nonnegative low-rank approximation of
  a similarity matrix for graph clustering,'' \emph{Journal of Global
  Optimization}, vol.~62, no.~3, pp. 545--574, 2015.

\bibitem[Li and Ding(2006)]{4053063}
T.~Li and C.~Ding, ``The relationships among various nonnegative matrix
  factorization methods for clustering,'' in \emph{ICDM}, 2006, pp. 362--371.

\bibitem[Cai et~al.(2011)Cai, He, Han, and Huang]{5674058}
D.~Cai, X.~He, J.~Han, and T.~S. Huang, ``Graph regularized nonnegative matrix
  factorization for data representation,'' \emph{IEEE Transactions on Pattern
  Analysis and Machine Intelligence}, vol.~33, no.~8, pp. 1548--1560, 2011.

\bibitem[Chen et~al.(2008)Chen, Rege, Dong, and Hua]{chen2008non}
Y.~Chen, M.~Rege, M.~Dong, and J.~Hua, ``Non-negative matrix factorization for
  semi-supervised data clustering,'' \emph{Knowledge and Information Systems},
  vol.~17, pp. 355--379, 2008.

\bibitem[Yang et~al.(2015)Yang, Cao, Jin, Wang, and Meng]{6985550}
L.~Yang, X.~Cao, D.~Jin, X.~Wang, and D.~Meng, ``A unified semi-supervised
  community detection framework using latent space graph regularization,''
  \emph{IEEE Transactions on Cybernetics}, vol.~45, no.~11, pp. 2585--2598,
  2015.

\bibitem[Zhang et~al.(2016)Zhang, Zong, Liu, and Luo]{7167693}
X.~Zhang, L.~Zong, X.~Liu, and J.~Luo, ``Constrained clustering with
  nonnegative matrix factorization,'' \emph{IEEE Transactions on Neural
  Networks and Learning Systems}, vol.~27, no.~7, pp. 1514--1526, 2016.

\bibitem[Wu et~al.(2018)Wu, Jia, Kwong, and Hou]{8361078}
W.~Wu, Y.~Jia, S.~Kwong, and J.~Hou, ``Pairwise constraint propagation-induced
  symmetric nonnegative matrix factorization,'' \emph{IEEE Transactions on
  Neural Networks and Learning Systems}, vol.~29, no.~12, pp. 6348--6361, 2018.

\bibitem[Qin et~al.(2023)Qin, Feng, Ren, and Zhang]{9543530}
Y.~Qin, G.~Feng, Y.~Ren, and X.~Zhang, ``Block-diagonal guided symmetric
  nonnegative matrix factorization,'' \emph{IEEE Transactions on Knowledge and
  Data Engineering}, vol.~35, no.~3, pp. 2313--2325, 2023.

\bibitem[Jia et~al.(2021)Jia, Liu, Hou, and Kwong]{9013063}
Y.~Jia, H.~Liu, J.~Hou, and S.~Kwong, ``Semi-supervised adaptive symmetric
  non-negative matrix factorization,'' \emph{IEEE Transactions on Cybernetics},
  vol.~51, no.~5, pp. 2550--2562, 2021.

\bibitem[Peng et~al.(2023)Peng, Yin, Yang, Chen, and Lin]{10076472}
S.~Peng, J.~Yin, Z.~Yang, B.~Chen, and Z.~Lin, ``Multiview clustering via
  hypergraph induced semi-supervised symmetric nonnegative matrix
  factorization,'' \emph{IEEE Transactions on Circuits and Systems for Video
  Technology}, pp. 1--1, 2023.

\bibitem[Jia et~al.(2023)Jia, Lu, Liu, and Hou]{10007868}
Y.~Jia, G.~Lu, H.~Liu, and J.~Hou, ``Semi-supervised subspace clustering via
  tensor low-rank representation,'' \emph{IEEE Transactions on Circuits and
  Systems for Video Technology}, pp. 1--1, 2023.

\bibitem[Luo et~al.(2021)Luo, Liu, Shang, Lou, and Zhou]{luo2021highly}
X.~Luo, Z.~Liu, M.~Shang, J.~Lou, and M.~C. Zhou, ``Highly-accurate community
  detection via pointwise mutual information-incorporated symmetric
  non-negative matrix factorization,'' \emph{IEEE Transactions on Network
  Science and Engineering}, vol.~8, no.~1, pp. 463--476, 2021.

\bibitem[Gao et~al.(2018)Gao, Guan, and Su]{8637461}
Z.~Gao, N.~Guan, and L.~Su, ``Graph regularized symmetric non-negative matrix
  factorization for graph clustering,'' in \emph{2018 IEEE International
  Conference on Data Mining Workshops}, 2018, pp. 379--384.

\bibitem[Li et~al.(2007)Li, Ding, and Jordan]{4470293}
T.~Li, C.~Ding, and M.~I. Jordan, ``Solving consensus and semi-supervised
  clustering problems using nonnegative matrix factorization,'' in
  \emph{Seventh IEEE International Conference on Data Mining},
  2007, pp. 577--582.

\bibitem[He et~al.(2022)He, Fei, Cheng, Li, Hu, and Tang]{9559733}
C.~He, X.~Fei, Q.~Cheng, H.~Li, Z.~Hu, and Y.~Tang, ``A survey of community
  detection in complex networks using nonnegative matrix factorization,''
  \emph{IEEE Transactions on Computational Social Systems}, vol.~9, no.~2, pp.
  440--457, 2022.

\bibitem[Fu et~al.(2019)Fu, Huang, Sidiropoulos, and Ma]{8653529}
X.~Fu, K.~Huang, N.~D. Sidiropoulos, and W.-K. Ma, ``Nonnegative matrix
  factorization for signal and data analytics: Identifiability, algorithms, and
  applications,'' \emph{IEEE Signal Processing Magazine}, vol.~36, no.~2, pp.
  59--80, 2019.

\bibitem[Yuvaraj and Vivekanandan(2013)]{6508193}
N.~Yuvaraj and P.~Vivekanandan, ``An efficient svm based tumor classification
  with symmetry non-negative matrix factorization using gene expression data,''
  in \emph{2013 International Conference on Information Communication and
  Embedded Systems}, 2013, pp. 761--768.

\bibitem[Liu et~al.(2013)Liu, Lin, Yan, Sun, Yu, and Ma]{6180173}
G.~Liu, Z.~Lin, S.~Yan, J.~Sun, Y.~Yu, and Y.~Ma, ``Robust recovery of subspace
  structures by low-rank representation,'' \emph{IEEE Transactions on Pattern
  Analysis and Machine Intelligence}, vol.~35, no.~1, pp. 171--184, 2013.

\bibitem[Jia et~al.(2022)Jia, Liu, Hou, Kwong, and Zhang]{9620082}
Y.~Jia, H.~Liu, J.~Hou, S.~Kwong, and Q.~Zhang, ``Self-supervised symmetric
  nonnegative matrix factorization,'' \emph{IEEE Transactions on Circuits and
  Systems for Video Technology}, vol.~32, no.~7, pp. 4526--4537, 2022.

\bibitem[Wang and Lu(2015)]{wang2015projection}
W.~Wang and C.~Lu, ``Projection onto the capped simplex,'' \emph{ArXiv Preprint
  arXiv:1503.01002}, 2015.

\bibitem[Chen et~al.(2017)Chen, Sun, and Toh]{chen2017note}
L.~Chen, D.~Sun, and K.-C. Toh, ``A note on the convergence of admm for
  linearly constrained convex optimization problems,'' \emph{Computational
  Optimization and Applications}, vol.~66, no.~2, pp. 327--343, 2017.

\bibitem[Zhao et~al.(2021)Zhao, Chen, and Chen]{8962252}
Y.-P. Zhao, L.~Chen, and C.~L.~P. Chen, ``Laplacian regularized nonnegative
  representation for clustering and dimensionality reduction,'' \emph{IEEE
  Transactions on Circuits and Systems for Video Technology}, vol.~31, no.~1,
  pp. 1--14, 2021.

\bibitem[Lu et~al.(2020)Lu, Feng, Chen, Liu, Lin, and Yan]{8606166}
C.~Lu, J.~Feng, Y.~Chen, W.~Liu, Z.~Lin, and S.~Yan, ``Tensor robust principal
  component analysis with a new tensor nuclear norm,'' \emph{IEEE Transactions
  on Pattern Analysis and Machine Intelligence}, vol.~42, no.~4, pp. 925--938,
  2020.

\bibitem[Ding et~al.(2010)Ding, Li, and Jordan]{ding2010convex}
C.~H. Ding, T.~Li, and M.~I. Jordan, ``Convex and semi-nonnegative matrix
  factorizations,'' \emph{IEEE Transactions on Pattern Analysis and Machine
  Intelligence}, vol.~32, no.~1, pp. 45--55, 2010.

\bibitem[Hartigan and Wong(1979)]{hartigan1979algorithm}
J.~A. Hartigan and M.~A. Wong, ``Algorithm as 136: A k-means clustering
  algorithm,'' \emph{Journal of the Royal Statistical Society. Series C (Applied Statistics)}, vol.~28, no.~1, pp. 100--108, 1979.

\bibitem[He et~al.(2011)He, Xie, Zdunek, Zhou, and Cichocki]{6061964}
Z.~He, S.~Xie, R.~Zdunek, G.~Zhou, and A.~Cichocki, ``Symmetric nonnegative
  matrix factorization: Algorithms and applications to probabilistic
  clustering,'' \emph{IEEE Transactions on Neural Networks}, vol.~22, no.~12,
  pp. 2117--2131, 2011.

\bibitem[Shi et~al.(2015)Shi, Lu, He, and He]{shi2015community}
X.~Shi, H.~Lu, Y.~He, and S.~He, ``Community detection in social network with
  pairwisely constrained symmetric non-negative matrix factorization,'' in
  \emph{Proceedings of the 2015 IEEE/ACM International Conference on Advances
  in Social Networks Analysis and Mining 2015}, 2015, pp. 541--546.

\bibitem[Wang et~al.(2008)Wang, Li, Zhu, and Ding]{wang2008multi}
D.~Wang, T.~Li, S.~Zhu, and C.~Ding, ``Multi-document summarization via
  sentence-level semantic analysis and symmetric matrix factorization,'' in
  \emph{Proceedings of the 31st Annual International ACM SIGIR Conference on Research and Development in Information Retrieval}, 2008, pp. 307--314.

\bibitem[Ng et~al.(2001)Ng, Jordan, and Weiss]{ng2001spectral}
A.~Ng, M.~Jordan, and Y.~Weiss, ``On spectral clustering: Analysis and an
  algorithm,'' \emph{Advances in Neural Information Processing Systems},
  vol.~14, 2001.

\bibitem[Wu et~al.(2013)Wu, Zhang, Hu, and Ji]{wu2013constrained}
B.~Wu, Y.~Zhang, B.-G. Hu, and Q.~Ji, ``Constrained clustering and its
  application to face clustering in videos,'' in \emph{Proceedings of the IEEE
  conference on Computer Vision and Pattern Recognition}, 2013, pp. 3507--3514.

\bibitem[Liu et~al.(2013{\natexlab{b}})Liu, Lin, Yan, Sun, Yu, and
  Ma]{liu2013robust}
G.~Liu, Z.~Lin, S.~Yan, J.~Sun, Y.~Yu, and Y.~Ma, ``Robust recovery of subspace
  structures by low-rank representation,'' \emph{IEEE Transactions on Pattern
  Analysis and Machine Intelligence}, vol.~35, no.~1, pp. 171--184, 2013.

  
\bibitem[Cover and Hart(1967)]{cover1967nearest}
T.~Cover and P.~Hart, ``Nearest neighbor pattern classification,'' \emph{IEEE
  transactions on information theory}, vol.~13, no.~1, pp. 21--27, 1967.


\bibitem[Zhang et~al.(2014)Zhang, Tang, Zhang, and Xue]{zhang2014addressing}
M.~Zhang, J.~Tang, X.~Zhang, and X.~Xue, ``Addressing cold start in recommender
  systems: A semi-supervised co-training algorithm,'' in \emph{Proceedings of
  the 37th international ACM SIGIR conference on Research \& development in
  information retrieval}, 2014, pp. 73--82.

\bibitem[Qiao et~al.(2018)Qiao, Shen, Zhang, Wang, and Yuille]{qiao2018deep}
S.~Qiao, W.~Shen, Z.~Zhang, B.~Wang, and A.~Yuille, ``Deep co-training for
  semi-supervised image recognition,'' in \emph{Proceedings of the european
  conference on computer vision (eccv)}, 2018, pp. 135--152.

\bibitem[Song and Lee(2013)]{song2013hierarchical}
H.~A. Song and S.-Y. Lee, ``Hierarchical representation using nmf,'' in
  \emph{Neural Information Processing: 20th International Conference, ICONIP
  2013, Daegu, Korea, November 3-7, 2013. Proceedings, Part I 20}.\hskip 1em
  plus 0.5em minus 0.4em\relax Springer, 2013, pp. 466--473.

\bibitem[Lu and Peng(2013)]{lu2013exhaustive}
Z.~Lu and Y.~Peng, ``Exhaustive and efficient constraint propagation: A
  graph-based learning approach and its applications,'' \emph{International
  Journal of Computer Vision}, vol. 103, no.~3, p. 306, 2013.

\bibitem[Lv et~al.(2023)Lv, Zhang, Li, Jia, and Chen]{10232925}
W.~Lv, C.~Zhang, H.~Li, X.~Jia, and C.~Chen, ``Joint projection learning and
  tensor decomposition-based incomplete multiview clustering,'' \emph{IEEE
  Transactions on Neural Networks and Learning Systems}, pp. 1--12, 2023.

\bibitem[Zhang et~al.(2023)Zhang, Li, Lv, Huang, Gao, and
  Chen]{zhang2023enhanced}
C.~Zhang, H.~Li, W.~Lv, Z.~Huang, Y.~Gao, and C.~Chen, ``Enhanced tensor
  low-rank and sparse representation recovery for incomplete multi-view
  clustering,'' in \emph{Proceedings of the AAAI conference on artificial
  intelligence}, vol.~37, no.~9, 2023, pp. 11\,174--11\,182.

\bibitem[Kolda and Bader(2009)]{kolda2009tensor}
T.~G. Kolda and B.~W. Bader, ``Tensor decompositions and applications,''
  \emph{SIAM review}, vol.~51, no.~3, pp. 455--500, 2009.

\bibitem[Liu et~al.(2013{\natexlab{c}})Liu, Musialski, Wonka, and Ye]{6138863}
J.~Liu, P.~Musialski, P.~Wonka, and J.~Ye, ``Tensor completion for estimating
  missing values in visual data,'' \emph{IEEE Transactions on Pattern Analysis
  and Machine Intelligence}, vol.~35, no.~1, pp. 208--220, 2013.

\bibitem[Zhang et~al.(2015)Zhang, Fu, Liu, Liu, and Cao]{7410542}
C.~Zhang, H.~Fu, S.~Liu, G.~Liu, and X.~Cao, ``Low-rank tensor constrained
  multiview subspace clustering,'' in \emph{2015 IEEE International Conference
  on Computer Vision (ICCV)}, 2015, pp. 1582--1590.

\bibitem[Wang et~al.(2023)Wang, Chen, Lin, Cen, and Cao]{10102285}
S.~Wang, Y.~Chen, Z.~Lin, Y.~Cen, and Q.~Cao, ``Robustness meets low-rankness:
  Unified entropy and tensor learning for multi-view subspace clustering,''
  \emph{IEEE Transactions on Circuits and Systems for Video Technology},
  vol.~33, no.~11, pp. 6302--6316, 2023.

\bibitem[Jia et~al.(2021{\natexlab{b}})Jia, Liu, Hou, Kwong, and
  Zhang]{9336710}
Y.~Jia, H.~Liu, J.~Hou, S.~Kwong, and Q.~Zhang, ``Multi-view spectral
  clustering tailored tensor low-rank representation,'' \emph{IEEE Transactions
  on Circuits and Systems for Video Technology}, vol.~31, no.~12, pp.
  4784--4797, 2021.

\bibitem[Nishihara et~al.(2015)Nishihara, Lessard, Recht, Packard, and
  Jordan]{nishihara2015general}
R.~Nishihara, L.~Lessard, B.~Recht, A.~Packard, and M.~Jordan, ``A general
  analysis of the convergence of admm,'' in \emph{International conference on
  machine learning}.\hskip 1em plus 0.5em minus 0.4em\relax PMLR, 2015, pp.
  343--352.

\bibitem[Jia et~al.(2020)Jia, Liu, Hou, and Kwong]{9072553}
Y.~Jia, H.~Liu, J.~Hou, and S.~Kwong, ``Clustering-aware graph construction: A
  joint learning perspective,'' \emph{IEEE Transactions on Signal and
  Information Processing over Networks}, vol.~6, pp. 357--370, 2020.
\fi 

\bibitem[Lee and Seung(2001)]{lee2001algorithms}
D.~D. Lee and H.~S. Seung, ``Algorithms for non-negative matrix
  factorization,'' in \emph{NeruIPS},
  2001, pp. 556--562.

\bibitem[Lee and Seung(1999)]{lee1999learning}
D.~D. Lee and H.~S. Seung, ``Learning the parts of objects by non-negative
  matrix factorization,'' \emph{Nature}, vol. 401, no. 6755, pp. 788--791,
  1999.

\bibitem[Kuang et~al.(2012)Kuang, Ding, and Park]{kuang2012symmetric}
D.~Kuang, C.~Ding, and H.~Park, ``Symmetric nonnegative matrix factorization
  for graph clustering,'' in \emph{ICDM}.\hskip 1em plus 0.5em minus 0.4em\relax SIAM,
  2012, pp. 106--117.

\bibitem[Kuang et~al.(2015)Kuang, Yun, and Park]{kuang2015symnmf}
D.~Kuang, S.~Yun, and H.~Park, ``Symnmf: nonnegative low-rank approximation of
  a similarity matrix for graph clustering,'' \emph{Journal of Global
  Optimization}, vol.~62, no.~3, pp. 545--574, 2015.

\bibitem[Li and Ding(2006)]{4053063}
T.~Li and C.~Ding, ``The relationships among various nonnegative matrix
  factorization methods for clustering,'' in \emph{ICDM}, 2006, pp. 362--371.

\bibitem[Cai et~al.(2011)Cai, He, Han, and Huang]{5674058}
D.~Cai, X.~He, J.~Han, and T.~S. Huang, ``Graph regularized nonnegative matrix
  factorization for data representation,'' \emph{IEEE TPAMI}, vol.~33, no.~8, pp. 1548--1560, 2011.

\bibitem[Chen et~al.(2008)Chen, Rege, Dong, and Hua]{chen2008non}
Y.~Chen, M.~Rege, M.~Dong, and J.~Hua, ``Non-negative matrix factorization for
  semi-supervised data clustering,'' \emph{Knowledge and Information Systems},
  vol.~17, pp. 355--379, 2008.

\bibitem[Yang et~al.(2015)Yang, Cao, Jin, Wang, and Meng]{6985550}
L.~Yang, X.~Cao, D.~Jin, X.~Wang, and D.~Meng, ``A unified semi-supervised
  community detection framework using latent space graph regularization,''
  \emph{IEEE TCyb}, vol.~45, no.~11, pp. 2585--2598,
  2015.

\bibitem[Zhang et~al.(2016)Zhang, Zong, Liu, and Luo]{7167693}
X.~Zhang, L.~Zong, X.~Liu, and J.~Luo, ``Constrained clustering with
  nonnegative matrix factorization,'' \emph{IEEE TNNLS}, vol.~27, no.~7, pp. 1514--1526, 2016.

\bibitem[Wu et~al.(2018)Wu, Jia, Kwong, and Hou]{8361078}
W.~Wu, Y.~Jia, S.~Kwong, and J.~Hou, ``Pairwise constraint propagation-induced
  symmetric nonnegative matrix factorization,'' \emph{IEEE TNNLS}, vol.~29, no.~12, pp. 6348--6361, 2018.

\bibitem[Qin et~al.(2023)Qin, Feng, Ren, and Zhang]{9543530}
Y.~Qin, G.~Feng, Y.~Ren, and X.~Zhang, ``Block-diagonal guided symmetric
  nonnegative matrix factorization,'' \emph{IEEE TKDE}, vol.~35, no.~3, pp. 2313--2325, 2023.

\bibitem[Jia et~al.(2021)Jia, Liu, Hou, and Kwong]{9013063}
Y.~Jia, H.~Liu, J.~Hou, and S.~Kwong, ``Semi-supervised adaptive symmetric
  non-negative matrix factorization,'' \emph{IEEE TCyb},
  vol.~51, no.~5, pp. 2550--2562, 2021.

\bibitem[Peng et~al.(2023)Peng, Yin, Yang, Chen, and Lin]{10076472}
S.~Peng, J.~Yin, Z.~Yang, B.~Chen, and Z.~Lin, ``Multiview clustering via
  hypergraph induced semi-supervised symmetric nonnegative matrix
  factorization,'' \emph{IEEE TCSVT}, pp. 1--1, 2023.

\bibitem[Jia et~al.(2023)Jia, Lu, Liu, and Hou]{10007868}
Y.~Jia, G.~Lu, H.~Liu, and J.~Hou, ``Semi-supervised subspace clustering via
  tensor low-rank representation,'' \emph{IEEE TCSVT}, pp. 1--1, 2023.

\bibitem[Luo et~al.(2021)Luo, Liu, Shang, Lou, and Zhou]{luo2021highly}
X.~Luo, Z.~Liu, M.~Shang, J.~Lou, and M.~C. Zhou, ``Highly-accurate community
  detection via pointwise mutual information-incorporated symmetric
  non-negative matrix factorization,'' \emph{IEEE TNNLS}, vol.~8, no.~1, pp. 463--476, 2021.

\bibitem[Gao et~al.(2018)Gao, Guan, and Su]{8637461}
Z.~Gao, N.~Guan, and L.~Su, ``Graph regularized symmetric non-negative matrix
  factorization for graph clustering,'' in \emph{ICDM Workshops}, 2018, pp. 379--384.

\bibitem[Li et~al.(2007)Li, Ding, and Jordan]{4470293}
T.~Li, C.~Ding, and M.~I. Jordan, ``Solving consensus and semi-supervised
  clustering problems using nonnegative matrix factorization,'' in
  \emph{ICDM},
  2007, pp. 577--582.

\bibitem[He et~al.(2022)He, Fei, Cheng, Li, Hu, and Tang]{9559733}
C.~He, X.~Fei, Q.~Cheng, H.~Li, Z.~Hu, and Y.~Tang, ``A survey of community
  detection in complex networks using nonnegative matrix factorization,''
  \emph{IEEE TCSS}, vol.~9, no.~2, pp.
  440--457, 2022.

\bibitem[Fu et~al.(2019)Fu, Huang, Sidiropoulos, and Ma]{8653529}
X.~Fu, K.~Huang, N.~D. Sidiropoulos, and W.-K. Ma, ``Nonnegative matrix
  factorization for signal and data analytics: Identifiability, algorithms, and
  applications,'' \emph{IEEE SPM}, vol.~36, no.~2, pp.
  59--80, 2019.

\bibitem[Yuvaraj and Vivekanandan(2013)]{6508193}
N.~Yuvaraj and P.~Vivekanandan, ``An efficient svm based tumor classification
  with symmetry non-negative matrix factorization using gene expression data,''
  in \emph{ICICES}, 2013, pp. 761--768.

\bibitem[Liu et~al.(2013)Liu, Lin, Yan, Sun, Yu, and Ma]{6180173}
G.~Liu, Z.~Lin, S.~Yan, J.~Sun, Y.~Yu, and Y.~Ma, ``Robust recovery of subspace
  structures by low-rank representation,'' \emph{IEEE TPAMI}, vol.~35, no.~1, pp. 171--184, 2013.

\bibitem[Jia et~al.(2022)Jia, Liu, Hou, Kwong, and Zhang]{9620082}
Y.~Jia, H.~Liu, J.~Hou, S.~Kwong, and Q.~Zhang, ``Self-supervised symmetric
  nonnegative matrix factorization,'' \emph{IEEE TCSVT}, vol.~32, no.~7, pp. 4526--4537, 2022.

\bibitem[Wang and Lu(2015)]{wang2015projection}
W.~Wang and C.~Lu, ``Projection onto the capped simplex,'' \emph{ArXiv Preprint
  arXiv:1503.01002}, 2015.

\bibitem[Chen et~al.(2017)Chen, Sun, and Toh]{chen2017note}
L.~Chen, D.~Sun, and K.-C. Toh, ``A note on the convergence of admm for
  linearly constrained convex optimization problems,'' \emph{Computational
  Optimization and Applications}, vol.~66, no.~2, pp. 327--343, 2017.

\bibitem[Zhao et~al.(2021)Zhao, Chen, and Chen]{8962252}
Y.-P. Zhao, L.~Chen, and C.~L.~P. Chen, ``Laplacian regularized nonnegative
  representation for clustering and dimensionality reduction,'' \emph{IEEE
  TCSVT}, vol.~31, no.~1,
  pp. 1--14, 2021.

\bibitem[Lu et~al.(2020)Lu, Feng, Chen, Liu, Lin, and Yan]{8606166}
C.~Lu, J.~Feng, Y.~Chen, W.~Liu, Z.~Lin, and S.~Yan, ``Tensor robust principal
  component analysis with a new tensor nuclear norm,'' \emph{IEEE TPAMI}, vol.~42, no.~4, pp. 925--938,
  2020.

\bibitem[Ding et~al.(2010)Ding, Li, and Jordan]{ding2010convex}
C.~H. Ding, T.~Li, and M.~I. Jordan, ``Convex and semi-nonnegative matrix
  factorizations,'' \emph{IEEE TPAMI}, vol.~32, no.~1, pp. 45--55, 2010.

\bibitem[Hartigan and Wong(1979)]{hartigan1979algorithm}
J.~A. Hartigan and M.~A. Wong, ``Algorithm as 136: A k-means clustering
  algorithm,'' \emph{Journal of the Royal Statistical Society. Series C (Applied Statistics)}, vol.~28, no.~1, pp. 100--108, 1979.

\bibitem[He et~al.(2011)He, Xie, Zdunek, Zhou, and Cichocki]{6061964}
Z.~He, S.~Xie, R.~Zdunek, G.~Zhou, and A.~Cichocki, ``Symmetric nonnegative
  matrix factorization: Algorithms and applications to probabilistic
  clustering,'' \emph{IEEE TNN}, vol.~22, no.~12,
  pp. 2117--2131, 2011.

\bibitem[Shi et~al.(2015)Shi, Lu, He, and He]{shi2015community}
X.~Shi, H.~Lu, Y.~He, and S.~He, ``Community detection in social network with
  pairwisely constrained symmetric non-negative matrix factorization,'' in
  \emph{ICASNAM}, 2015, pp. 541--546.

\bibitem[Wang et~al.(2008)Wang, Li, Zhu, and Ding]{wang2008multi}
D.~Wang, T.~Li, S.~Zhu, and C.~Ding, ``Multi-document summarization via
  sentence-level semantic analysis and symmetric matrix factorization,'' in
  \emph{SIGIR}, 2008, pp. 307--314.

\bibitem[Ng et~al.(2001)Ng, Jordan, and Weiss]{ng2001spectral}
A.~Ng, M.~Jordan, and Y.~Weiss, ``On spectral clustering: Analysis and an
  algorithm,'' \emph{NeurIPS},
  vol.~14, 2001.

\bibitem[Wu et~al.(2013)Wu, Zhang, Hu, and Ji]{wu2013constrained}
B.~Wu, Y.~Zhang, B.-G. Hu, and Q.~Ji, ``Constrained clustering and its
  application to face clustering in videos,'' in \emph{CVPR}, 2013, pp. 3507--3514.

\bibitem[Liu et~al.(2013{\natexlab{b}})Liu, Lin, Yan, Sun, Yu, and
  Ma]{liu2013robust}
G.~Liu, Z.~Lin, S.~Yan, J.~Sun, Y.~Yu, and Y.~Ma, ``Robust recovery of subspace
  structures by low-rank representation,'' \emph{IEEE TPAMI}, vol.~35, no.~1, pp. 171--184, 2013.

  
\bibitem[Cover and Hart(1967)]{cover1967nearest}
T.~Cover and P.~Hart, ``Nearest neighbor pattern classification,'' \emph{IEEE
  TIT}, vol.~13, no.~1, pp. 21--27, 1967.


\bibitem[Zhang et~al.(2014)Zhang, Tang, Zhang, and Xue]{zhang2014addressing}
M.~Zhang, J.~Tang, X.~Zhang, and X.~Xue, ``Addressing cold start in recommender
  systems: A semi-supervised co-training algorithm,'' in \emph{SIGIR}, 2014, pp. 73--82.

\bibitem[Qiao et~al.(2018)Qiao, Shen, Zhang, Wang, and Yuille]{qiao2018deep}
S.~Qiao, W.~Shen, Z.~Zhang, B.~Wang, and A.~Yuille, ``Deep co-training for
  semi-supervised image recognition,'' in \emph{ECCV}, 2018, pp. 135--152.

\bibitem[Song and Lee(2013)]{song2013hierarchical}
H.~A. Song and S.-Y. Lee, ``Hierarchical representation using nmf,'' in
  \emph{ ICONIP
  2013, Daegu, Korea, November 3-7, 2013. Proceedings, Part I 20}.\hskip 1em
  plus 0.5em minus 0.4em\relax Springer, 2013, pp. 466--473.

\bibitem[Lu and Peng(2013)]{lu2013exhaustive}
Z.~Lu and Y.~Peng, ``Exhaustive and efficient constraint propagation: A
  graph-based learning approach and its applications,'' \emph{IJCV}, vol. 103, no.~3, p. 306, 2013.

\bibitem[Lv et~al.(2023)Lv, Zhang, Li, Jia, and Chen]{10232925}
W.~Lv, C.~Zhang, H.~Li, X.~Jia, and C.~Chen, ``Joint projection learning and
  tensor decomposition-based incomplete multiview clustering,'' \emph{IEEE
  TNNLS}, pp. 1--12, 2023.

\bibitem[Zhang et~al.(2023)Zhang, Li, Lv, Huang, Gao, and
  Chen]{zhang2023enhanced}
C.~Zhang, H.~Li, W.~Lv, Z.~Huang, Y.~Gao, and C.~Chen, ``Enhanced tensor
  low-rank and sparse representation recovery for incomplete multi-view
  clustering,'' in \emph{AAAI}, vol.~37, no.~9, 2023, pp. 11\,174--11\,182.

\bibitem[Kolda and Bader(2009)]{kolda2009tensor}
T.~G. Kolda and B.~W. Bader, ``Tensor decompositions and applications,''
  \emph{SIAM review}, vol.~51, no.~3, pp. 455--500, 2009.

\bibitem[Liu et~al.(2013{\natexlab{c}})Liu, Musialski, Wonka, and Ye]{6138863}
J.~Liu, P.~Musialski, P.~Wonka, and J.~Ye, ``Tensor completion for estimating
  missing values in visual data,'' \emph{IEEE TPAMI}, vol.~35, no.~1, pp. 208--220, 2013.

\bibitem[Zhang et~al.(2015)Zhang, Fu, Liu, Liu, and Cao]{7410542}
C.~Zhang, H.~Fu, S.~Liu, G.~Liu, and X.~Cao, ``Low-rank tensor constrained
  multiview subspace clustering,'' in \emph{ICCV}, 2015, pp. 1582--1590.

\bibitem[Wang et~al.(2023)Wang, Chen, Lin, Cen, and Cao]{10102285}
S.~Wang, Y.~Chen, Z.~Lin, Y.~Cen, and Q.~Cao, ``Robustness meets low-rankness:
  Unified entropy and tensor learning for multi-view subspace clustering,''
  \emph{IEEE TCSVT},
  vol.~33, no.~11, pp. 6302--6316, 2023.

\bibitem[Jia et~al.(2021{\natexlab{b}})Jia, Liu, Hou, Kwong, and
  Zhang]{9336710}
Y.~Jia, H.~Liu, J.~Hou, S.~Kwong, and Q.~Zhang, ``Multi-view spectral
  clustering tailored tensor low-rank representation,'' \emph{IEEE TCSVT}, vol.~31, no.~12, pp.
  4784--4797, 2021.

\bibitem[Nishihara et~al.(2015)Nishihara, Lessard, Recht, Packard, and
  Jordan]{nishihara2015general}
R.~Nishihara, L.~Lessard, B.~Recht, A.~Packard, and M.~Jordan, ``A general
  analysis of the convergence of admm,'' in \emph{ICML}.\hskip 1em plus 0.5em minus 0.4em\relax PMLR, 2015, pp.
  343--352.

\bibitem[Jia et~al.(2020)Jia, Liu, Hou, and Kwong]{9072553}
Y.~Jia, H.~Liu, J.~Hou, and S.~Kwong, ``Clustering-aware graph construction: A
  joint learning perspective,'' \emph{IEEE TSIPN}, vol.~6, pp. 357--370, 2020.


\end{thebibliography}
\end{document}